\pdfoutput=1
\documentclass[11pt]{article}

\usepackage{amsmath,amssymb,amsthm}
\usepackage{graphicx}
\usepackage{longtable}
\usepackage{placeins}
\usepackage{tikz}
\usetikzlibrary{positioning,arrows.meta}
\usepackage{forest}
\usepackage[authoryear,round]{natbib}

\usepackage[margin=1in]{geometry}

\usepackage[plain,noend]{algorithm2e}

\theoremstyle{plain}
\newtheorem{theorem}{Theorem}

\newtheorem{corollary}{Corollary}

\theoremstyle{definition}

\theoremstyle{remark}

\makeatletter
\renewcommand{\algocf@captiontext}[2]{#1\algocf@typo. \AlCapFnt{}#2}

\def\@algocf@capt@plain{top}
\renewcommand{\algocf@makecaption}[2]{%
	\addtolength{\hsize}{\algomargin}%
	\sbox\@tempboxa{\algocf@captiontext{#1}{#2}}%
	\ifdim\wd\@tempboxa >\hsize%
	\hskip .5\algomargin%
	\parbox[t]{\hsize}{\algocf@captiontext{#1}{#2}}%
	\else%
	\global\@minipagefalse%
	\hbox to\hsize{\box\@tempboxa}%
	\fi%
	\addtolength{\hsize}{-\algomargin}%
}
\makeatother

\graphicspath{{./art/}{./figures/}}

\DeclareMathOperator{\Var}{Var}
\DeclareMathOperator{\Cov}{Cov}

\usepackage[hidelinks]{hyperref}

\let\origparagraph\paragraph \renewcommand{\paragraph}[1]{\origparagraph{#1.}}

\title{A Theory of Cross-Validation Variance}
\author{Nathaniel S. O'Connell \quad Jaime L. Speiser\\
	Department of Biostatistics and Data Science\\
	Wake Forest University School of Medicine\\
	Winston-Salem, NC, United States\\
	Corresponding author: \texttt{nathaniel.oconnell@wfusm.edu}}
\date{} 

\begin{document}
	
	\maketitle
	
\begin{abstract}
	Cross-validation estimates predictive performance by repeatedly splitting the data into training and validation sets, fitting the model on the first and evaluating it on the second. Every observation influences the estimate in two roles, through its own evaluation when it is held out and through the models it helps fit when it trains. The random splits determine its weight in each role. From this representation, we prove an exact finite-sample identity for the variance of the cross-validated estimate. For a given splitting scheme with a common training set size, the variance is a function of these weights, with constants of the model, the metric, and the population. It covers metrics averaged over observations, such as mean squared error, and over case--control pairs, such as the AUC. The identity shows which schemes to use. The variance decomposes into sampling variance, which depends only on the training size, and variance due to the random fold splitting, which arises when the weights change from one draw of the splits to the next. \(K\)-fold cross-validation, whose weights are equal across observations by design, is therefore variance optimal at first order; Monte Carlo and bootstrap cross-validation are not. The first-order variance of \(K\)-fold does not depend on \(K\), and under reasonable conditions on the model leave-one-out has the smallest variance of any \(K\)-fold scheme. For the AUC the same holds for leave-pair-out, which evaluates every case--control pair. Finally, we show the cross-validated estimate is asymptotically normal with the identity's full variance. Together, these results provide a general theoretical foundation for inference on cross-validated performance and establish principles for selecting variance-optimal cross-validation designs.
\end{abstract}


	\section{Introduction}
\label{sec:introduction}

Cross-validation is among the most widely used procedures in statistics, machine learning, and data science. It estimates a model's predictive performance by repeatedly splitting a dataset into training and validation subsets, fitting the model on the first, evaluating it on the second, and averaging the held-out evaluations \citep{arlot2010}. The most common forms of cross-validation --  $K$-fold and repeated $K$-fold \citep{stone1974,geisser1975}, Monte Carlo split-sampling, and the bootstrap \citep{efron1983,efron1997} -- differ only in how they generate the splits. Under every form, each observation influences the estimate in two ways: it is evaluated when it is held out, and when it is in a training set it helps fit the models that evaluate the other observations. The held-out evaluations are correlated through the training data they share.

Despite its widespread use, no general theory explains how the model, training fraction, and resampling design jointly determine cross-validation variance. Existing results study individual cross-validation designs in isolation. \citet{burman1989} derived variance expansions for $K$-fold and repeated $K$-fold cross-validation, \citet{nadeau2003} proposed a corrected variance for Monte Carlo splitting based on the correlation between split estimates, and \citet{bengio2004} parameterized the variance of $K$-fold at a fixed partition by three covariances among the held-out errors and showed that no distribution-free unbiased estimator of it exists. For least-squares density estimation, \citet{arlot2016} showed that $K$-fold has smaller variance than Monte Carlo splitting at matched training size and attributed the difference to the balanced use of observations. Fixed-model variance formulas, such as DeLong's for the AUC \citep{delong1988}, quantify only the uncertainty from the evaluation observations. None of these separates an observation's two roles, validation and training.

In parallel, others have studied the asymptotic distribution of cross-validation estimates under stability and regularity conditions \citep{dudoit2005,ledell2015,lei2020,li2023}. \citet{austern2020} proved a central limit theorem centered at the expected performance of the model, for a number of folds that may grow with $n$, whose asymptotic variance includes the variability of the fitted models. \citet{Bayle2020} built their interval from the sample variance of the out-of-fold losses, valid for a random target, the average performance of the $K$ models fitted in that run, and \citet{lei2025cv} gave a stability condition under which the two targets coincide. \citet{bates2023} made a related point from the estimation side, that cross-validation estimates the performance of a model trained on a random sample rather than of one fitted model. \citet{cai2025} and \citet{bayle2025} studied the bootstrap of the whole procedure and the comparison of two models. These results identify validation, training, and cross-covariance components of the asymptotic variance of $K$-fold. They do not express these components through the roles an observation plays across designs, or give an exact finite-sample decomposition of the variance itself.

We treat cross-validation as a randomized design: both the sample and the splits vary, and the splits determine how each observation is weighted in validation and training. The main result is a variance identity that holds exactly, at every sample size, for every design that trains on the same fraction of the data in every split (Theorems~\ref{thm:dual_role} and~\ref{thm:variance_identity}). It covers continuous and binary outcomes alike: metrics that average over observations, such as squared error and the Brier score, and, given the labels, metrics that average over case--control pairs, such as the AUC (Theorem~\ref{thm:pairwise_identity}). Existing representations of the variance \citep{burman1989,nadeau2003,bengio2004,austern2020} are special cases of the identity (Sections~\ref{subsec:refitting}, \ref{subsec:split_noise}, and~\ref{sec:normality}).

The identity separates sampling variance, attributable to random sampling of the data, from split noise, caused by randomly assigning observations to folds. At a fixed training fraction every design has the same sampling variance and differs only in its split noise, which arises when the weights vary from one draw of the splits to the next, so a balanced design, whose weights do not vary, is variance optimal. $K$-fold is balanced at first order, unlike Monte Carlo and bootstrap CV (Theorem~\ref{thm:balanced_optimal}, Corollary~\ref{cor:kfold_optimal}). Its first-order variance does not depend on $K$ (Corollary~\ref{cor:choice_of_K}). Under additional conditions, a fixed fitting budget favors one run with as many folds as possible, approaching leave-one-out as its limit (Corollary~\ref{cor:loo_limit}). For AUC and other pairwise metrics, ordinary $K$-fold evaluates only pairs sharing a fold, adding split noise that grows with $K$. Paired $K$-fold \citep{yousef2019} evaluates every case--control pair, removes this noise, and approaches leave-pair-out (Section~\ref{subsec:folds_auc}).  Finally, we prove asymptotic normality with the full variance given the splits, including higher orders. With positive limiting variance, a consistent estimate yields valid Wald intervals and tests (Theorem~\ref{thm:clt}). Related to this, we identify which terms several existing variance estimators capture (Section~\ref{subsec:existing_estimators}).

Section~\ref{sec:framework} sets up the design, the role weights and the influences, Section~\ref{sec:identity} proves the identity, Section~\ref{sec:implications} derives the design results, Section~\ref{sec:normality} proves normality, and Section~\ref{sec:empirical_validation} checks the identity against variances measured over repeated datasets. Proofs and further numerical results are in the Supplementary Material, whose sections are numbered S1--S6.

\section{A Design-Based Representation of Cross-Validation}
\label{sec:framework}

We show how validation and training weights arise from the random splits of the data, and how each observation's influence in these roles enters the variance identity of Section~\ref{sec:identity}. We also state the two conditions used throughout. Unit-level metrics, such as the Brier score and log-loss, and pairwise metrics, such as the AUC, are set up side by side, since the identity treats both with the same weights and influences. Notation is tabulated in Supplement~\ref{appendix:notation}.

\subsection{Data, models, and performance metrics}
\label{subsec:data_algorithm_metric}

Let $P$ denote the underlying, unknown population distribution, and let $\mathcal{D}_n = \{Z_i\}_{i=1}^n$ be an i.i.d.\ sample from $P$. Each observation $Z_i = (X_i, Y_i)$ consists of a covariate vector $X_i$ and an outcome $Y_i$. A model is fit to a training set $T \subset \{1,\dots,n\}$ to produce the fitted model $\hat{f}_T$, a function mapping covariates to predictions. Any model that produces predictions from data is covered, including but not limited to parametric regression models, tree-based ensembles, neural networks, and other statistical and machine learning procedures. 

Predictive performance is assessed through a metric that averages contributions from \emph{basic units}. For a unit-level metric, such as the Brier score, log-loss, or squared error, a basic unit is one observation and its contribution is the loss of its prediction; for a pairwise metric, such as the AUC, a basic unit is a pair of observations and its contribution compares the two predictions. Let $\mathcal{B}$ be the set of basic units and $\ell_i(\hat{f}_T)$ the contribution of unit $i$ at $\hat{f}_T$.

For a binary outcome the pairs are case--control pairs, and the contribution \(\mathcal C(s,t)\) compares the prediction \(s\) for the case with the prediction \(t\) for the control; the AUC has \(\mathcal C(s,t)=\mathbb I(s>t)+\tfrac12\mathbb I(s=t)\), and the Gini coefficient, Somers' \(D\) and the pairwise ranking loss are affine in it. The results for pairwise metrics rest on two properties: each observation enters all of its pairs in the same role, as a case or as a control, so its contributions averaged over its partners form one function of it, and the pairs are fixed by the design and the labels. Metrics without these properties are not covered, such as the concordance index under censoring, in which the comparable pairs are selected by the data.

\subsection{Cross-validation as a randomized design}
\label{subsec:cv_design}

A cross-validation procedure is specified by a \emph{design} $D$, which fixes the resampling mechanism, a per-split training fraction $p \in (0,1]$, and the number of splits $S$. Applied to the index set $\{1,\dots,n\}$, the design generates $S$ splits $\{(T_r, V_r)\}_{r=1}^S$ into a training set and a validation set, represented by their \emph{inclusion structure}, the $n \times S$ array $I = (I_{ir})$ with $I_{ir} = \mathbb I(i\in T_r)$. The design is a probability distribution over $I$: its training fraction fixes the probability that an observation is in a given training set, and its resampling mechanism fixes how the indicators depend on one another within and across splits. For the bootstrap, observation $i$ enters the training resample of split $r$ with multiplicity $\tilde I_{ir}\in\{0,1,2,\dots\}$, and $I_{ir}=\mathbb I(\tilde I_{ir}\ge 1)$. The inclusion structure, with the multiplicities when applicable, is the only random variable the design introduces. Fitting the model to each $T_r$ and evaluating it on $V_r$ yields the cross-validation estimate, $\widehat{R}_D$.

We restrict attention to \emph{ordinary} designs:
\begin{enumerate}
	\item[(i)] An observation is never evaluated on a model it helped fit: $V_r = T_r^c$ for each split $r$.
	\item[(ii)] Each split operates at a fixed training fraction $p \in (0,1]$, in the sense that $|T_r|/n \to p$ as $n \to \infty$.
	\item[(iii)] The design does not distinguish observations: permuting the observation indices leaves the distribution of $I$ unchanged.
\end{enumerate}
This class includes $K$-fold cross-validation with $p = (K-1)/K$ and its repeated version with $R$ repetitions, leave-one-out, Monte Carlo cross-validation, and the bootstrap, whose distinct training fraction tends to $p = 1 - e^{-1}$. Throughout, $K$ divides $n$. A design is \emph{fixed-size} if every training set has the same size $m$, with $m/n\to p$. $K$-fold, its repetitions, leave-one-out, and Monte Carlo cross-validation are fixed-size. The bootstrap is not, because sampling with replacement makes the size of the validation set vary. For a pairwise metric, a design is \emph{class-stratified} if every validation set has the same numbers of cases and of controls, so that every training set has \(m_1\) cases and \(m_0\) controls. The number of splits $S$ is set by the analyst, with $S=RK$ for repeated $K$-fold, and the results of Sections~\ref{sec:identity} and~\ref{sec:implications} hold for every $S$. Bootstrap variants that mix training and validation contributions, such as the .632 rule, fit the weighted form below but need a separate variance treatment (Supplement~\ref{appendix:alt_cv_methods}).

The cross-validation estimate targets $R_{p,n} := \mathbb{E}\{\ell_i(\hat f_T)\}$, the expected contribution of a held-out unit when the model is fit to $m$ other observations. The expectation is over the training sample and the held-out unit. Bootstrap training sets have no common size, so for the bootstrap the target is $\mathbb E(\widehat R_D)$. For a binary outcome, the \emph{labels} are the observed outcomes \(Y\), which mark each observation as a case or a control and so determine which pairs are case--control pairs. The pairwise analysis conditions on the labels. Under a class-stratified design every training set then has \(m_1\) cases and \(m_0\) controls, fixed proportions of the \(n_1\) cases and \(n_0\) controls in the sample, and the target is \(R_{p,n}(Y)\), the expected value of the metric for a model fitted to \(m_1\) cases and \(m_0\) controls. Unlike \(R_{p,n}\), it varies from sample to sample, through the class counts \(n_1\) and \(n_0\).

Whichever the target, the quantity this paper studies is the variance of $\widehat R_D$ under the joint randomness of the data and the inclusion structure,
\[
\Var\!\left(\widehat{R}_D\right) := \Var_{\mathcal{D}_n, I}\!\left(\widehat{R}_D(\mathcal{D}_n, I)\right).
\]
Conditional on $I$, $\Var(\widehat{R}_D \mid I)$ measures variation across datasets with the splits held fixed.

\subsection{Weighted representation and role weights}
\label{subsec:weighted_representation}

Cross-validation designs generate splits differently, but their estimates share one weighted form: each held-out unit contributes whenever it is validated, and those contributions are averaged across splits. Let $\ell_i^{(r)}$ denote the contribution of unit $i$ when it is validated in split $r$. A pair is validated when both of its members are, so that both predictions come from the same fitted model.

The estimate averages all validation evaluations equally. Let \(M:=\sum_r|V_r|\) be the total number of evaluations, \(N_i:=\sum_r\mathbb I(i\in V_r)\) the number of times unit \(i\) is validated, and \(\ell_i\) the average of unit \(i\)'s contributions. Then
\[
\widehat R_D=\frac1M\sum_{r=1}^S\sum_{i\in V_r}\ell_i^{(r)}=\sum_{i\in\mathcal B}v_i\ell_i,
\qquad v_i:=\frac{N_i}{M},
\]
where \(v_i\), the \emph{validation weight} of unit \(i\), is the fraction of all validation evaluations contributed by unit \(i\), with \(\sum_i v_i=1\).
An observation also influences the splits where it is not evaluated through the models it helps train. The \emph{training weight} of observation \(i\) is its share of all training use,
\[
w_i:=\sum_{r=1}^S\frac{|V_r|}{M}\,\frac{I_{ir}}{|T_r|},\qquad\sum_{i=1}^n w_i=1,
\]
It multiplies split \(r\)'s share \(|V_r|/M\) of the estimate by observation \(i\)'s share \(1/|T_r|\) of its training set, then sums over splits that train on \(i\). For the bootstrap, \(I_{ir}/|T_r|\) is replaced by \(\tilde I_{ir}/n\). Together, \(v_i\) and \(w_i\) are the \emph{role weights}, one for each of the two roles. Other ways of aggregating the evaluations, such as averaging split-level estimates, change only these weights and their higher-order counterparts (Supplement~\ref{appendix:alt_cv_methods}).

Under each design, with $q := 1 - p$, the weights are
\begin{itemize}
	\item $K$-fold cross-validation with $R$ repetitions: every observation is validated $R$ times and included in $R(K-1)$ training sets of size $m$, so $v_i = w_i = 1/n$ for every $i$.
	\item Monte Carlo cross-validation with $S$ splits of validation size $nq$: $v_i = N_i/(Snq)$ and $w_i = (S - N_i)/(Snp)$, so that $w_i = 1/(np) - (q/p)\,v_i$.
	\item Bootstrap cross-validation with $S$ resamples: $v_i = N_i/M$, with $N_i$ the number of resamples that leave $i$ out, and $w_i = \sum_r (|V_r|/M)(\tilde I_{ir}/n)$. The two weights are separate random quantities, since $N_i$ does not determine how many times the other resamples draw $i$.
\end{itemize}
In each design the validation weights are of order $1/n$. A design is \emph{balanced} if $v_i = w_i = 1/n$ for every observation. $K$-fold cross-validation is balanced. Monte Carlo and bootstrap cross-validation are not, because the number of times an observation is validated is random.

The units of a pairwise metric are the case--control pairs. Let \(M_A>0\) be the total number of validated case--control pairs, summed over the splits. Each pair's validation weight is the number of splits that validate it divided by \(M_A\), and with these weights the estimate is the averaged \(K\)-fold AUC of \citet{airola2011} under exact class stratification. An observation influences the estimate through every validated pair that contains it, so its validation weight is its total weight \(\tilde v_i\) through those pairs. Every validated pair contains one case and one control, so the \(\tilde v_i\) sum to one within each class. The training weight \(w_i\) is as defined above, with the share \(|V_r|/M\) of split \(r\) replaced by its share of the \(M_A\) validated pairs. With \(n_y\) the number of observations of class \(y\), under class-stratified \(K\)-fold every observation has \(\tilde v_i=1/n_{Y_i}\) and \(w_i=1/n\), the pairwise counterpart of balance.

\subsection{Influences and conditions}
\label{subsec:assumptions}

The identity depends on how much an observation moves the estimate in each of its roles. We distinguish three effects: a unit's validation influence, a training observation's effect on its contribution, and the joint effect of several observations (Section~\ref{subsec:representation}). The first-order variance uses the first two. Take a unit-level metric, so that unit \(i\) is observation \(i\), and let \(i\) be validated in split \(r\). Let
\[
\ell_i^\star := \mathbb{E}\big(\ell_i^{(r)} \,\big|\, Z_i\big)
\]
be its contribution with the training observations averaged out, its \emph{validation influence}. Its mean is \(R_{p,n}\). For a training observation \(j\in T_r\), let
\[
\phi_i(z) := |T_r|\,\Big\{\mathbb{E}\big(\ell_i^{(r)} \,\big|\, Z_i,\, Z_j = z\big) - \mathbb{E}\big(\ell_i^{(r)} \,\big|\, Z_i\big)\Big\}
\]
be the influence of a training observation at \(z\) on unit \(i\)'s expected contribution, scaled by the training-set size. For each \(Z_i\) its mean over \(z\) is zero. Averaging over the validated observation gives the \emph{training influence} \(\bar\phi(z):=\mathbb{E}_{Z_i}\{\phi_i(z)\}\), the influence of a training observation at \(z\) on the fitted model's expected performance. The validation and training influences are defined for any model and any metric with finite variance. For a fixed-size design neither depends on the split, and both depend on \(n\) through the training size, a subscript we suppress.

A pairwise metric has the same two influences, defined given the labels. Which pairs are case--control pairs depends on the labels, so the pairwise analysis conditions on the label vector \(Y\) throughout. Given \(Y\), the covariates are independent, with \(X_i\) distributed as \(X\) given \(Y_i\), and they are the only random part of the data. For a case \(i\) and a control \(j\) validated together in split \(r\), the case's validation influence
\[
\ell_i^\star:=\mathbb E\big[\mathcal C\{\hat f_{T_r}(X_i),\hat f_{T_r}(X_j)\}\,\big|\,X_i,Y\big]
\]
is its contribution with its partner and the training observations averaged out. A control's validation influence is defined in the same way, with its case partner averaged out. For a class-stratified design its mean is \(R_{p,n}(Y)\) in either class. The training influence \(\bar\phi_Y\) is \(\bar\phi\) with every expectation also given \(Y\), so that \(\bar\phi_Y(x)\) is the influence of a training observation at \(x\) on the fitted model's expected metric given the labels. Given \(Y\), a training observation's covariate has the distribution of its class, so \(\bar\phi_Y\) has mean zero within each class. Both influences are defined within a class because a case and a control play different roles in their pairs.

Two conditions are used throughout: A1 on the data, the design and the model, and A2 on how much of the variance the joint effects of several observations carry.

\smallskip\noindent\textbf{A1 (Data, design, model).} \emph{\(\mathcal{D}_n\) is i.i.d.\ from \(P\). The inclusion structure is independent of \(\mathcal D_n\), or for a pairwise metric of the covariates given the labels, and the design is ordinary. The contributions \(\ell_i^{(r)}\) have finite variance, and \(\sup_n[\Var(\ell_i^\star)+\Var\{\bar\phi(Z)\}]<\infty\). The model is fit symmetrically in its training observations, with any internal randomization held fixed or averaged out.}

A1 states that the splits are independent of the data. A class-stratified design forms its folds from the labels, so for a pairwise metric they need only be independent of the covariates given the labels. A symmetric fit depends on its training set and not on the order of its members, which, with i.i.d.\ data and a common training size, makes an observation's influence the same in every split that trains on it (Section~\ref{subsec:representation}). For a model with internal randomization, such as a random forest, the same symmetric fitting rule is applied in every split, with the seeds held fixed in a way that preserves its symmetry or averaged over, so the variance studied here is over the data and the splits, not over the seeds.

\smallskip\noindent\textbf{A2 (Dominance).} A2 concerns the parts of the estimate that depend jointly on sets of observations. Fix the inclusion structure \(I\). For a set \(\mathcal S\) of observations write \(Z_{\mathcal S}\) for their values. Let \(g\) be a square-integrable function of the observations. For each subset \(\mathcal A\) of \(\mathcal S\), including the empty set and \(\mathcal S\) itself, \(\mathbb E(g\mid Z_{\mathcal A})\) is the conditional mean of \(g\) given the observations in \(\mathcal A\). The Hoeffding component of \(g\) on \(\mathcal S\) is the alternating sum of these conditional means,
\[
g_{\mathcal S}:=\sum_{\mathcal A\subseteq\mathcal S}(-1)^{|\mathcal S|-|\mathcal A|}\,\mathbb E(g\mid Z_{\mathcal A}).
\]
For one observation, \(g_{\{i\}}=\mathbb E(g\mid Z_i)-\mathbb E(g)\) is what knowing \(Z_i\) changes in the expected value of \(g\). For a pair, \(g_{\{i,j\}}=\mathbb E(g\mid Z_i,Z_j)-\mathbb E(g\mid Z_i)-\mathbb E(g\mid Z_j)+\mathbb E(g)\) is what knowing both changes beyond what each changes alone. In general \(g_{\mathcal S}\) is the part of \(g\) that depends jointly on the observations in \(\mathcal S\), with every effect of a smaller set removed. Components on different sets are uncorrelated, and \(g-\mathbb E(g)=\sum_{\mathcal S\ne\varnothing}g_{\mathcal S}\) \citep{hoeffding1948}. Write \(h_{\mathcal S}\) for the components of \(\widehat R_D\) given \(I\), or given \(I\) and \(Y\) for a pairwise metric, and let
\[
\rho_n^{(k)}(I):=\sum_{|\mathcal S|=k}\mathbb E\big(h_{\mathcal S}^2\,\big|\,I\big),
\qquad
\rho_n^{(k+)}(I):=\sum_{l\ge k}\rho_n^{(l)}(I)
\]
be the variance carried by the components of order \(k\), and by those of order \(k\) and above. For a pairwise metric they are written \(\rho_n^{(k)}(I,Y)\) and \(\rho_n^{(k+)}(I,Y)\).

\emph{At order \(k\), \(\mathbb E_I\{\rho_n^{((k+1)+)}(I)\}=o(n^{-k})\). A2(i) and A2(ii) denote A2 at orders one and two.}

A2 limits the variance from joint effects of observations, rather than restricting the form of the model. For a function \(g\) of the data let \(\nabla_jg:=g(\mathcal D_n)-g(\mathcal D_n^{(j)})\), where \(\mathcal D_n^{(j)}\) replaces \(Z_j\) by an independent copy, and \(\nabla_j\nabla_kg:=\nabla_j(\nabla_kg)\). For fixed-size designs whose validation weights are of order \(1/n\), bounds on such differences of the contributions imply A2 at every order, and A2(i) holds in the influence-function case, in which a held-out contribution is its value at the limit of the fitted model plus an average of one term per training observation, up to a remainder small in mean square; this covers M-estimators and fixed-bandwidth smoothers (Supplement~\ref{appendix:stability_assumptions}). The models that meet the stability conditions of \citet{austern2020} at a fixed number of folds also satisfy A2(i) (Supplement~\ref{appendix:proof_containment}). A2(i) is not needed for a normal limit (Theorem~\ref{thm:clt}).

\section{The Variance Identity}
\label{sec:identity}

At every sample size, the cross-validation variance is exactly a finite series. Each order is a quadratic form in design-dependent weights, with constants determined by the model, metric, and population. Theorems~\ref{thm:dual_role} and~\ref{thm:variance_identity} prove it for unit-level metrics and Theorem~\ref{thm:pairwise_identity} for pairwise metrics.

\subsection{The dual-role representation}
\label{subsec:representation}

Expectations are under the joint distribution of the data and the inclusion structure unless stated otherwise. Conditioning on \(I\) and applying the law of total variance,
\begin{equation}
	\label{eq:lotv}
	\Var\!\left(\widehat{R}_D\right)
	=
	\mathbb{E}_I\!\left\{\Var\!\left(\widehat{R}_D \mid I\right)\right\}
	+
	\Var_I\!\left\{\mathbb{E}\!\left(\widehat{R}_D \mid I\right)\right\}.
\end{equation}
For a fixed-size design the second term is zero. Given \(I\), a validated observation is independent of the training data used to fit its model, and every training set has size \(m\), so \(\mathbb{E}(\widehat R_D\mid I)=R_{p,n}\) for every realized \(I\). The variance is then the average of \(\Var(\widehat R_D\mid I)\) over the design, and the rest of this section decomposes that conditional variance.

A held-out contribution \(\ell_i^{(r)}\) depends only on \(Z_i\) and the observations in \(T_r\). Thus a set \(\mathcal S\) of \(k\ge2\) observations can contribute only when one member is validated and the rest train, or when every member trains. In the latter case, its component is that of the fitted model's expected performance, \(R(\hat f_T):=\mathbb{E}_{Z_i}\{\ell_i(\hat f_T)\}\), because a held-out observation is independent of the training data. Under i.i.d. sampling, a symmetric fit, and common training size \(m\), each split type contributes the same function of \(Z_{\mathcal S}\). The component of \(\widehat R_D\) on \(\mathcal S\) is therefore that function times the number of relevant splits.

The counts are the weights of order \(k\). For an observation \(i\) and a set \(\mathcal J\) of \(k-1\ge1\) other observations, and for a set \(\mathcal S\) of \(k\ge2\) observations, let
\[
c^{(k)}_{i;\mathcal J}:=\frac{1}{M m^{k-1}}\sum_{r=1}^S\mathbb I(i\in V_r)\prod_{j\in\mathcal J}I_{jr},
\qquad
d^{(k)}_{\mathcal S}:=\frac{1}{m^{k}}\sum_{r=1}^S\frac{|V_r|}{M}\prod_{j\in\mathcal S}I_{jr}.
\]

Here \(c^{(k)}_{i;\mathcal J}\) weights the joint influence of validated observation \(i\) and training observations \(\mathcal J\), reflecting how often they occur in those roles. The weight \(d^{(k)}_{\mathcal S}\) reflects how often all members of \(\mathcal S\) train together, scaling their joint influence on expected performance. Both are determined by \(I\) alone. At \(k=2\) they are the \emph{pair weights} \(c_{ij}\) and \(d_{ij}\), and at \(k=1\), with \(\mathcal J\) empty, they are the role weights \(v_i\) and \(w_i\).

The fixed functions are the \emph{kernels} of the Hoeffding components: for a set \(\mathcal S\), the part of a held-out contribution and the part of the fitted model's expected performance that depend on the observations in \(\mathcal S\) jointly. For \(i\in V_r\) and sets \(\mathcal J\subseteq T_r\) of \(k-1\) and \(\mathcal S\subseteq T_r\) of \(k\) training observations, let
\[
\phi^{(k)}(Z_i;Z_{\mathcal J}):=m^{k-1}\,\big(\ell_i^{(r)}\big)_{\{i\}\cup\mathcal J},
\qquad
\chi^{(k)}(Z_{\mathcal S}):=m^{k}\,\{R(\hat f_{T_r})\}_{\mathcal S},
\]
where a set in the subscript denotes the Hoeffding component on that set. By the argument above, neither kernel depends on the split or on which training observations are chosen. At \(k=1\) they are the two influences of Section~\ref{subsec:assumptions}, \(\ell_i^\star-R_{p,n}\) and \(\bar\phi\). At \(k=2\) the first is \(\phi_i(Z_j)-\bar\phi(Z_j)\), the part of a training observation's influence that is specific to validated observation \(i\). The second, defined as \(\chi\), measures how much two training observations together move expected performance beyond the sum of their separate moves. When every training observation affects the fitted model everywhere, as for a parametric fit, one observation moves expected performance at order \(1/m\) and a pair at order \(1/m^2\), and the scaling makes these kernels of order one. When the prediction at a point rests on a few training observations, joint effects are larger and the variance of the kernels can grow with \(m\).

\begin{theorem}[Dual-role representation]
	\label{thm:dual_role}
	Assume A1, let the design be fixed-size, and fix a realized inclusion structure \(I\). For a unit-level metric, the Hoeffding component of \(\widehat R_D\) on one observation is
	\[
	h_i(Z_i)
	=
	v_i\,(\ell_i^\star-R_{p,n})
	+
	w_i\,\bar\phi(Z_i),
	\qquad i=1,\dots,n,
	\]
	and on a set \(\mathcal S\) of \(k\ge2\) observations is
	\begin{equation}
		\label{eq:order_k_kernel}
		h_{\mathcal S}
		=
		\sum_{i\in\mathcal S}c^{(k)}_{i;\mathcal S\setminus\{i\}}\,\phi^{(k)}(Z_i;Z_{\mathcal S\setminus\{i\}})
		+
		d^{(k)}_{\mathcal S}\,\chi^{(k)}(Z_{\mathcal S}).
	\end{equation}
\end{theorem}
The proof is in Supplement~\ref{appendix:IF_derivation}. At \(k=2\) the two kinds of split give the two orientations of the validation--training interaction, \(c_{ij}\phi^{(2)}(Z_i;Z_j)\) and \(c_{ji}\phi^{(2)}(Z_j;Z_i)\), and the training--training interaction \(d_{ij}\chi(Z_i,Z_j)\). Summing over the observations,
\[
\widehat R_D-R_{p,n}
=
\sum_{i=1}^n v_i\,(\ell_i^\star-R_{p,n})
+
\sum_{i=1}^n w_i\,\bar\phi(Z_i)
+
\varrho_D,
\]
where \(\varrho_D\) contains the components on two or more observations, is uncorrelated with every function of a single observation, and has \(\mathbb{E}(\varrho_D^2\mid I)=\rho_n^{(2+)}(I)\). The first sum is the aggregate validation influence and the second is the aggregate training influence from refitting. Observations are independent, so the validation and training sums can covary only through the two influences of the same observation.

\subsection{Role moments and the variance identity}
\label{subsec:variance_decomposition}

The variance of the single-observation component involves three constants, the variance of an observation's validation influence, \(\sigma_V^2:=\Var(\ell_i^\star)\), the variance of its training influence on expected performance, \(\sigma_T^2:=\Var\{\bar\phi(Z)\}\), and their covariance, \(\gamma_0:=\Cov\{\ell_i^\star,\bar\phi(Z_i)\}\). The \(h_i(Z_i)\) are independent across \(i\) and uncorrelated with \(\varrho_D\), so for a fixed-size design,
\begin{equation}
	\label{eq:cond_var_roles}
	\Var\!\left(\widehat{R}_D \mid I\right)
	=
	\sigma_V^2\sum_{i=1}^n v_i^2
	+
	\sigma_T^2\sum_{i=1}^n w_i^2
	+
	2\gamma_0\sum_{i=1}^n v_iw_i
	+
	\rho_n^{(2+)}(I),
\end{equation}
with \(\rho_n^{(2+)}(I)\ge0\). The design affects this part only through the three weight sums.

The part \(\rho_n^{(2+)}(I)\) of \eqref{eq:cond_var_roles} is, order by order, a quadratic form in the weights of that order, as the first-order part is, with one constant for each kind of product of the kernels in \eqref{eq:order_k_kernel}. For observations \(Z_1,\dots,Z_k\), \(k\ge2\), write \(\phi^{(k)}_l\) for \(\phi^{(k)}\) with \(Z_l\) validated and the others training, and \(\chi^{(k)}\) for \(\chi^{(k)}(Z_1,\dots,Z_k)\). Define
\[
\tau_{k,n}^2:=\Var(\phi^{(k)}_1),
\quad
\kappa_{k,n}:=\mathbb E(\phi^{(k)}_1\phi^{(k)}_2),
\quad
\sigma_{\chi,k,n}^2:=\Var(\chi^{(k)}),
\quad
\xi_{k,n}:=\mathbb E(\phi^{(k)}_1\chi^{(k)}).
\]
Here \(\tau_{k,n}^2\) is the variance of the validation--training interaction, \(\kappa_{k,n}\) is the covariance between two of its orientations, \(\sigma_{\chi,k,n}^2\) is the variance of the training--training interaction, and \(\xi_{k,n}\) is the covariance between the two kinds of interaction. At \(k=2\) the order is dropped, giving the \emph{pair constants} \(\tau_n^2\), \(\kappa_n\), \(\sigma_{\chi,n}^2\), and \(\xi_n\), with \(\tau_n^2=\mathbb E\{\phi_i(Z_j)^2\}-\sigma_T^2\). Squaring the components \eqref{eq:order_k_kernel} and summing over the sets of size \(k\),
\begin{equation}
	\label{eq:cond_var_order}
	\begin{aligned}
		\rho_n^{(k)}(I)
		=
		\sum_{|\mathcal S|=k}\Big\{&
		\tau_{k,n}^2\sum_{i\in\mathcal S}\big(c^{(k)}_{i;\mathcal S\setminus\{i\}}\big)^2
		+\kappa_{k,n}\sum_{i\ne i'\in\mathcal S}c^{(k)}_{i;\mathcal S\setminus\{i\}}c^{(k)}_{i';\mathcal S\setminus\{i'\}}\\
		&+\sigma_{\chi,k,n}^2\big(d^{(k)}_{\mathcal S}\big)^2
		+2\xi_{k,n}\,d^{(k)}_{\mathcal S}\sum_{i\in\mathcal S}c^{(k)}_{i;\mathcal S\setminus\{i\}}
		\Big\}.
	\end{aligned}
\end{equation}
The simplest model shows what these constants are. Take the training-set mean as the fitted model and evaluate it by squared error, which for a binary outcome is the Brier score of the fitted risk. Each held-out contribution \((Y_i-\bar Y_{T_r})^2\) is a quadratic in the observations, so its components on three or more observations are zero and the identity stops at order two. With \(\sigma^2:=\Var(Y)\), which is \(\theta(1-\theta)\) for a binary outcome with mean \(\theta\), the four pair constants are explicit and of order \(\sigma^4\),
\[
\tau_n^2=\kappa_n=\sigma_{\chi,n}^2=-\xi_n=4\sigma^4.
\]
For the sample mean under squared error, the validation--training and training--training pair interactions are equal in magnitude and opposite in sign, giving \(\xi_n=-4\sigma^4\). This negative covariance matters for comparing fold counts, but its sign depends on the learner and loss (Section~\ref{subsec:loo}).

Averaging \eqref{eq:cond_var_roles} and \eqref{eq:cond_var_order} over the inclusion structure replaces each weight sum by its average, and these averages are all the design contributes. Define the first-order role moments
\[
\Gamma_D:=\mathbb E_I\Big(\sum_i v_i^2\Big),\qquad
\Lambda_D:=\mathbb E_I\Big(\sum_i w_i^2\Big),\qquad
\Omega_D:=\mathbb E_I\Big(\sum_i v_iw_i\Big),
\]
and, at order \(k\ge2\), the role moments \(\Theta_D^{(k)}\), \(\Theta_D'^{(k)}\), \(\Psi_D^{(k)}\), and \(\Psi_D'^{(k)}\), the expectations over \(I\) of the four weight sums in \eqref{eq:cond_var_order}, in order. At \(k=2\) the order is dropped. These moments measure how unevenly the design weights observations and their interactions.

\begin{theorem}[Variance identity]
	\label{thm:variance_identity}
	Let \(D\) be a fixed-size design with \(S\) splits, and assume A1. For a unit-level metric,
	\[
	\begin{aligned}
		\Var(\widehat R_D)
		={}&\sigma_V^2\Gamma_D+\sigma_T^2\Lambda_D+2\gamma_0\Omega_D\\
		&+\sum_{k=2}^{n}\big\{\tau_{k,n}^2\Theta_D^{(k)}+\kappa_{k,n}\Theta_D'^{(k)}+\sigma_{\chi,k,n}^2\Psi_D^{(k)}+2\xi_{k,n}\Psi_D'^{(k)}\big\},
	\end{aligned}
	\]
	where the term of order \(k\) equals \(\mathbb E_I\{\rho_n^{(k)}(I)\}\ge0\).
\end{theorem}

The proof is in Supplement~\ref{appendix:proof_variance_identity}. The identity is exact at every sample size, and all design dependence enters through the role moments, which is how Section~\ref{sec:implications} compares designs at a common training size. The terms of order \(k>m+1\) are zero, because a set of observations enters a split together only as one validated observation and at most \(m\) observations of its training set. Each term of the series is nonnegative, so every truncation is a lower bound on the variance under A1 alone, even though individual constants such as \(\kappa_{k,n}\) and \(\xi_{k,n}\) can be negative. For a balanced design \(\Gamma_D=\Lambda_D=\Omega_D=1/n\) exactly, and Section~\ref{subsec:design_comparison} presents them for Monte Carlo and bootstrap cross-validation.

Under A2(i) the components on two or more observations carry variance of smaller order than \(1/n\), and only the single-observation terms remain. The variance of a unit-level metric is then the first-order part of Theorem~\ref{thm:variance_identity},
\[
\sigma_V^2\Gamma_D+\sigma_T^2\Lambda_D+2\gamma_0\Omega_D,
\]
up to \(o(n^{-1})\), a quadratic form in the two role weights with one constant for each role and one for their covariance.

The bootstrap is not fixed-size, so Theorem~\ref{thm:variance_identity} does not cover it. Under a condition A3 on how the fit responds to the multiplicities of a resample, stated in Supplement~\ref{appendix:proof_bootstrap_identity}, the first-order part of the identity holds for it up to \(o(n^{-1})\), with the influences defined per unit of multiplicity and \(\mathbb E(\widehat R_D)\) in place of \(R_{p,n}\).

\subsection{Pairwise metrics}
\label{subsec:pairwise_metrics}

For pairwise metrics, one evaluation involves two observations rather than one. The variance identity extends to this setting, with three changes to account for the case--control pair.

First, which pairs count depends on the labels, so all expectations in this subsection are taken given the labels, with the components in the covariates. Let \(m_{1r}:=\sum_{l\in T_r}Y_l\) be the number of cases in the training set of split \(r\), its \emph{training composition}, and \(\mu_m(m_1/m)\) the expected value of the metric for a model fitted to \(m_1\) cases and \(m-m_1\) controls, so that a pair validated in split \(r\) has expected contribution \(\mu_m(m_{1r}/m)\). Changing the training class mix changes the fitted model and hence its kernels. Under a class-stratified design every split has the same composition, so the constants are single numbers and the target is fixed, as under a common training size for a unit-level metric.

Second, an observation is weighted through its validated pairs, which it enters in one role, so its kernels are specific to its class. The weights count a split by its validated pairs that contain \(i\), and by all its validated pairs, over the splits of one composition at a time:
\[
\begin{gathered}
	c^{(k)}_{i;\mathcal J,m_1}:=\frac{1}{M_Am^{k-1}}\sum_{r:\,m_{1r}=m_1}\mathbb I(i\in V_r)\Big\{\sum_{j\in V_r}\mathbb I(Y_j\ne Y_i)\Big\}\prod_{l\in\mathcal J}I_{lr},\\
	d^{(k)}_{\mathcal S,m_1}:=\frac{1}{M_Am^{k}}\sum_{r:\,m_{1r}=m_1}\Big\{\sum_{j,j'\in V_r}\mathbb I(Y_j=1,\,Y_{j'}=0)\Big\}\prod_{l\in\mathcal S}I_{lr}.
\end{gathered}
\]
Summed over compositions and at order one they are \(\tilde v_i\) and \(w_i\) of Section~\ref{subsec:weighted_representation}; \(\tilde v_{i,m_1}\) and \(w_{i,m_1}\) are the parts from composition \(m_1\), and \(s_{m_1}\) is the share of the validated pairs from splits of that composition. The kernels \(\phi^{(k)}_{y,m_1}\) and \(\chi^{(k)}_{m_1}\) are those of Section~\ref{subsec:representation} for a validated observation of class \(y\), with its partner averaged out, and a model trained at composition \(m_1\); at order one they are \(\ell^\star_{i,m_1}-\mu_m(m_1/m)\) and \(\bar\phi_{Y,m_1}\).

Third, a pair has two validated observations, so a split can contribute to the component on \(\mathcal S\) in a third way: a case \(i\) and a control \(j\) of \(\mathcal S\) are validated together while the rest of \(\mathcal S\) trains their model. With \(\mathcal J\) a set of \(k-2\) training observations, its weight and kernel are
\[
\begin{gathered}
	e^{(k)}_{ij;\mathcal J,m_1}:=\frac{1}{M_Am^{k-2}}\sum_{r:\,m_{1r}=m_1}\mathbb I(i,j\in V_r)\prod_{l\in\mathcal J}I_{lr},\\
	\zeta^{(k)}_{m_1}(X_i,X_j;X_{\mathcal J}):=m^{k-2}\big[\mathcal C\{\hat f_{T_r}(X_i),\hat f_{T_r}(X_j)\}\big]_{\{i,j\}\cup\mathcal J}.
\end{gathered}
\]
At \(k=2\), \(e^{(2)}_{ij}\) summed over compositions is the validation weight of the pair, and \(\zeta^{(2)}(X_i,X_j)\), the validation-pair interaction, is the part of the pair's contribution that depends on both members jointly. Unit-level metrics have no counterpart, since a unit-level contribution has one validated observation.

The argument of Theorem~\ref{thm:dual_role}, applied given the labels, then gives the Hoeffding component of \(\widehat R_D\) on one observation,
\[
h_i(X_i)=\sum_{m_1}\big[\tilde v_{i,m_1}\{\ell^\star_{i,m_1}-\mu_m(m_1/m)\}+w_{i,m_1}\,\bar\phi_{Y,m_1}(X_i)\big],
\qquad i=1,\dots,n,
\]
and on a set \(\mathcal S\) of \(k\ge2\) observations,
\[
\begin{aligned}
	h_{\mathcal S}
	=\sum_{m_1}\bigg\{&
	\sum_{i\in\mathcal S}c^{(k)}_{i;\mathcal S\setminus\{i\},m_1}\,\phi^{(k)}_{Y_i,m_1}(X_i;X_{\mathcal S\setminus\{i\}})
	+
	d^{(k)}_{\mathcal S,m_1}\,\chi^{(k)}_{m_1}(X_{\mathcal S})\\
	&+
	\sum_{\substack{i,j\in\mathcal S\\ Y_i=1,\,Y_j=0}}e^{(k)}_{ij;\mathcal S\setminus\{i,j\},m_1}\,\zeta^{(k)}_{m_1}(X_i,X_j;X_{\mathcal S\setminus\{i,j\}})\bigg\},
\end{aligned}
\]
with kernels that depend on the set only through its class counts (Supplement~\ref{appendix:proof_auc_identity}).

The first-order constants are taken within each class and at a pair of compositions: for an observation of class \(y\), \(\sigma_{V\mid y}(m_1,m_1')\), \(\sigma_{T\mid y}(m_1,m_1')\) and \(\gamma_{0\mid y}(m_1,m_1')\) are the covariances, given \(Y\), of its validation influences, of its training influences, and of a validation influence with a training influence, under models trained at compositions \(m_1\) and \(m_1'\). At a single composition they are the class constants \(\sigma_{V\mid y}^2\), \(\sigma_{T\mid y}^2\) and \(\gamma_{0\mid y}\), and the class role moments \(\Gamma_{D,y}\), \(\Lambda_{D,y}\), \(\Omega_{D,y}\) are the averages over the design of \(\sum_{Y_i=y}\tilde v_i^2\), \(\sum_{Y_i=y}w_i^2\), \(\sum_{Y_i=y}\tilde v_iw_i\) given \(Y\). For any design whose law is invariant to relabeling within a class, all of these depend on \(Y\) only through the class counts \(n_1\) and \(n_0=n-n_1\).

\begin{theorem}[Variance identity for pairwise metrics]
	\label{thm:pairwise_identity}
	Assume A1, let the metric be pairwise, and let the design be fixed-size. Given the labels and a realized inclusion structure \(I\):
	\begin{itemize}
		\item[(a)] The conditional mean is \(\mathbb E(\widehat R_D\mid I,Y)=\sum_{m_1}s_{m_1}\,\mu_m(m_1/m)\), and
		\[
		\begin{aligned}
			\Var(\widehat R_D\mid I,Y)
			={}&
			\sum_i\sum_{m_1,m_1'}\big\{
			\tilde v_{i,m_1}\tilde v_{i,m_1'}\,\sigma_{V\mid Y_i}(m_1,m_1')
			+w_{i,m_1}w_{i,m_1'}\,\sigma_{T\mid Y_i}(m_1,m_1')\\
			&\qquad\qquad
			+2\tilde v_{i,m_1}w_{i,m_1'}\,\gamma_{0\mid Y_i}(m_1,m_1')\big\}
			+\sum_{k=2}^n\rho_n^{(k)}(I,Y),
		\end{aligned}
		\]
		where \(\rho_n^{(k)}(I,Y)\ge0\) is a quadratic form in the weights of order \(k\) above, with constants indexed by pairs of compositions and by the numbers of cases and controls in a set.
		\item[(b)] If the design is class-stratified, \(\mathbb E(\widehat R_D\mid I,Y)=R_{p,n}(Y)\) for every realized \(I\), and the variance given the class counts is
		\[
		\begin{aligned}
			\Var(\widehat R_D\mid n_1)
			={}&\sum_{y\in\{0,1\}}\big(\sigma_{V\mid y}^2\Gamma_{D,y}+\sigma_{T\mid y}^2\Lambda_{D,y}+2\gamma_{0\mid y}\Omega_{D,y}\big)\\
			&+\sum_{k=2}^{n}\mathbb E_I\big\{\rho_n^{(k)}(I,Y)\,\big|\,Y\big\},
		\end{aligned}
		\]
		where \(\rho_n^{(k)}(I,Y)\) is \eqref{eq:cond_var_order} given \(Y\), with constants indexed by the numbers of cases and controls in a set and the weights \(e^{(k)}\) added.
	\end{itemize}
\end{theorem}

The proof is in Supplement~\ref{appendix:proof_auc_identity}. Without class stratification the conditional mean in (a) varies with the splits, and by \eqref{eq:lotv} given \(Y\) its variance across split draws is added to the average of (a) over the design. Every term of (a) and (b) is nonnegative, so every truncation is a lower bound. Averaging (b) over the class counts adds \(\Var\{R_{p,n}(Y)\}\), the effect of the labels on the target through the training composition. The proofs use only the two properties of Section~\ref{subsec:data_algorithm_metric}, so they cover a pair average without labels, whose first-order identity is Theorem~\ref{thm:variance_identity} with the validation influence doubled.

Sections~\ref{sec:implications} and~\ref{sec:normality} are stated in the unit-level form, a quadratic form in the two role weights with three constants, and the corollary below gives that form for pairwise metrics. The pair-specific part \(\zeta^{(2)}\) drops out at first order because a pair's validation weight is of order \(n^{-2}\): summed over the \(n^2\) pairs, its variance is of order \(n^{-2}\). It enters at second order, where under class-stratified \(K\)-fold it adds split noise that increases in \(K\) (Section~\ref{subsec:folds_auc}). It remains to replace the composition-indexed constants by limits. For a design that splits without regard to the labels, the training case fractions \(m_{1r}/m\) concentrate near the population class proportion \(\pi_1\), where \(\pi_y:=\mathbb P(Y=y)\), so the limits are taken there.

\smallskip\noindent\textbf{A4 (Pairwise metric).} \emph{(i) \(\mathbb E_{I,Y}\{\rho_n^{(2+)}(I,Y)\}=o(n^{-1})\). (ii) For a training set of \(m\) observations with \(m_1\) cases, the centered validation influence \(\ell_i^\star-\mu_m(m_1/m)\) and the training influence \(\bar\phi_Y\) of an observation of either class converge in \(L_2\) as \(m\to\infty\), uniformly for \(m_1/m\) near \(\pi_1\), to limits continuous in \(m_1/m\) at \(\pi_1\). (iii) \(\mu_m\) has a differentiable extension near \(\pi_1\) whose derivative converges uniformly there to a limit \(\mu_\infty'\) continuous at \(\pi_1\).}

A4(i) is the counterpart of A2(i). Under A4(ii) the composition-indexed covariances in Theorem~\ref{thm:pairwise_identity}(a) converge to the class constants, and two features of pairwise metrics determine the unit-level constants. The weights \(\tilde v_i\) sum to one within each class, so an observation of class \(y\) has validation weight \(v_i/\pi_y\) to first order, and its validation influence on the unit-level scale is its centered validation influence divided by \(\pi_y\). The label of a training observation shifts the case fraction of the training set and with it the expected metric, by \(\mu_\infty'(\pi_1)\) per unit of case fraction under A4(iii). The unit-level constants are therefore
\[
\sigma_A^2:=\frac{\sigma_{V\mid1}^2}{\pi_1}+\frac{\sigma_{V\mid0}^2}{\pi_0},
\qquad
\gamma_A:=\gamma_{0\mid1}+\gamma_{0\mid0},
\qquad
\sigma_T^2=\pi_1\sigma_{T\mid1}^2+\pi_0\sigma_{T\mid0}^2+\pi_1\pi_0\,\mu_\infty'(\pi_1)^2,
\]
the last being \(\sigma_T^2\) of Section~\ref{subsec:representation} with the labels random (Supplement~\ref{appendix:auc_label_part}); its final term is the \emph{label part}, which in Theorem~\ref{thm:pairwise_identity} comes from the variance of the conditional mean in (a) across split draws and from \(\Var\{R_{p,n}(Y)\}\).

\begin{corollary}[Pairwise metrics at first order]
	\label{prop:pairwise_reduction}
	Assume A1 and A4 with a bounded pair contribution, and let the design be fixed-size with \(p<1\), independent of the labels, and with a number of splits polynomial in \(n\). Then \eqref{eq:cond_var_roles} holds with \(\sigma_V^2\) and \(\gamma_0\) replaced by \(\sigma_A^2\) and \(\gamma_A\) at their limits, \(v_i\) and \(w_i\) the unit-level role weights, and \(\rho_n^{(2+)}(I)\) replaced by a remainder of expected absolute value \(o(n^{-1})\) over the design. The same holds with \(v_i=w_i=1/n\), that is, \(\Var(\widehat R_D)=(\sigma_A^2+\sigma_T^2+2\gamma_A)/n+o(n^{-1})\), for stratified \(K\)-fold cross-validation, whose folds have class counts that differ by at most one, if adding one case or one control to a training set moves the expected value of the metric by \(o(n^{-1/2})\), uniformly over training sets of size at least \(pn-2\) with case fraction near \(\pi_1\).
\end{corollary}
The proof is in Supplement~\ref{appendix:pairwise_kernel_mapping}. At first order a pairwise metric behaves as a unit-level metric with constants \(\sigma_A^2\), \(\sigma_T^2\), \(\gamma_A\). For stratified \(K\)-fold every training composition is fixed up to one observation, so under the corollary's condition the conditional mean in Theorem~\ref{thm:pairwise_identity}(a) varies by \(o(n^{-1/2})\) and the label part enters the first-order variance only through the variance of the target across class counts, \(n\Var\{R_{p,n}(Y)\}\to\pi_1\pi_0\mu_\infty'(\pi_1)^2\). If \(K\) divides both class counts, stratified \(K\)-fold is class-stratified.

\subsection{Refitting and established variance formulas}
\label{subsec:refitting}

Refitting enters first-order variance through training-influence variance \(\sigma_T^2\) and its covariance with validation, \(\gamma_0\) (or \(\gamma_A\) for pairwise metrics). A variance formula for a fixed model treats the fitted model as given and keeps only the validation term.

For a learner evaluated with the smooth loss it minimizes, \(\bar\phi\) converges to the directional derivative of expected loss. That derivative goes to zero at the limiting minimizer, so \(\sigma_T^2\) and \(\gamma_0\) vanish and only the validation term remains at first order (Supplement~\ref{appendix:proof_refit_second_order}). A fixed-model variance formula is therefore right at first order for logistic regression fit by maximum likelihood and evaluated by log-loss, but misses the training terms when cross-validation reports a loss the model was not fit to, such as the Brier score, unless the limiting fit minimizes that loss as well. The argument does not apply to the AUC, whose contribution is an indicator on a pair, not a smooth loss of one observation.

Established variance formulas use the variance \(\sigma_\ell^2:=\Var(\ell_i^{(r)})\) of a held-out contribution and the covariances between held-out contributions, and the training terms are inside those covariances. In the influence-function case of Section~\ref{subsec:assumptions}, \(\sigma_\ell^2=\sigma_V^2+o(1)\), and two contributions of units \(i\ne i'\), validated on splits \(r\) and \(r'\) of a fixed-size design, have covariance
\begin{equation}
	\label{eq:pair_cov}
	\Cov\big(\ell_i^{(r)},\ell_{i'}^{(r')}\,\big|\,I\big)
	=\frac{|T_r\cap T_{r'}|}{m^2}\,\sigma_T^2
	+\frac{\mathbb I(i\in T_{r'})+\mathbb I(i'\in T_r)}{m}\,\gamma_0
	+o(n^{-1}).
\end{equation}
The first term comes from the training observations the two fitted models share, and the second from each unit training the other's model. At independently drawn training sets the covariance is \(C_\ell=(\sigma_T^2+2\gamma_0)/n+o(n^{-1})\), and Theorem~\ref{thm:variance_identity} reads
\begin{equation}
	\label{eq:pairwise_form}
	\Var(\widehat R_D)
	=\sigma_\ell^2\,\Gamma_D+C_\ell
	+\sigma_T^2\Big(\Lambda_D-\frac1n\Big)+2\gamma_0\Big(\Omega_D-\frac1n\Big)
	+o(n^{-1})
\end{equation}
(Supplement~\ref{appendix:proof_pairwise_form}). For \(K\)-fold the last two terms are zero, and the sample variance of held-out contributions divided by \(n\) estimates \(\sigma_\ell^2/n\) and misses \(C_\ell\), the training terms that refitting adds. The within-fold and between-fold covariances \(\omega\) and \(\gamma^{\mathrm{BG}}\) of \citet{bengio2004} are \eqref{eq:pair_cov} at the two overlap patterns of \(K\)-fold, and only \(q\,\omega+p\,\gamma^{\mathrm{BG}}=C_\ell\) enters the variance at first order (Supplement~\ref{appendix:proof_pairwise_form}). For the AUC under class-stratified \(K\)-fold, the fixed-model variance is the first term of Theorem~\ref{thm:pairwise_identity}(b), \(\sigma_{V\mid1}^2/n_1+\sigma_{V\mid0}^2/n_0\), the within-fold analogue of the variance of \citet{delong1988}, and by Corollary~\ref{prop:pairwise_reduction} the cross-validation variance exceeds it at first order by the training terms \((\sigma_T^2+2\gamma_A)/n\), which can have either sign because \(\gamma_A\) can be negative; given the class counts, as DeLong's estimator is computed, the label part of \(\sigma_T^2\) drops out of this difference.

\section{Balanced Designs and Design Comparison}
\label{sec:implications}

Sections~\ref{subsec:design_comparison} to~\ref{subsec:loo} are stated for unit-level metrics. Their first-order results hold for pairwise metrics at the constants of Corollary~\ref{prop:pairwise_reduction}, and the balance results of Section~\ref{subsec:design_comparisons_fixed_p} hold for class-stratified designs (Supplement~\ref{appendix:proof_order_balance} and~\ref{appendix:proof_pairwise_second}). The fold-count comparison of Section~\ref{subsec:loo} does not carry over, because \(K\)-fold evaluates only the case--control pairs that share a fold, so Section~\ref{subsec:folds_auc} treats pairwise metrics separately, with leave-pair-out \citep{yousef2005,airola2011} in the role of leave-one-out.

\subsection{The first-order imbalance penalty}
\label{subsec:design_comparison}

Across designs, imbalance in role weights can add split noise at every order. At first order the weights are \(v_i\) and \(w_i\), and balanced weights attain the minimum. When every observation has \(v_i=w_i=1/n\), the first-order component of Theorem~\ref{thm:dual_role} is \(\sum_iH(Z_i)/n\), with \(H(Z_i):=\ell_i^\star-R_{p,n}+\bar\phi(Z_i)\) the combined influence of an observation, the sum of its two influences. Its variance is \(\mathcal V/n\), with
\[
\mathcal V:=\Var\{H(Z)\}=\sigma_V^2+\sigma_T^2+2\gamma_0.
\]
For any other design, write each weight as \(1/n\) plus a deviation. Since \(\sum_iv_i=\sum_iw_i=1\), the deviations sum to zero, the cross terms between \(1/n\) and the deviations drop out, and \eqref{eq:cond_var_roles} becomes
\begin{equation}
	\label{eq:imbalance_decomposition}
	\Var(\widehat R_D\mid I)
	=
	\frac{\mathcal V}{n}+\sum_{i=1}^n\Big(v_i-\frac1n,\,w_i-\frac1n\Big)\,\Sigma\,\Big(v_i-\frac1n,\,w_i-\frac1n\Big)^\top+\rho_n^{(2+)}(I),
\end{equation}
with \(\Sigma\) the covariance matrix of \((\ell_i^\star,\bar\phi(Z_i))\), exactly at every realized set of splits. The second term is the \emph{first-order imbalance penalty}: extra variance from unequal role weights. It is nonnegative because \(\Sigma\) is a covariance matrix and vanishes for a balanced design. Averaged over the splits it is the imbalance penalty of the design,
\[
Q_D
:=
\sigma_V^2\Big(\Gamma_D-\frac1n\Big)+\sigma_T^2\Big(\Lambda_D-\frac1n\Big)+2\gamma_0\Big(\Omega_D-\frac1n\Big),
\]
so that \(\Var(\widehat R_D)=\mathcal V/n+Q_D+\mathbb E_I\{\rho_n^{(2+)}(I)\}\). The leading term of every fixed-size design is therefore at least \(\mathcal V/n\), and a balanced design attains it. For a pairwise metric, class-stratified \(K\)-fold is balanced in the sense of Section~\ref{subsec:weighted_representation}, and \(K\)-fold independent of the labels has a penalty below first order, because its unit-level weights in Corollary~\ref{prop:pairwise_reduction} are all \(1/n\).

The penalty of a design follows from the moments of its weights. Under Monte Carlo cross-validation \(N_i\sim\mathrm{Binomial}(S,q)\), and the bootstrap is taken at \(p=1-e^{-1}\). The binomial and multinomial moments of the weights of Section~\ref{subsec:weighted_representation} give (Supplement~\ref{appendix:proof_role_moments})
\begin{equation}
	\label{eq:role_moments}
	\begin{aligned}
		K\text{-fold:}\quad & \Gamma_D=\Lambda_D=\Omega_D=\frac1n,\\
		\text{Monte Carlo:}\quad & \Gamma_D=\frac1n\Big(1+\frac{p}{qS}\Big),\quad \Lambda_D=\frac1n\Big(1+\frac{q}{pS}\Big),\quad \Omega_D=\frac1n\Big(1-\frac1S\Big),\\
		\text{bootstrap:}\quad & \Gamma_D=\frac1n\Big(1+\frac{p}{qS}\Big),\quad \Lambda_D=\frac1n\Big(1+\frac1S\Big),\quad \Omega_D=\frac1n\Big(1-\frac1S\Big),
	\end{aligned}
\end{equation}
the first two exactly and the third up to \(o(n^{-1})\). Substituting into \(Q_D\) gives, under A1 and, for the bootstrap, A3,
\begin{equation}
	\label{eq:QD_designs}
	Q_D=
	\begin{cases}
		0, & K\text{-fold, any number of repetitions},\\[4pt]
		\dfrac{1}{nS}\Big(\dfrac{p}{q}\,\sigma_V^2+\dfrac{q}{p}\,\sigma_T^2-2\gamma_0\Big), & \text{Monte Carlo},\\[10pt]
		\dfrac{1}{nS}\Big(\dfrac{p}{q}\,\sigma_V^2+\sigma_T^2-2\gamma_0\Big)+o(n^{-1}), & \text{bootstrap}.
	\end{cases}
\end{equation}
Monte Carlo and the bootstrap validate each observation a random number of times, and their penalties fall at \(1/S\). Under Monte Carlo cross-validation the two weights are also tied, since an observation validated more often trains less often: \(w_i-1/n=-(q/p)(v_i-1/n)\), so its penalty is \(p/(qnS)\) times \(\Var\{\ell_i^\star-(q/p)\bar\phi(Z_i)\}\). At a fixed number of splits the penalty is of order \(1/n\), the order of the variance itself.

\subsection{Sampling variance and split noise}
\label{subsec:split_noise}

To derive the identity, we fixed the splits and varied the independent observations. To compare designs, we reverse that conditioning: at a common training size, designs differ only in the added variance caused by their split assignments. By the law of total variance conditional on the data,
\begin{equation}
	\label{eq:lotv_data}
	\Var(\widehat R_D)
	=
	\Var_{\mathcal D}\!\left\{
	\mathbb E_I(\widehat R_D\mid\mathcal D_n)
	\right\}
	+
	\mathbb E_{\mathcal D}\!\left\{
	\Var_I(\widehat R_D\mid\mathcal D_n)
	\right\}.
\end{equation}
The first term, the \emph{sampling variance}, is the variance from the random draw of the data: the variance across datasets of the estimate averaged over all split assignments. It is the variance the estimate would have after unlimited repetition, and no amount of repetition brings the variance below it. The second term, the \emph{split noise}, is the variance across split assignments at a fixed dataset, averaged over datasets. It is the part of the variance that comes from the random splitting of the data.

The separation holds at every order: averaging over splits changes the weights but not the kernels. By condition (iii) of Section~\ref{subsec:cv_design} the weights of each order are exchangeable, so the averages over \(I\) of the first-order weights are \(1/n\), and for a fixed-size design the averages of the weights of order \(k\) depend only on \(n\), \(m\), and \(k\). Because \(I\) is independent of the data, the Hoeffding components of \(\mathbb E_I(\widehat R_D\mid\mathcal D_n)\) are those of \(\widehat R_D\) with every weight replaced by its average over \(I\) (Supplement~\ref{appendix:proof_split_lotv}). The sampling variance is therefore the variance identity evaluated at the averaged weights, and the split noise is the rest of each term. At first order the rest is \(Q_D\). Hence
\begin{equation}
	\label{eq:floor_split}
	\Var_{\mathcal D}\{\mathbb E_I(\widehat R_D\mid\mathcal D_n)\}
	=
	\frac{\mathcal V}{n}+\sum_{k\ge2}\Pi_n^{(k)},
	\qquad
	\mathbb E_{\mathcal D}\{\Var_I(\widehat R_D\mid\mathcal D_n)\}
	=
	Q_D+\sum_{k\ge2}\mathcal N_D^{(k)},
\end{equation}
where \(\Pi_n^{(k)}\), written \(\Pi_n\) at \(k=2\), is the term of order \(k\) of the identity at the averaged weights, and \(\mathcal N_D^{(k)}:=\mathbb E_I\{\rho_n^{(k)}(I)\}-\Pi_n^{(k)}\) is the variance of order \(k\) from the weights of order \(k\) changing from one split draw to the next.

The sampling variance depends on the design only through the sample and training sizes, so every design at a given training size has the same sampling variance, written \(\Var_{\mathrm{samp}}(m)\). For a pairwise metric, given the class counts, it depends on the design only through the training composition \((m_1,m_0)\), and averaging over the class counts adds the variance of the target \(R_{p,n}(Y)\) across them (Supplement~\ref{appendix:proof_split_lotv_auc}). Because split noise is nonnegative, sampling variance is the lower bound for every design at that training size.

The split noise is where designs differ. For a fixed-size design with \(S\) splits, the estimate is the average of the split estimates, \(\widehat R_D=S^{-1}\sum_r\widehat R_r\), and
\begin{equation}
	\label{eq:split_noise_S}
	\mathbb E_{\mathcal D}\!\left\{\Var_I(\widehat R_D\mid\mathcal D_n)\right\}
	=
	\frac1S\,\mathbb E_{\mathcal D}\!\left\{\Var_I(\widehat R_1\mid\mathcal D_n)\right\}
	+
	\frac1{S^2}\sum_{r\ne r'}\mathbb E_{\mathcal D}\!\left\{\Cov_I(\widehat R_r,\widehat R_{r'}\mid\mathcal D_n)\right\}.
\end{equation}
Each split by itself is a uniformly drawn training set of size \(m\), so the first term is the same for every design at that training size. Designs differ only through the covariances between their splits. Monte Carlo splits are drawn independently, so the covariances are zero, and every part of the split noise falls at \(1/S\). The folds of a \(K\)-fold partition are dependent. Every observation is validated in exactly one fold and trains in the others, so the covariance sum cancels the first-order part of the first term for any number of repetitions, and \(Q_D=0\). What remains is \(\sum_{k\ge2}\mathcal N_D^{(k)}\), the split noise left after averaging over the folds.

Repetition reduces this remaining split noise and does not change the sampling variance. Write \(\widehat R_{K,R}\) for the average of \(R\) independently repeated \(K\)-fold estimates and \(\mathcal N_{K,1}\) for the split noise of one partition. Given the data the \(R\) estimates are independent and identically distributed, so \eqref{eq:lotv_data} gives
\begin{equation}
	\label{eq:repeated_floor}
	\Var(\widehat R_{K,R})=\Var_{\mathrm{samp}}(m)+\frac{\mathcal N_{K,1}}{R}
\end{equation}
exactly, under A1 alone.

Two established results follow from these decompositions. \citet{burman1989} expands the variance of repeated \(K\)-fold cross-validation for a bias-corrected version of the estimate, and finds the number of repetitions only in a term of order \(n^{-2}\). By \eqref{eq:repeated_floor} the number of repetitions divides only the split noise of one partition. Under A2(i) that split noise is \(o(n^{-1})\), because \(Q_D=0\) for \(K\)-fold (Step~4 of Supplement~\ref{appendix:proof_split_lotv}), and under A2(ii) with convergent pair constants it is of order \(n^{-2}\) (Section~\ref{subsec:loo}). When the order-two term is of first order, so is the split noise of one partition, which repetition then reduces.

\citet{nadeau2003} estimate the variance of Monte Carlo cross-validation by \(\hat\sigma^2(1/S+q/p)\), with \(\hat\sigma^2\) the sample variance of the \(S\) split estimates at the fixed dataset, on the argument that two split estimates are correlated by about \(q\). In the identity's terms \(\hat\sigma^2\) estimates the split noise of a single split, whose first-order part is \(Q_D\) at \(S=1\), so their estimator targets
\begin{equation}
	\label{eq:nadeau_bengio}
	\frac{\mathcal V}{n} + Q_D-\frac{(2p-1)\sigma_T^2+2p\gamma_0}{p^2n}
\end{equation}
up to \(o(n^{-1})\) under A2(i) (Supplement~\ref{appendix:proof_nadeau_bengio}). The first two terms are the variance of Monte Carlo cross-validation. The correction is therefore exact at first order if and only if \(2p\gamma_0=(q-p)\sigma_T^2\), which is the condition for two split estimates to have correlation \(q\). When the model is fit by minimizing the reported loss, both sides go to zero with the training terms, so the condition holds in the limit (Section~\ref{subsec:refitting}). Otherwise the error can have either sign.

\subsection{Optimal designs at a fixed training size}
\label{subsec:design_comparisons_fixed_p}

The sampling variance sets a lower bound; we now ask which designs attain it, and through which orders. At each order the variance given the splits is a quadratic form in the weights of that order, \eqref{eq:cond_var_roles} and \eqref{eq:cond_var_order}, and the sampling variance is the same form at the weights averaged over the splits, so variation of the weights around their average adds variance. A design is \emph{balanced at order \(k\)} if its weights of order \(k\) do not vary with \(I\); at order one this is balance in the sense of Section~\ref{subsec:weighted_representation}, since a weight that does not vary equals its average \(1/n\). Balance through order \(k\) removes the split noise through order \(k\) and leaves the parts above it.

Balance at order one does not imply balance at order two. Under \(K\)-fold the pair weights \(c_{ij}\) and \(d_{ij}\) depend on whether the two observations share a fold, in which case they are validated and trained together, and otherwise one is validated while the other trains. Which pairs share a fold changes from one partition to the next, and that variation is the second-order split noise of \(K\)-fold. For repeated \(K\)-fold the pair weights depend only on how many repetitions put the two in the same fold, so balance at order two, \emph{pair balance}, means that every pair shares a fold equally often.

\begin{theorem}[Optimality of balanced designs]
	\label{thm:balanced_optimal}
	Under A1, let \(1\le k\le m+1\). A fixed-size design balanced at order \(k\) is balanced at every lower order and the parts of its split noise through order \(k\) are zero, so that its variance is the first expression below exactly, while every fixed-size design at that training size satisfies the second:
	\[
	\Var(\widehat R_D)=\frac{\mathcal V}{n}+\sum_{l=2}^{k}\Pi_n^{(l)}+\mathbb E_I\{\rho_n^{((k+1)+)}(I)\},
	\qquad
	\Var(\widehat R_D)\ge\frac{\mathcal V}{n}+\sum_{l=2}^{k}\Pi_n^{(l)}+Q_D+\sum_{l=2}^{k}\mathcal N_D^{(l)},
	\]
	where \(\mathcal N_D^{(l)}\) is zero only if the components of order \(l\) of \(\widehat R_D\) do not vary with \(I\). Under A2 at order \(k\), a fixed-size design is therefore variance-optimal at its training size up to \(o(n^{-k})\) if and only if \(Q_D+\sum_{l=2}^{k}\mathcal N_D^{(l)}=o(n^{-k})\), and a design balanced at order \(k\) is.
\end{theorem}
The proof is in Supplement~\ref{appendix:proof_order_balance}. Balance through order \(k\) ensures optimality through that order. Conversely, when the kernels of an order are linearly independent, eliminating its split noise requires balance at that order (Theorem~\ref{thm:dual_role}). Asymptotic optimality needs only the condition of the theorem, which a design with small imbalance also meets. A design that is not balanced at order one exceeds the optimum by \(Q_D\) at first order.

\begin{corollary}[First-order optimality of \(K\)-fold cross-validation]
	\label{cor:kfold_optimal}
	Under A1 and A2(i), \(K\)-fold cross-validation with any number of repetitions has variance \(\mathcal V/n+o(n^{-1})\) and is first-order variance-optimal among fixed-size designs at training fraction \(p=(K-1)/K\).
\end{corollary}
This is Theorem~\ref{thm:balanced_optimal} at \(k=1\) together with \eqref{eq:QD_designs}. Balance at order two is not achieved by random partitions and must be constructed. Independent repetitions of \(K\)-fold divide its second-order split noise by \(R\) but do not remove it, because the numbers \(\lambda_{ij}\) of partitions in which \(i\) and \(j\) share a fold still differ across pairs. Choosing the partitions so that every pair shares a fold equally often removes it exactly, whatever its size, and by Theorem~\ref{thm:balanced_optimal}, under A1 alone, the variance of pair-balanced \(K\)-fold then equals the sampling variance through order two. For a fixed set of \(R\) partitions with the observations in random order, let \(\bar\lambda:=R(n/K-1)/(n-1)\) be the average of the \(\lambda_{ij}\). The second-order split noise is then exactly
\begin{equation}
	\label{eq:split_noise_counts}
	\mathcal N_D^{(2)}
	=
	\frac1{R^2}\Big\{\frac{\tau_n^2+\kappa_n}{n^2m^2}-\frac{2\xi_n}{nKm^3}+\frac{\sigma_{\chi,n}^2}{2K^2m^4}\Big\}\sum_{i\ne j}(\lambda_{ij}-\bar\lambda)^2
\end{equation}
(Supplement~\ref{appendix:proof_order_balance}), which ranks any set of partitions by how unevenly its pairs share folds. Pair-balanced partitions make it zero, and exact pair balance needs at least \(K(n-1)/(K-1)\) fits, more than leave-one-out (Supplement~\ref{appendix:proof_order_balance}). When the sampling variance does not rise with the training size, leave-one-out also has the smaller variance (Section~\ref{subsec:loo}), so pair balance is generally not worth its compute.

\subsection{Fold counts at first order}
\label{subsec:endpoint_variance_optimality}

Corollary~\ref{cor:kfold_optimal} compares designs at one training fraction, not different fold counts. Changing \(K\) changes the training size and thus both \(R_{p,n}\) and \(\mathcal V_{pn}\). The model used in practice is fit to all \(n\) observations, while cross-validation trains on \(m<n\), so when the model improves with more data, cross-validation is conservative for the deployed model. A larger \(K\) trains closer to the deployed model, and leave-one-out trains closest. Bias is therefore smallest at the largest \(K\), and the choice of \(K\) is a question of variance.

Throughout, \(K\ge2\) and \(R\) are fixed as \(n\to\infty\). The validation and training influences, and with them \(\sigma_V^2\), \(\sigma_T^2\), and \(\gamma_0\), depend on the training size (a dependence suppressed so far). Write \(H_m\) and \(\mathcal V_m\) for \(H\) and \(\mathcal V\) at training size \(m\), so that \(\mathcal V=\mathcal V_{pn}\). Comparing fold counts needs these influences to converge as the training size grows, as A4(ii) requires for pairwise metrics.

\smallskip\noindent\textbf{A5 (Limit).} \emph{As \(m\to\infty\) with \(m/n\to p\), the centered validation influence \(\ell_i^\star-R_{p,n}\) and the training influence \(\bar\phi\) converge in \(L_2(P)\). Write \(\mathcal V_\infty\) for the limit of \(\mathcal V_m\).}

\smallskip
The common claim that more folds increase variance because training sets overlap is misleading. Overlap makes held-out contributions covary through \(\sigma_T^2\) and \(\gamma_0\) \eqref{eq:pair_cov}, but for \(K\)-fold these covariances total \((\sigma_T^2+2\gamma_0)/n\) at first order for every \(K\), because each observation has training weight \(1/n\). Thus \(K\) affects first-order variance only through training size, and not at all when \(\mathcal V_m\) has a common limit.

\begin{corollary}[\(K\)-invariance at first order]
	\label{cor:choice_of_K}
	Under A1, A2(i) and A5, if \(\mathcal V_\infty\) is common across the training fractions compared, then
	\[
	\Var(\widehat R_{K,R})=\frac{\mathcal V_\infty}{n}+o(n^{-1}).
	\]
\end{corollary}
This follows from Corollary~\ref{cor:kfold_optimal}, since \(\mathcal V=\mathcal V_{pn}\to\mathcal V_\infty\), and it is the \(K\)-independence of the limiting variance in \citet{austern2020}. The condition holds whenever each fit depends on its training set alone, including tuning by an inner cross-validation on that set, since then \(\mathcal V_m\) depends only on \(m\) and every training fraction has the same limit. It fails when the fit also depends on the full sample size. For example, a ridge penalty set to \(cn\) and applied to a training set of size \(pn\) penalizes at \(c/p\) per training observation, so the limit of the fitted model, and with it \(\mathcal V_\infty\), changes with \(K\).

\subsection{Leave-one-out and a fixed number of fits}
\label{subsec:loo}

Leave-one-out uses every holdout of size one, so it has no split noise and its variance is \(\Var_{\mathrm{samp}}(n-1)\). It is the limit of the \(K\)-fold family. When the sampling variance does not rise with the training size, no \(K\)-fold design has a smaller variance. Further, when also the covariance between the two kinds of pair interaction is not positive, a fixed number of fits is best spent on one run with as many folds as they allow, not on repetitions (Corollary~\ref{cor:loo_limit}, below). At first order leave-one-out and \(K\)-fold have the same variance, by Corollary~\ref{cor:choice_of_K}.

Let the pair constants converge to common limits \(\tau^2\), \(\kappa\), \(\sigma_\chi^2\), \(\xi\) at the training sizes compared, and write \(G(K)/n^2\) for the limiting second-order split noise of one \(K\)-fold partition, where
\[
G(K):=\frac{\tau^2+\kappa}{K-1}-\frac{2\xi}{(K-1)^2}+\frac{\sigma_\chi^2}{2(K-1)^3}.
\]
Under A1, and A2(i) and (ii) for \(K\)-fold and for leave-one-out, the variance of repeated \(K\)-fold through second order is (Supplement~\ref{app:endpoint_variance_optimality})
\begin{equation}
	\label{eq:K_second_corrected}
	\Var(\widehat R_{K,R})
	=\frac{\mathcal V_{pn}}{n}+\frac{\tau^2+\kappa+\tfrac12\sigma_\chi^2+2\xi}{n^2}+\frac{G(K)}{Rn^2}+o(n^{-2}),
\end{equation}
and hence
\begin{equation}
	\label{eq:loo_endpoint_variance_gap}
	\Var(\widehat R_{K,R})-\Var(\widehat R_{\mathrm{LOO}})
	=\frac{\mathcal V_{pn}-\mathcal V_{n-1}}{n}+\frac{G(K)}{Rn^2}+o(n^{-2}),
\end{equation}
whatever the rate at which \(\mathcal V_m\) converges. The first two terms of \eqref{eq:K_second_corrected} are the sampling variance. The first depends on \(K\) only through the training size, and the second, from the pair constants, is the same at every \(K\). Repetition divides only the third term of \eqref{eq:K_second_corrected}, the split noise, so repeated \(K\)-fold converges to its own sampling variance, not that of leave-one-out.

If the model learns so that \(\mathcal V_m=\mathcal V_\infty+bm^{-\gamma}+o(m^{-\gamma})\) for some \(0<\gamma\le1\), then
\begin{equation}
	\label{eq:rate_gap}
	\frac{\mathcal V_{pn}-\mathcal V_{n-1}}{n}=\frac{b\,(p^{-\gamma}-1)}{n^{1+\gamma}}+o(n^{-1-\gamma}).
\end{equation}
At \(\gamma=1\) this is \(b/\{(K-1)n^2\}+o(n^{-2})\), of the order of the split noise, and at \(\gamma<1\) it is of larger order and determines the sign of the difference.

With \(S=RK\) fits, the split noise in \eqref{eq:K_second_corrected} is \(KG(K)/(Sn^2)\), so a fixed number of fits compares designs by \(KG(K)\), the split noise per fit. Its first and third terms are nonnegative (\(|\kappa|\le\tau^2\) by Cauchy--Schwarz) and decrease in \(K\), so more folds lower it if \(\xi<0\).

Neither sign is guaranteed mathematically, although both conditions arise in typical model settings. The sign of \(b\) is that of the covariance between the limiting and finite-training parts of an observation's influence, when \(H_m\) itself expands at that rate (Supplement~\ref{app:sign_training_correction}): \(b>0\) when the observations that move the estimate most move it more with less training data. For the sample mean under squared error or the Brier score an observation far from the mean has a large loss when validated and pulls the fitted mean away from the population mean when it trains, so \(b=2\Var\{(Y-\mathbb E(Y))^2\}\ge0\), or \(2\theta(1-\theta)(1-2\theta)^2\) for a binary outcome with mean \(\theta\). For the sample mean under squared error, \(\xi=-4\sigma^4<0\): a training observation lowers the loss of a similarly deviating validation observation, whereas two such training observations jointly increase expected loss. This opposition need not hold for other learners; convexity alone does not determine the sign. Examples of both reversals appear in Supplements~\ref{app:sign_training_correction} and~\ref{app:sign_xi}. Although these sign reversals are mathematically possible, a negative \(b\) corresponds to the counterintuitive situation in which increasing the training size increases the sampling variance of estimated risk, contrary to the behavior ordinarily expected of a well-behaved learner. Thus, under conventional learning conditions, leave-one-out is generally the minimum-variance endpoint of the \(K\)-fold family, rather than the high-variance extreme it is often argued to be.

\begin{corollary}[Leave-one-out as the limit of \(K\)-fold cross-validation]
	\label{cor:loo_limit}
	Under A1, A2(i) and (ii) for \(K\)-fold and for leave-one-out, and convergence of the pair constants to common limits, suppose that \(\mathcal V_m\) does not increase in \(m\).
	\begin{itemize}
		\item[(a)] Up to \(o(n^{-2})\), the variance of repeated \(K\)-fold cross-validation is at least that of leave-one-out, the case \(K=n\), for every \(K\) and \(R\).
		\item[(b)] If also \(\xi\le0\), then up to \(o(n^{-2})\) the variance of repeated \(K\)-fold cross-validation does not increase in \(K\), and at a fixed number \(S=RK\) of fits one run of \(S\)-fold cross-validation has the smallest variance among these designs.
	\end{itemize}
\end{corollary}
The proof is in Supplement~\ref{app:loo_corollaries}. It uses only that the smaller training fraction has the larger sampling variance for all large \(n\), so \(b>0\) suffices, and part (a) holds whatever the pair constants, because the split noise is nonnegative and goes to zero as \(K\) grows. Leave-one-out is also the smallest among fixed-size designs whose training sizes include \(n-1\), exhaustive designs included, exactly when \(\Var_{\mathrm{samp}}(n-1)\) is the smallest sampling variance among them (Supplement~\ref{app:loo_corollaries}).

\subsection{Pairwise metrics: cross-validation on case--control pairs}
\label{subsec:folds_auc}

Ordinary class-stratified \(K\)-fold evaluates only case--control pairs in the same validation fold: \(n_1n_0/K\) of the \(n_1n_0\) pairs. Thus each partition averages the pair-specific interaction \(\zeta^{(2)}\) over a different \(1/K\) share of pairs. The resulting split noise grows linearly in \(K\) and, scaling as \(1/(\pi_1\pi_0)\), is larger for rare outcomes. Unit-level metrics have no counterpart, since each partition evaluates every observation. More folds therefore raise the training size and lower pair coverage.

Leave-pair-out \citep{yousef2005,airola2011} evaluates every case--control pair once, eliminating split noise. For a class-stratified design, the first-order sampling variance follows from Theorem~\ref{thm:pairwise_identity}(b) at the averaged weights \(\Gamma_{D,y}=1/n_y\), \(\Lambda_{D,y}=m_y^2/(n_ym^2)\), and \(\Omega_{D,y}=m_y/(n_ym)\). Averaging over class counts also adds the variance of the target:
\begin{equation}
	\label{eq:auc_first_floor}
	\mathcal W_{1,n}
	=
	\mathbb E_Y\bigg\{\sum_{y\in\{0,1\}}\frac1{n_y}\Big(\sigma_{V\mid y}^2+\frac{m_y^2}{m^2}\,\sigma_{T\mid y}^2+\frac{2m_y}{m}\,\gamma_{0\mid y}\Big)\bigg\}
	+
	\Var\{R_{p,n}(Y)\},
\end{equation}
with the constants of Theorem~\ref{thm:pairwise_identity} at training composition \(m_1\) (Supplement~\ref{appendix:proof_split_lotv_auc}). Write \(\mathcal W_{1,n}(K)\) and \(\mathcal W_{1,n}(\mathrm{LPO})\) for its values at \(m_y=pn_y\) and \(m_y=n_y-1\). Under A1, A4, A2(ii) given the labels and \(L_2\) convergence of the pair kernels (Supplement~\ref{appendix:proof_pairwise_second}), class-stratified \(K\)-fold with \(R\) independent repetitions has
\begin{equation}
	\label{eq:lpo_endpoint_variance_gap}
	\Var(\widehat R_{K,R})-\Var(\widehat R_{\mathrm{LPO}})
	=\mathcal W_{1,n}(K)-\mathcal W_{1,n}(\mathrm{LPO})+\frac{G_A(K)}{Rn^2}+o(n^{-2}),
\end{equation}
where \(G_A(K)\ge0\) has the terms of \(G(K)\) of Section~\ref{subsec:loo} at the pairwise analogues of the pair constants, covariances between \(\zeta^{(2)}\) and the training effects, and the \emph{pair-coverage cost} \(\lambda_4(K-1)\), \(\lambda_4:=\varsigma^2/(\pi_1\pi_0)\) with \(\varsigma^2\) the limiting variance of \(\zeta^{(2)}\): a case--control pair has validation weight \(K/(n_1n_0)\) when its members share a fold and zero otherwise, and the variance this adds, summed over the \(n_1n_0\) pairs, is \((K-1)\varsigma^2/(n_1n_0)=\lambda_4(K-1)/n^2+o(n^{-2})\). Because this term increases in \(K\), Corollary~\ref{cor:loo_limit}(b) does not carry over and an intermediate fold count can be best, while Corollary~\ref{cor:loo_limit}(a) carries over to leave-pair-out whenever its first-order sampling variance is lower (Supplement~\ref{app:endpoint_auc}).

\emph{Paired \(K\)-fold cross-validation} \citep{yousef2019} divides each class into \(K\) folds and fits all \(K^2\) combinations, holding out one case fold and one control fold per fit. Every case--control pair is then evaluated once. Its training composition and averaged weights agree with class-stratified \(K\)-fold, so sampling variance is unchanged, while the pair-coverage noise, including \(\lambda_4(K-1)\) and the other case--control terms in \(G_A(K)\), is removed.With \(R\) independent repetitions (Supplement~\ref{appendix:proof_pairwise_second})
\begin{equation}
	\label{eq:paired_gap}
	\Var\big(\widehat R_{K,R}^{\mathrm{paired}}\big)-\Var(\widehat R_{\mathrm{LPO}})
	=\mathcal W_{1,n}(K)-\mathcal W_{1,n}(\mathrm{LPO})+\frac{G_s(K)}{Rn^2}+o(n^{-2}),
\end{equation}
where \(G_s\) is \(G\) at the same-class parts of \(\tau_A^2+\kappa_A\), \(\sigma_{\chi,A}^2\) and \(\xi_A\). Paired \(K\)-fold is therefore never worse than class-stratified \(K\)-fold at the same \(R\), and at the same number of fits it has the smaller variance for all large enough \(K\) whenever \(\lambda_4>0\), since \(G_A(K)/K\to\lambda_4\) while \(G_s(K)\to0\). Its split noise does not increase in \(K\) when the same-class part of \(\xi_A\) is not positive, and its variance does not when also \(\mathcal W_{1,n}(K)\) does not. With class-specific fold counts, leave-pair-out, whose folds are single cases and single controls, is its limit, as leave-one-out is of \(K\)-fold.

\section{Asymptotic Normality}
\label{sec:normality}\label{subsec:clt}
The exact identity captures every order of variance. Under the conditions of Theorem~\ref{thm:clt}, the estimate is asymptotically normal with the full variance given the splits. A2(i) states that the first-order term dominates the variance. Theorem~\ref{thm:clt}(a) does not require it; it requires that \(n\) times the conditional variance converge and that single observations move the estimate little, so it covers models whose higher-order terms are of the same order as the first (Supplement~\ref{appendix:clt_local} gives an example). Under A2(i) the limiting variance is \(\mathcal V_\infty\), the first-order variance of a balanced design, plus the imbalance penalty. Except in the degenerate case in which the first-order influence has zero variance, so that the variance of the estimate is \(o(1/n)\), the limiting variance is positive, and normality with a consistent estimate of it yields Wald intervals and tests, and it makes the design comparisons of Section~\ref{sec:implications}, which compare variances, comparisons of limiting laws.

Earlier central limit theorems differ in their target. \citet{austern2020} center at \(R_{p,n}\), a fixed target that is the same for every design at a given training size, and prove their theorem for \(K\)-fold cross-validation under stability conditions. \citet{Bayle2020} and \citet{lei2025cv} center at a random target, the average expected performance of the \(K\) models fitted in the run. Theorem~\ref{thm:clt} extends the limit at a fixed target to Monte Carlo and bootstrap cross-validation, to pairwise metrics, and to models for which A2(i) fails.

\subsection{A normal approximation for functions of independent observations}

Given the splits, the estimate depends on \(n\) independent observations but is not a sum of independent terms. Sum-based CLTs need higher-order effects to vanish, as in A2(i). Theorem~\ref{thm:clt}(a) instead uses the normal approximation of \citet{chatterjee2008} for functions of independent inputs, allowing higher-order terms to contribute to variance. Recall from Section~\ref{subsec:assumptions} that \(\nabla_jg\) is the change in \(g\) when \(Z_j\) is replaced by an independent copy, and \(\nabla_j\nabla_kg\) the change in \(\nabla_kg\) when \(Z_j\) is replaced as well. For a pairwise metric, conditioning on \(I\) below means conditioning on \((I,Y)\), and the independent inputs are the covariates given their labels.

\begin{theorem}[Asymptotic normality]
	\label{thm:clt}
	Assume A1.
	\begin{itemize}
		\item[(a)] If \(n\Var(\widehat R_D\mid I)\to V\) in probability, then
		\[
		\sqrt n\,\big\{\widehat R_D-\mathbb E(\widehat R_D\mid I)\big\}\Rightarrow N(0,V),
		\]
		given \(I\) and unconditionally, whenever some function \(\widetilde R_D\) of the observations with \(n\Var(\widehat R_D-\widetilde R_D\mid I)\to0\) satisfies, in probability,
		\begin{gather*}
			\sum_{j}\mathbb E\big\{(\nabla_j\widetilde R_D)^4\,\big|\,I\big\}=o(n^{-2}),\qquad
			\sum_{j}\mathbb E\big\{|\nabla_j\widetilde R_D|^3\,\big|\,I\big\}=o(n^{-3/2}),\\
			\sum_{k}\mathbb E\Big[\Big\{\sum_{j\ne k}|\nabla_j\nabla_k\widetilde R_D|\,|\nabla_j\widetilde R_D|\Big\}^2\,\Big|\,I\Big]=o(n^{-2}),
		\end{gather*}
		where each difference may be evaluated at any point whose \(j\)th coordinate is \(Z_j\) or one of its independent copies, for every \(j\), and the rates hold uniformly over these choices. For a unit-level metric, \(\sqrt n(\widehat R_D-R_{p,n})\Rightarrow N(0,V)\), under A3 with \(\mathbb E(\widehat R_D)\) in place of \(R_{p,n}\) for the bootstrap.
		\item[(b)] For balanced fixed-size designs, and for Monte Carlo and bootstrap cross-validation with a fixed number of splits, the conditions of (a) hold under A2(i) and A5, and A3 for the bootstrap, with \(V=\mathcal V_\infty+\lim_nnQ_D\).
	\end{itemize}
\end{theorem}

The proof is in Supplement~\ref{appendix:proof_clt}. The limit holds given the splits, which is the position in practice, and unconditionally, and for a unit-level metric and a fixed-size design \(\mathbb E(\widehat R_D\mid I)=R_{p,n}\) for every realized \(I\), so it is centered at the fixed target. The conditions of part (a) say that no single observation moves the estimate by a share of its standard deviation that does not vanish, and that the effect of one observation depends on the others weakly, or strongly on only a few; they may be checked on a surrogate \(\widetilde R_D\) because a part of the estimate with variance \(o(1/n)\) changes neither the limit nor \(V\), which is how part (b) needs no moment condition beyond A1 and the \(L_2\) convergence in A5.

For a balanced design the imbalance penalty is zero and \(V=\mathcal V_\infty\), for any number of repetitions and, when each fit depends only on its own training set, for any number of folds, leave-one-out included. For Monte Carlo and bootstrap cross-validation, \(V\) exceeds \(\mathcal V_\infty\) by the imbalance penalty of the realized splits, which remains in the limiting variance given those splits, so an interval for \(R_{p,n}\) from one run must use \(V/n\), penalty included. An interval at \(\mathcal V_\infty/n\) falls short of its nominal coverage whenever the penalty is positive, and one at the validation term \(\sigma_V^2/n\) alone omits, in addition, the training terms \(\sigma_T^2+2\gamma_0\), which can have either sign. Without A2(i), an interval at the identity's full variance given the splits is asymptotically valid by part (a), while one at its first-order part omits the higher orders' share.

\subsection{Pairwise metrics}

Pairwise metrics have the same normal-limit form, but a different target. For \(K\)-fold and Monte Carlo cross-validation with a fixed number of splits, independent of the labels, and for stratified \(K\)-fold under the condition of Corollary~\ref{prop:pairwise_reduction}, under A4
\[
\sqrt n\big\{\widehat R_D-\mu_m(\pi_1)\big\}\Rightarrow N\big(0,\ \sigma_A^2+\sigma_T^2+2\gamma_A+\lim_nnQ_D\big),
\]
where \(Q_D\) is the imbalance penalty at the constants of Corollary~\ref{prop:pairwise_reduction} (Supplement~\ref{appendix:proof_clt}). Only the constants and the target differ. The target is \(\mu_m(\pi_1)\), the expected metric of a model fitted to \(m\) observations at the population class proportion, because the conditional mean \(\mathbb E(\widehat R_D\mid I,Y)\) varies with the labels through the class compositions of the training sets, unlike its unit-level counterpart. That variation is asymptotically normal and independent of the limit given the labels, and its variance is the label part of \(\sigma_T^2\) (Section~\ref{subsec:pairwise_metrics}) times \(\lim_nn\Lambda_D\), which is one for \(K\)-fold and larger for Monte Carlo, whose training use is uneven. For \(K\)-fold the penalty is zero and the limiting variance is \(\sigma_A^2+\sigma_T^2+2\gamma_A\).

\subsection{Relation to earlier central limit theorems}

\citet{austern2020} prove, for \(K\)-fold with a number of folds that may grow with \(n\), that \(\sqrt n(\widehat R_{K,1}-R_{p,n})\) is asymptotically normal with variance \(\sigma_{1,\mathrm{AZ}}^2+\sigma_{2,\mathrm{AZ}}^2+2\rho_{\mathrm{AZ}}\), under \(L_4\) stability conditions on the first and second differences of a contribution and of the expected performance of the fitted model. Their \(\rho_{\mathrm{AZ}}\) is \(\lim\gamma_0\) exactly, and at a fixed number of folds their conditions imply A2(i), \(\sigma_{2,\mathrm{AZ}}^2=\lim\sigma_T^2\), and \(\sigma_{1,\mathrm{AZ}}^2=\lim\sigma_V^2\) (Supplement~\ref{appendix:proof_containment}). Their limiting variance is therefore \(\mathcal V_\infty\) term by term, and Theorem~\ref{thm:clt} obtains their conclusion with A2(i) and A5 in place of their stability and convergence conditions. Their Proposition~3 with its Remark~6, that \(\rho_{\mathrm{AZ}}=0\) when a parametric M-estimator is evaluated with its own fitting loss, is the parametric case of the matched-loss argument of Section~\ref{subsec:refitting}.

\citet{Bayle2020}, centering at the random target, need only the perturb-one stability of the centered contribution, and their limiting variance is \(\sigma_V^2\), the validation term. \citet{lei2025cv} gives the same randomly centered theorem under perturb-one stability of the uncentered contribution, together with a condition under which the two centerings coincide. In the identity's terms they coincide when \(\sigma_T^2=0\), which forces \(\gamma_0=0\). The two limiting variances agree under the weaker condition \(\sigma_T^2+2\gamma_0=0\), and otherwise differ in either direction.

\subsection{What existing variance estimators estimate}
\label{subsec:existing_estimators}

Theorem~\ref{thm:clt} supports Wald inference with a consistent estimator of \(V\). The identity identifies what it must contain: at first order the validation term, the training terms \(\sigma_T^2+2\gamma_0\) and, for a design that is not balanced, the imbalance penalty; beyond first order, the higher-order terms whenever they carry variance of order \(1/n\). Each estimator in use captures part of this.

The standard variance estimate, the sample variance of held-out losses divided by \(n\), estimates the validation term and misses \(C_\ell=(\sigma_T^2+2\gamma_0)/n\), the covariance that held-out contributions acquire through the training observations their models share \eqref{eq:pairwise_form}. The interval of \citet{Bayle2020} uses it, which is valid for their random target but not for \(R_{p,n}\) unless the training terms are negligible. \citet{bates2023} show that intervals from it undercover and trace the shortfall to the within-fold and between-fold covariances of \citet{bengio2004}, the \(\omega\) and \(\gamma^{\mathrm{BG}}\) of Section~\ref{subsec:refitting}; their nested cross-validation estimates the error about a random target, the performance of the model fitted to the data at hand, at the cost of many more fits. For the AUC, DeLong's variance and the influence-curve variance of \citet{ledell2015}, which holds each fold's fitted model fixed, both estimate the validation term \(\sigma_A^2/n\) and under stratified \(K\)-fold miss the training terms \((\sigma_T^2+2\gamma_A)/n\) of Corollary~\ref{prop:pairwise_reduction}.

Repeating cross-validation does not estimate \(V\) either. Given the data, the spread of \(R\) repeated \(K\)-fold estimates estimates the split noise \(\mathcal N_{K,1}\) of one partition \eqref{eq:repeated_floor}, which under A2(i) is \(o(n^{-1})\), so as a standard error it is asymptotically negligible relative to the total variance. The correction of \citet{nadeau2003} for Monte Carlo splits is exact at first order only when \(2p\gamma_0=(q-p)\sigma_T^2\) (Section~\ref{subsec:split_noise}).

\citet{cai2025} target \(R_{p,n}\) itself. Their first-order expansion of the Monte Carlo estimate averaged over many splits has the validation influence and the limiting training influence of Section~\ref{subsec:refitting} as its two terms, so its variance is \(\mathcal V_\infty\), and they estimate it by a weighted bootstrap of the whole procedure with a random-effects fit that removes the split noise within each replicate. The result is the first-order sampling variance, training terms included. It omits the higher-order terms that Theorem~\ref{thm:clt}(a) keeps in \(V\), since it rests on a first-order expansion, and the penalty \(Q_D\) of a run with few Monte Carlo splits, since it removes split noise. Training on reweighted observations reproduces the deployed procedure's training influence only under the comparability that A3(iii) asks of bootstrap cross-validation, and its own Monte Carlo error takes thousands of fits to make small.

No unbiased estimate of the variance of \(K\)-fold exists from the held-out contributions of one run \citep{bengio2004}, so a consistent estimator needs structure beyond independence. Stability conditions or a random target set the training terms aside, and resampling the observations keeps them but changes the training sets the model sees. A third approach is to draw replicate datasets from an estimated population and rerun the procedure with its realized splits. This reproduces all variance components, including refitting and imbalance, to the extent that the estimated population is accurate.

\section{Empirical Verification}
\label{sec:empirical_validation}

We compare predicted variance components with their empirical values across replicated datasets. Every constant is computed from its definition, not fitted to the results. The first-order and pair terms of Theorems~\ref{thm:variance_identity} and~\ref{thm:pairwise_identity} are checked against the total variance, and the split of Section~\ref{subsec:split_noise} against the split noise.

\begin{figure}[t]
	\centering
	\includegraphics[width=\textwidth]{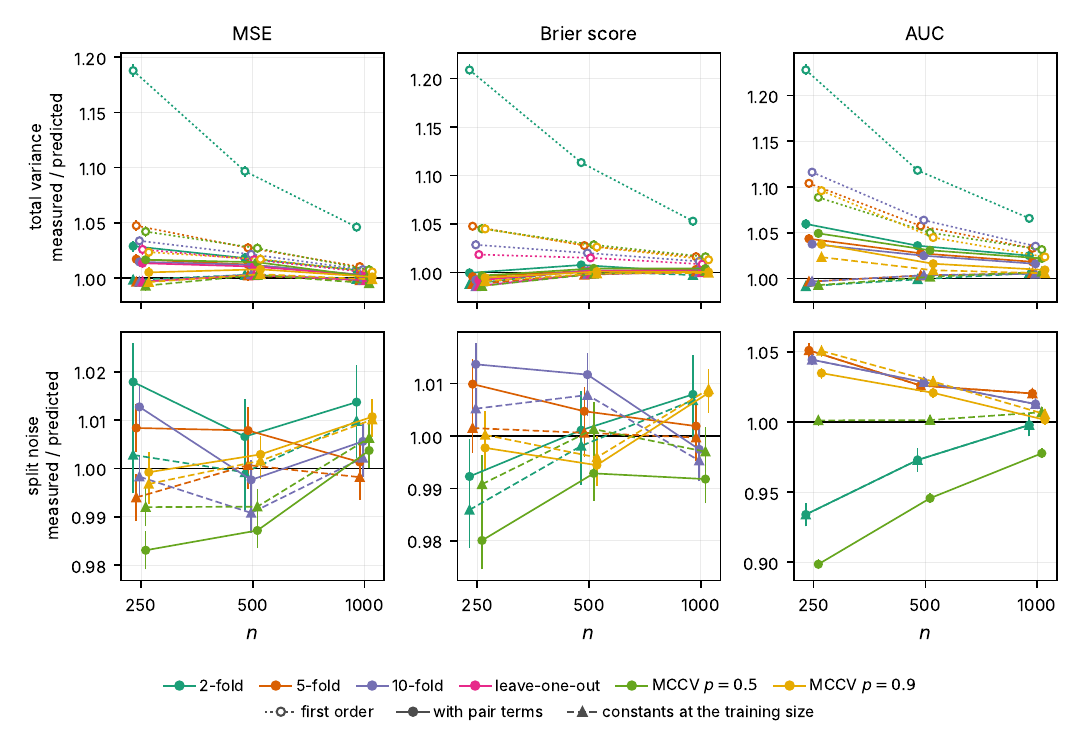}
	\caption{Measured variance against the identity's predictions, for three metrics, six designs, and three sample sizes. Error bars are one standard error. Top row: total variance over $10^5$ datasets, as a ratio to the prediction at first order (hollow markers, dotted) and to the prediction with the limiting pair terms added (filled markers, solid). Bottom row: split noise over $6000$ datasets with $30$ redraws, as a ratio to the prediction with the pair terms. Triangles: ratio to the prediction with the constants at the training size, through order three for MSE and the Brier score and at first order for the AUC. Leave-one-out has no split noise.}
	\label{fig:identity}
\end{figure}

The simulation fits a fixed-bandwidth Nadaraya--Watson smoother, which is in the influence-function case of Section~\ref{subsec:assumptions}, for three metrics, three sample sizes, and several designs. The population is $X\sim\mathrm{Uniform}(-1,1)^2$ with $\eta(X)=1.5X_1+2\sin(3X_2)-0.5$. The continuous outcome $Y=\eta(X)+\epsilon$, $\epsilon\sim N(0,1)$, is evaluated by squared error (MSE). The binary outcome $Y\sim\mathrm{Bernoulli}\{\mathrm{expit}\,\eta(X)\}$ is evaluated by the Brier score and by the within-fold AUC with class-stratified folds. The Gaussian kernel has bandwidth $h=0.7$, held fixed. Its population limit is smoothed relative to the conditional mean, which allows nonzero first-order training and cross terms (Section~\ref{subsec:refitting}).

The limiting constants are evaluated by quadrature over the covariates and exact integration of the outcome moments, and their finite-training corrections by differentiating the expected-performance expansion of Supplement~\ref{appendix:validation_finite_m}. The simulation evaluates the AUC with the indicator; its constants are computed with the indicator smoothed to $\Phi(u/\varepsilon)$ and extrapolated to $\varepsilon=0$ (Supplement~\ref{appendix:validation_finite_m}, which also gives $b_A$). The limit of this classifier ranks covariates by the ratio of its kernel averages over cases and over controls, which does not depend on the training class fraction, so the label part of $\sigma_T^2$ is zero. Supplement~\ref{appendix:validation_finite_m} gives the constants. The designs are $K$-fold with $K\in\{2,5,10\}$, leave-one-out for the unit-level metrics, which holds out no case--control pair, and Monte Carlo cross-validation with $(p,S)=(0.5,10)$ or $(0.9,10)$, at $n\in\{250,500,1000\}$. Figure~\ref{fig:identity} shows the comparisons and Supplement~\ref{appendix:validation_more} the numbers.

The identity's predictions agree with the measured variances. Across $51$ cells the total-variance ratios lie between $0.992$ and $1.060$, and across $45$ nonzero cells the split-noise ratios lie between $0.899$ and $1.051$, with relative standard errors of about $0.5\%$ and $0.4$--$0.9\%$. The pair terms are needed: at first order alone the discrepancies reach $23\%$ for total variance and $73\%$ for split noise. With the finite-training correction $b$ added as well, the total-variance ratios range from $0.994$ to $1.004$ for MSE and from $0.986$ to $1.002$ for the Brier score, and with its AUC analogue $b_A$ (Supplement~\ref{appendix:validation_finite_m}) from $0.992$ to $1.023$ for the AUC. The remaining differences are shared by the designs at each $n$, which use the same datasets. The terms the predictions omit, of order $n^{-3}$, move them by less than $1\%$ at these sample sizes, except the finite-training corrections to the pair constants of the AUC, which are not computed and leave its $K$-fold split noise up to $7\%$ from the prediction at $n=250$ (Supplement~\ref{appendix:validation_finite_m}). Refitting matters: for the Brier score, the term $\sigma_T^2 + 2\gamma_0$ omitted by a fixed-model formula accounts for $45\%$ of first-order variance (Section~\ref{subsec:refitting}). The measured variances order the fold counts as Section~\ref{sec:implications} predicts. At $n=250$ the totals for both MSE and the Brier score fall from two-fold through five- and ten-fold to leave-one-out (Corollary~\ref{cor:loo_limit}), and for the AUC five-fold is below ten-fold at each sample size (Section~\ref{subsec:folds_auc}).

Additional checks compare the identity with the exact sample-mean variance (Supplement~\ref{appendix:validation_mean}), test the effect of pair balance (Supplement~\ref{appendix:validation_pair_balanced}), and verify Theorem~\ref{thm:clt}(a) when A2(i) fails (Supplements~\ref{appendix:validation_soclt} and~\ref{appendix:clt_local}).

\section{Conclusion}
\label{sec:conclusion}

Cross-validation is typically understood through how the data are partitioned and how much the resulting training sets overlap. Neither alone determines the variance of the performance estimate. The relevant structure is how each observation contributes through its own evaluation and through the models it helps fit. Expressing these contributions as design-dependent weights yields an exact finite-sample variance identity and separates the uncertainty due to sampling the dataset from the additional variation introduced by assigning observations to folds. This distinction provides a common basis for understanding and comparing resampling procedures that have traditionally been studied separately.

The implications for choosing a cross-validation design are substantial. At a common training size, designs share the same sampling variance and differ only in their split noise. Ordinary \(K\)-fold is balanced at first order, so increasing the overlap between training sets does not inflate its leading variance, which under the conditions developed here is independent of \(K\). Moreover, leave-one-out eliminates split noise entirely and has the smallest variance in the \(K\)-fold family whenever sampling variance does not increase with training size. Thus, the commonly held view that leave-one-out suffers from high variance because its training sets overlap is not generally justified. The preferred fold count still depends on the learner and how sampling variance changes with training size, but under the additional second-order conditions established here, increasing the number of folds is also a more efficient use of a fixed fitting budget than repeating partitions with fewer folds.

The appropriate design also depends on the performance metric. For metrics averaged over individual observations, \(K\)-fold evaluates every observation once per partition. For pairwise metrics such as the AUC, however, ordinary \(K\)-fold evaluates only case--control pairs assigned to the same fold, introducing a source of split noise that increases with the number of folds. Paired \(K\)-fold removes this pair-coverage cost by evaluating every case--control pair. This distinction illustrates why balancing observations alone is not always sufficient: the design must also account for the units over which predictive performance is evaluated.

Beyond design selection, the decomposition has implications for statistical inference. Repeating cross-validation on a fixed dataset measures variation due to the splitting, not the sampling uncertainty of the performance estimate. Conversely, variance estimates based only on held-out contributions can omit the uncertainty arising from refitting the model. The exact identity, together with the normal approximation developed here, provides a basis for inference that accounts for both roles and, when necessary, higher-order interactions. An important direction for future research is the development of computationally practical estimators of these variance components for complex learning procedures, as well as extensions to dependent observations and metrics whose evaluation units are determined by the observed data.

Taken together, these results shift the focus of cross-validation design from the overlap of training sets to the distribution of statistical influence across observations and their interactions. Understanding that structure is essential not only for choosing how to cross-validate, but also for quantifying what can be learned from the resulting performance estimate.

	\bibliographystyle{plainnat}
	\bibliography{cv-optimality}
	
	\clearpage
	\appendix
	\renewcommand{\thesection}{S\arabic{section}}
	\let\paragraph\origparagraph 
	\section{Aggregation and Sufficient Conditions for A2}
\label{appendix:assumptions_preliminaries}

This section extends the estimate to general aggregation (Section~\ref{appendix:alt_cv_methods}) and gives sufficient conditions for A2 (Section~\ref{appendix:stability_assumptions}). Throughout, $\mathcal{D}_n = \{Z_i\}_{i=1}^n$ is i.i.d.\ from $P$, a model is fit to a training set $T$ to produce the fitted model $\hat f_T$, and $\ell_i(f)$ is the contribution of unit $i$ evaluated at the model $f$. For fixed-size designs every training set has size $m$ with $m/n \to p$. A1 and A2 are as stated in Section~\ref{subsec:assumptions}, A3 in Section~\ref{appendix:proof_bootstrap_identity}, A4 in Section~\ref{subsec:pairwise_metrics}, and A5 in Section~\ref{subsec:endpoint_variance_optimality}.

\subsection{General aggregation and variants of the estimate}
\label{appendix:alt_cv_methods}

The appendices allow any aggregation of the evaluations that is determined by the inclusion structure. Let $a_{ir} \ge 0$ be the weight on unit $i$'s contribution in split $r$, with $a_{ir} = 0$ unless $i \in V_r$ and $\sum_{r,i}a_{ir} = 1$, so that
\[
\widehat R_D = \sum_r\sum_{i \in V_r}a_{ir}\ell_i^{(r)},
\]
and let $s_r := \sum_ia_{ir}$ be the share of the estimate produced by split $r$. The role weights are
\[
v_i := \sum_ra_{ir},
\qquad
w_i := \sum_rs_r\frac{I_{ir}}{|T_r|},
\]
with $\tilde I_{ir}/n$ in place of $I_{ir}/|T_r|$ for the bootstrap, and $\ell_i := v_i^{-1}\sum_ra_{ir}\ell_i^{(r)}$. For a fixed-size design the weights of order $k$ are
\[
c^{(k)}_{i;\mathcal J}:=\frac{1}{m^{k-1}}\sum_{r=1}^S a_{ir}\prod_{j\in\mathcal J}I_{jr},
\qquad
d^{(k)}_{\mathcal S}:=\frac{1}{m^{k}}\sum_{r=1}^S s_r\prod_{j\in\mathcal S}I_{jr},
\]
for a set $\mathcal J$ of $k-1$ observations other than $i$ and a set $\mathcal S$ of $k$ observations. At $k=1$ they are $v_i$ and $w_i$, and at $k=2$ they are the pair weights $c_{ij}$ and $d_{ij}$. Pooled aggregation, the case of the main text, is $a_{ir} = \mathbb{I}(i \in V_r)/M$ with $M = \sum_r|V_r|$, so that $s_r = |V_r|/M$ and the weights reduce to those of Sections~\ref{subsec:weighted_representation} and~\ref{subsec:representation}. The representation and the identities are proved for general $a_{ir}$. Theorem~\ref{thm:variance_identity} applies exactly whenever the design is fixed-size and the coefficients are determined by the inclusion structure and sum to one. Changing the aggregation changes only the role weights. Averaging split-level estimates, $a_{ir}=\mathbb I(i\in V_r)/(S|V_r|)$, coincides with pooling for fixed-size designs.

\paragraph{The .632 and .632+ estimators.}
Both mix the out-of-bag estimate with the apparent error of the full-data fit. The apparent error evaluates every observation on a model it helped fit, which breaks condition (i), and each of its contributions depends on \(Z_i\) through the prediction as well as the outcome, a term Theorem~\ref{thm:dual_role} does not contain, so the mixture is outside Theorem~\ref{thm:variance_identity}.

\subsection{Sufficient conditions for A2}
\label{appendix:stability_assumptions}

This proves the statements of Section~\ref{subsec:assumptions} that bounds on finite differences of the contributions give A2 at every order and that the influence-function case satisfies A2(i). With \(\nabla_j\) as in Section~\ref{subsec:assumptions}, for a set \(\mathcal J\) of observations let \(\nabla_{\mathcal J}\) apply \(\nabla_j\) for every \(j\in\mathcal J\), and let \(\bar\ell^{(r)}:=s_r^{-1}\sum_{b\in V_r}a_{br}\ell_b^{(r)}\) be the average contribution of split \(r\). For \(k\ge1\) let
\[
\beta_k':=\sup\big\|\nabla_{\mathcal J}\ell_i^{(r)}\big\|_2,
\qquad
\beta_{k+1}:=\sup\big\|\nabla_{\mathcal S}\bar\ell^{(r)}\big\|_2,
\]
the suprema over splits \(r\), validated units \(i\in V_r\), and sets \(\mathcal J\) of \(k\) and \(\mathcal S\) of \(k+1\) distinct members of \(T_r\). The first measures how much \(k\) training observations jointly move one held-out contribution, the second how much \(k+1\) of them jointly move the average contribution of a split. With \(Z_i\) fixed, write \(\varrho_{ir}\) for the part of \(\ell_i^{(r)}\) that depends jointly on two or more of its training observations.

The following holds under A1 for a fixed-size design with \(\max_iv_i\le c/n\), with \(h_{\mathcal S}\) the components of \(\widehat R_D\) given \(I\).
\begin{itemize}
	\item[(a)] For \(k\ge1\) and every realized \(I\),
	\[
	\rho_n^{((k+1)+)}(I)\le\frac{k+2}{2^{k+1}}\Big\{\frac{4m^k\max_iv_i}{k!}\,\beta_k'^2+\frac{m^{k+1}}{(k+1)!}\,\beta_{k+1}^2\Big\}.
	\]
	Hence A2 at order \(k\) holds when \(\beta_k'=o(n^{-(2k-1)/2})\) and \(\beta_{k+1}=o(n^{-(2k+1)/2})\). When \(\beta_k'=O(m^{-k})\) and \(\beta_{k+1}=O(m^{-(k+1)})\), \(\rho_n^{((k+1)+)}(I)\) is \(O(n^{-(k+1)})\).
	\item[(b)] A2(i) holds for every realized \(I\) when
	\[
	\text{(S1)}\quad \mathbb E\{\phi_i(Z)^2\}=o(n),
	\qquad
	\text{(S2)}\quad \sup_{i,r}\mathbb E(\varrho_{ir}^2)=o(n^{-1}).
	\]
\end{itemize}

\begin{proof}[Proof of (a)]
	\noindent\emph{Step 1: Replacing the members of a set isolates the components that contain it.}\quad
	Let \(\mathcal S\) be a set of \(k+1\) observations and \(\mathcal S'\) any set. Then
	\[
	\nabla_{\mathcal S}h_{\mathcal S'}=\sum_{\mathcal A\subseteq\mathcal S}(-1)^{|\mathcal A|}h_{\mathcal S'}^{(\mathcal A)},
	\]
	where the superscript marks which members of \(\mathcal S\) are replaced by independent copies. If \(\mathcal S\not\subseteq\mathcal S'\), some member of \(\mathcal S\) is not an argument of \(h_{\mathcal S'}\), and \(\nabla_{\mathcal S}h_{\mathcal S'}=0\). If \(\mathcal S\subseteq\mathcal S'\), any two of the \(2^{k+1}\) terms differ in a replaced observation, in which \(h_{\mathcal S'}\) has conditional mean zero given its other arguments. The terms are therefore uncorrelated, and
	\[
	\|\nabla_{\mathcal S}h_{\mathcal S'}\|_2^2=2^{k+1}\|h_{\mathcal S'}\|_2^2.
	\]
	
	\noindent\emph{Step 2: The remainder is bounded by a sum of differences.}\quad
	The operator \(\nabla_{\mathcal S}\) removes \(\mathbb E(\widehat R_D\mid I)\), and components on different sets stay orthogonal after the replacements, since for \(\mathcal S'\ne\mathcal S''\) one of the two depends on an observation, original or copy, that the other does not. Summing Step~1 over the sets of size \(k+1\),
	\[
	\sum_{|\mathcal S|=k+1}\|\nabla_{\mathcal S}\widehat R_D\|_2^2
	=
	2^{k+1}\sum_{\mathcal S'}\binom{|\mathcal S'|}{k+1}\|h_{\mathcal S'}\|_2^2
	\ge
	2^{k+1}\rho_n^{((k+1)+)}(I).
	\]
	
	\noindent\emph{Step 3: One difference in terms of the stability constants.}\quad
	Apply \(\nabla_{\mathcal S}\) to \(\widehat R_D=\sum_r\sum_{b\in V_r}a_{br}\ell_b^{(r)}\). The contribution \(\ell_b^{(r)}\) depends only on \(Z_b\) and \(Z_{T_r}\), so a term survives only if \(\mathcal S\subseteq\{b\}\cup T_r\), in one of two ways.
	\begin{itemize}
		\item[(1)] \(b\in\mathcal S\) and \(\mathcal S\setminus\{b\}\subseteq T_r\). Then \(\nabla_{\mathcal S}\ell_b^{(r)}\) is the difference of two values of \(\nabla_{\mathcal S\setminus\{b\}}\ell_b^{(r)}\), at \(Z_b\) and at its copy, so
		\[
		\|\nabla_{\mathcal S}\ell_b^{(r)}\|_2\le2\beta_k'.
		\]
		\item[(2)] \(\mathcal S\subseteq T_r\). Then the terms over \(b\in V_r\) sum to \(s_r\nabla_{\mathcal S}\bar\ell^{(r)}\), and
		\[
		\Big\|\sum_{b\in V_r}a_{br}\nabla_{\mathcal S}\ell_b^{(r)}\Big\|_2\le s_r\beta_{k+1}.
		\]
	\end{itemize}
	Summing over splits by Minkowski's inequality, with \(|T_r|=m\),
	\[
	\|\nabla_{\mathcal S}\widehat R_D\|_2
	\le
	2\beta_k'\,m^k\sum_{i\in\mathcal S}c^{(k+1)}_{i;\mathcal S\setminus\{i\}}
	+
	\beta_{k+1}\,m^{k+1}d^{(k+1)}_{\mathcal S}.
	\]
	
	\noindent\emph{Step 4: The bound in the weights.}\quad
	The right side of Step~3 has \(k+2\) terms, and \((\sum_{l=1}^{k+2}x_l)^2\le(k+2)\sum_lx_l^2\). With Step~2,
	\[
	\rho_n^{((k+1)+)}(I)
	\le
	\frac{k+2}{2^{k+1}}\Big\{4\beta_k'^2m^{2k}\sum_{|\mathcal S|=k+1}\sum_{i\in\mathcal S}\big(c^{(k+1)}_{i;\mathcal S\setminus\{i\}}\big)^2
	+\beta_{k+1}^2m^{2k+2}\sum_{|\mathcal S|=k+1}\big(d^{(k+1)}_{\mathcal S}\big)^2\Big\}.
	\]
	
	\noindent\emph{Step 5: The weight sums.}\quad
	Each \(c^{(k+1)}_{i;\mathcal J}\) is at most \(v_i/m^k\), and a training set contains \(\binom mk\le m^k/k!\) sets \(\mathcal J\) of size \(k\), so
	\[
	\sum_{\mathcal J}c^{(k+1)}_{i;\mathcal J}\le\frac{v_i}{k!},
	\qquad
	\sum_{|\mathcal S|=k+1}\sum_{i\in\mathcal S}\big(c^{(k+1)}_{i;\mathcal S\setminus\{i\}}\big)^2
	\le
	\sum_i\frac{v_i^2}{k!\,m^k}
	\le
	\frac{\max_iv_i}{k!\,m^k},
	\]
	using \(\sum_iv_i^2\le\max_iv_i\). In the same way \(d^{(k+1)}_{\mathcal S}\le1/m^{k+1}\) and \(\sum_{\mathcal S}d^{(k+1)}_{\mathcal S}\le1/(k+1)!\), so
	\[
	\sum_{|\mathcal S|=k+1}\big(d^{(k+1)}_{\mathcal S}\big)^2\le\frac{1}{(k+1)!\,m^{k+1}}.
	\]
	Substituting into Step~4 gives the bound of (a).
	
	\noindent\emph{Step 6: The rates.}\quad
	With \(\max_iv_i\le c/n\), \(m\le n\) and \(m/n\to p>0\), the bound is at most a constant times
	\[
	n^{k-1}\beta_k'^2+n^{k+1}\beta_{k+1}^2,
	\]
	which is \(o(n^{-k})\) when \(\beta_k'=o(n^{-(2k-1)/2})\) and \(\beta_{k+1}=o(n^{-(2k+1)/2})\), and \(O(n^{-(k+1)})\) when \(\beta_k'=O(m^{-k})\) and \(\beta_{k+1}=O(m^{-(k+1)})\).
\end{proof}

The condition \(\max_iv_i\le c/n\) holds for \(K\)-fold, including leave-one-out, for Monte Carlo cross-validation with \(p<1\), where \(v_i\le1/(n-m)\), and for exhaustive leave-\((n-m)\)-out.

\begin{proof}[Proof of (b)]
	\noindent\emph{Step 1: The projection of a contribution.}\quad
	At fixed \(Z_i\), the Hoeffding decomposition of \(\ell_i^{(r)}\) in its training observations is
	\begin{equation}
		\label{eq:app_if_expansion}
		\ell_i^{(r)} = \ell_i^\star + \frac{1}{m}\sum_{j \in T_r}\phi_i(Z_j) + \varrho_{ir},
	\end{equation}
	since \(\mathbb E(\ell_i^{(r)}\mid Z_i,Z_j)-\mathbb E(\ell_i^{(r)}\mid Z_i)=\phi_i(Z_j)/m\). Write \(\phi_i^\circ:=\phi_i-\bar\phi\). Substituting into \(\widehat R_D\), and using \(\sum_ic_{ij}=w_j\),
	\[
	\widehat R_D-R_{p,n}
	=
	\sum_i h_i(Z_i)
	+
	U_D+\bar\varrho_D,
	\qquad
	U_D:=\sum_{i\ne j}c_{ij}\phi_i^\circ(Z_j),
	\qquad
	\bar\varrho_D:=\sum_r\sum_{i\in V_r}a_{ir}\varrho_{ir}.
	\]
	Neither \(U_D\) nor \(\bar\varrho_D\) has a component on a single observation, so
	\[
	\rho_n^{(2+)}(I)
	=
	\Var(U_D+\bar\varrho_D\mid I)
	\le
	2\Var(U_D\mid I)+2\Var(\bar\varrho_D\mid I).
	\]
	
	\noindent\emph{Step 2: The pair term.}\quad
	The function \(\phi_i^\circ(Z_j)\) has conditional mean zero given either \(Z_i\) or \(Z_j\), so two terms of \(U_D\) are uncorrelated unless they involve the same unordered pair. With
	\[
	\tau_n^2=\mathbb E\{\phi_i^\circ(Z)^2\}\le\mathbb E\{\phi_i(Z)^2\}=o(n)
	\]
	by (S1),
	\[
	\Var(U_D\mid I)\le2\tau_n^2\sum_{i\ne j}c_{ij}^2.
	\]
	Since \(c_{ij}\le v_i/m\) and \(\sum_jc_{ij}\le v_i\),
	\[
	\sum_{i\ne j}c_{ij}^2\le\frac{\max_iv_i}{m}=O(n^{-2}),
	\]
	so \(\Var(U_D\mid I)=o(n^{-1})\).
	
	\noindent\emph{Step 3: The remainder.}\quad
	By Minkowski's inequality and (S2),
	\[
	\Var(\bar\varrho_D\mid I)
	\le
	\Big(\sum_r\sum_{i\in V_r}a_{ir}\Big)^2
	\sup_{i,r}\mathbb E(\varrho_{ir}^2)
	=
	o(n^{-1}).
	\]
	With Step~1, \(\rho_n^{(2+)}(I)=o(n^{-1})\) for every realized \(I\).
\end{proof}

\paragraph{The influence-function case.}
Suppose the fitted model converges to a limit \(f_p^\star\) and has an influence function \(\psi\), and the induced contribution satisfies the mean-square expansion
\[
\ell_i(\hat f_T)=L\{Y_i,f_p^\star(X_i)\}
+\frac1{m}\sum_{j\in T}\partial_2L\{Y_i,f_p^\star(X_i)\}\psi(Z_j;X_i)+\varrho_{iT},
\]
with \(\sup_{i,T}\mathbb E(\varrho_{iT}^2)=o(n^{-1})\) and \(\psi\) centered, and with
\[
\mathbb E\{\psi(Z;X)^2\}<\infty,
\qquad
\mathbb E\big[\partial_2L\{Y,f_p^\star(X)\}\psi(Z;X)\big]^2\ \text{bounded}.
\]
These are conditions on the contribution, since an \(L_2\) expansion of the fitted prediction alone does not control the quadratic Taylor remainder in \(L_2\). Under them (S1) and (S2) hold, so A2(i) holds, and A5 holds with the limits in closed form.

\begin{proof}
	\noindent\emph{Step 1: The training influence.}\quad
	Write \(\ell_i^{\mathrm{lin}}(Z_j):=\partial_2L\{Y_i,f_p^\star(X_i)\}\psi(Z_j;X_i)\). Since \(\psi\) is centered,
	\[
	\phi_i(z)=\ell_i^{\mathrm{lin}}(z)+m\big\{\mathbb E(\varrho_{iT}\mid Z_i,Z_j=z)-\mathbb E(\varrho_{iT}\mid Z_i)\big\}.
	\]
	At fixed \(Z_i\) the projections of \(\varrho_{iT}\) onto the \(m\) single training observations are orthogonal and, by symmetry, have equal norms, so
	\[
	m\,\big\|\mathbb E(\varrho_{iT}\mid Z_i,Z_j)-\mathbb E(\varrho_{iT}\mid Z_i)\big\|_2\le\big\{m\,\mathbb E(\varrho_{iT}^2)\big\}^{1/2}=o(1),
	\]
	and \(\phi_i=\ell_i^{\mathrm{lin}}+o_{L_2}(1)\).
	
	\noindent\emph{Step 2: The validation influence.}\quad
	Averaging over the training observations,
	\[
	\ell_i^\star=L\{Y_i,f_p^\star(X_i)\}+\mathbb E(\varrho_{iT}\mid Z_i),
	\qquad
	\|\mathbb E(\varrho_{iT}\mid Z_i)\|_2\le\|\varrho_{iT}\|_2=o(n^{-1/2}).
	\]
	
	\noindent\emph{Step 3: Conditions (S1) and (S2).}\quad
	By Step~1 and the bounded second moment of \(\ell_i^{\mathrm{lin}}\), \(\mathbb E\{\phi_i(Z)^2\}=O(1)\), which is (S1). The linear term of the expansion is a sum of functions of one training observation each, so the components of \(\ell_i^{(r)}\) on two or more training observations are those of \(\varrho_{iT}\), and
	\[
	\mathbb E(\varrho_{ir}^2)\le\mathbb E(\varrho_{iT}^2)=o(n^{-1}),
	\]
	which is (S2).
	
	\noindent\emph{Step 4: The closed-form limits.}\quad
	Averaging Step~1 over \(Z_i\), by Jensen's inequality \(\bar\phi\) converges in \(L_2(P)\) to
	\[
	\bar\phi_\infty(z)=\mathbb E\!\left[\partial_2L\{Y,f_p^\star(X)\}\psi(z;X)\right],
	\]
	and by Step~2 \(\ell_i^\star\) converges to \(L\{Y_i,f_p^\star(X_i)\}\). So \(\sigma_V^2\), \(\sigma_T^2\), and \(\gamma_0\) converge to the variances and covariance of these two limits, which is A5 with the limiting covariance in closed form.
\end{proof}

\section{Proofs for Section~\ref{sec:identity}}
\label{appendix:proofs_main_results}

Throughout, \(a_{ir}\), \(s_r\), \(v_i\), \(w_i\), and the weights of order \(k\) are those of general aggregation (Section~\ref{appendix:alt_cv_methods}). The main text is the pooled case \(a_{ir}=\mathbb I(i\in V_r)/M\), \(s_r=|V_r|/M\).

\subsection{Proof of Theorems~\ref{thm:dual_role} and~\ref{thm:variance_identity}}
\label{appendix:IF_derivation}
\label{appendix:proof_variance_identity}

The design is fixed-size, so every training set has \(m\) observations, and by A1 every contribution is a symmetric function of the training observations of its split.

\begin{proof}
	\noindent\emph{Step 1: The components and their support.}\quad
	With the components of A2, taken given \(I\), each component has conditional mean zero given all but one of its arguments, components on different sets are orthogonal, and \(g\mapsto g_{\mathcal S}\) is linear. Let \(g\) be a function of \(Z_{\mathcal U}\) alone and \(u\in\mathcal S\setminus\mathcal U\). For every \(\mathcal A\subseteq\mathcal S\setminus\{u\}\), \(Z_u\) is independent of \((g,Z_{\mathcal A})\), so
	\[
	\mathbb E(g\mid Z_{\mathcal A\cup\{u\}})=\mathbb E(g\mid Z_{\mathcal A}).
	\]
	The terms for \(\mathcal A\) and \(\mathcal A\cup\{u\}\) in \(g_{\mathcal S}\) have opposite signs and cancel, so
	\begin{equation}
		\label{eq:support_rule}
		g_{\mathcal S}=0\quad\text{unless }\mathcal S\subseteq\mathcal U.
	\end{equation}
	The components of \(\widehat R_D\) given \(I\) are \(h_{\mathcal S}=(\widehat R_D)_{\mathcal S}\), and by linearity
	\[
	h_{\mathcal S}=\sum_{r=1}^S\sum_{b\in V_r}a_{br}\big(\ell_b^{(r)}\big)_{\mathcal S}.
	\]
	
	\noindent\emph{Step 2: Which terms contribute.}\quad
	The contribution \(\ell_b^{(r)}\) is a function of \(Z_b\) and \(Z_{T_r}\). By \eqref{eq:support_rule}, \((\ell_b^{(r)})_{\mathcal S}\ne0\) requires \(\mathcal S\subseteq\{b\}\cup T_r\). Since \(b\in V_r\) and \(V_r\cap T_r=\varnothing\), either \(b\in\mathcal S\) and \(\mathcal S\setminus\{b\}\subseteq T_r\), or \(b\notin\mathcal S\) and \(\mathcal S\subseteq T_r\). A split in which two members of \(\mathcal S\) are validated contributes nothing to \(h_{\mathcal S}\).
	
	\noindent\emph{Step 3: A member of \(\mathcal S\) is validated.}\quad
	Let \(b=i\in\mathcal S\) and \(\mathcal J:=\mathcal S\setminus\{i\}\subseteq T_r\), with \(|\mathcal J|=k-1\). The component \((\ell_i^{(r)})_{\{i\}\cup\mathcal J}\) is a function of \((Z_i,Z_{\mathcal J})\). The observations are i.i.d., the law of the fit is unchanged by permuting its training observations, and every training set has size \(m\), so this function is the same for every split and every choice of \(\mathcal J\subseteq T_r\), is symmetric in \(Z_{\mathcal J}\), and equals \(m^{-(k-1)}\phi^{(k)}(Z_i;Z_{\mathcal J})\). Summing over the splits in which \(i\) is validated and \(\mathcal J\) trains,
	\[
	\sum_{r}a_{ir}\prod_{j\in\mathcal J}I_{jr}\;m^{-(k-1)}\phi^{(k)}(Z_i;Z_{\mathcal J})
	=
	c^{(k)}_{i;\mathcal S\setminus\{i\}}\,\phi^{(k)}(Z_i;Z_{\mathcal S\setminus\{i\}}).
	\]
	
	\noindent\emph{Step 4: All of \(\mathcal S\) trains.}\quad
	Let \(b\notin\mathcal S\) and \(\mathcal S\subseteq T_r\). Since \(Z_b\) is independent of \(Z_{T_r}\),
	\[
	\mathbb E(\ell_b^{(r)}\mid Z_{T_r})=\mathbb E_{Z_b}\{\ell_b(\hat f_{T_r})\}=R(\hat f_{T_r})=:R_r,
	\]
	and by the tower property \(\mathbb E(\ell_b^{(r)}\mid Z_{\mathcal A})=\mathbb E(R_r\mid Z_{\mathcal A})\) for every \(\mathcal A\subseteq\mathcal S\). Hence \((\ell_b^{(r)})_{\mathcal S}=(R_r)_{\mathcal S}\), which does not depend on \(b\), and by the symmetry of the fit equals \(m^{-k}\chi^{(k)}(Z_{\mathcal S})\) for every split and every \(\mathcal S\subseteq T_r\). No member of \(\mathcal S\) is in \(V_r\), so summing over \(b\in V_r\) gives the weight \(s_r\), and summing over splits,
	\[
	\sum_r s_r\prod_{j\in\mathcal S}I_{jr}\;m^{-k}\chi^{(k)}(Z_{\mathcal S})
	=
	d^{(k)}_{\mathcal S}\,\chi^{(k)}(Z_{\mathcal S}).
	\]
	Steps 2--4 give \eqref{eq:order_k_kernel}.
	
	\noindent\emph{Step 5: Order one.}\quad
	At \(k=1\) the validated case has \(\mathcal J=\varnothing\), and
	\[
	\big(\ell_i^{(r)}\big)_{\{i\}}=\ell_i^\star-R_{p,n},
	\]
	the same in every split because every training set has size \(m\). The training case has, by the definition of \(\bar\phi\),
	\[
	\big(R_r\big)_{\{j\}}=\mathbb E(R_r\mid Z_j)-\mathbb E(R_r)=\frac{\bar\phi(Z_j)}{m}.
	\]
	The weights of order one are \(c^{(1)}_{i;\varnothing}=v_i\) and \(d^{(1)}_{\{i\}}=w_i\), so Steps 2--4 give
	\[
	h_i(Z_i)=v_i(\ell_i^\star-R_{p,n})+w_i\bar\phi(Z_i),
	\]
	which completes Theorem~\ref{thm:dual_role}.
	
	\noindent\emph{Step 6: The second moment of one component.}\quad
	Write \(c_i:=c^{(k)}_{i;\mathcal S\setminus\{i\}}\) and \(d:=d^{(k)}_{\mathcal S}\). The observations are i.i.d., \(\phi^{(k)}\) is symmetric in its last \(k-1\) arguments, and \(\chi^{(k)}\) is symmetric, so relabelling the members of \(\mathcal S\) gives, for \(i\ne i'\) in \(\mathcal S\),
	\[
	\begin{aligned}
		\mathbb E\{\phi^{(k)}(Z_i;Z_{\mathcal S\setminus\{i\}})^2\}&=\tau_{k,n}^2,
		&
		\mathbb E\{\chi^{(k)}(Z_{\mathcal S})^2\}&=\sigma_{\chi,k,n}^2,\\
		\mathbb E\{\phi^{(k)}(Z_i;Z_{\mathcal S\setminus\{i\}})\,\phi^{(k)}(Z_{i'};Z_{\mathcal S\setminus\{i'\}})\}&=\kappa_{k,n},
		&
		\mathbb E\{\phi^{(k)}(Z_i;Z_{\mathcal S\setminus\{i\}})\,\chi^{(k)}(Z_{\mathcal S})\}&=\xi_{k,n}.
	\end{aligned}
	\]
	Squaring \eqref{eq:order_k_kernel} and taking the expectation given \(I\),
	\begin{equation}
		\label{eq:component_second_moment}
		\begin{aligned}
			\mathbb E(h_{\mathcal S}^2\mid I)
			={}&
			\tau_{k,n}^2\sum_{i\in\mathcal S}c_i^2
			+
			\kappa_{k,n}\sum_{i\ne i'\in\mathcal S}c_ic_{i'}\\
			&+
			\sigma_{\chi,k,n}^2d^2
			+
			2\xi_{k,n}\,d\sum_{i\in\mathcal S}c_i .
		\end{aligned}
	\end{equation}
	With \(\phi^{(1)}(Z_i):=\ell_i^\star-R_{p,n}\) and \(\chi^{(1)}:=\bar\phi\), the constants at order one are \(\tau_{1,n}^2=\sigma_V^2\), \(\sigma_{\chi,1,n}^2=\sigma_T^2\), and \(\xi_{1,n}=\gamma_0\).
	
	\noindent\emph{Step 7: The variance given \(I\).}\quad
	Every evaluated contribution has expectation \(R_{p,n}\) given \(I\) and the aggregation weights sum to one, so \(\mathbb E(\widehat R_D\mid I)=R_{p,n}\), and \eqref{eq:lotv} reduces to \(\Var(\widehat R_D)=\mathbb E_I\{\Var(\widehat R_D\mid I)\}\). By orthogonality of the components,
	\[
	\Var(\widehat R_D\mid I)=\sum_{\mathcal S\ne\varnothing}\mathbb E(h_{\mathcal S}^2\mid I)=\sum_{k=1}^n\rho_n^{(k)}(I).
	\]
	
	\noindent\emph{Step 8: The terms of order \(k\ge2\).}\quad
	Summing \eqref{eq:component_second_moment} over the sets of size \(k\ge2\) and averaging over \(I\) gives, by the definitions of the role moments,
	\[
	\mathbb E_I\{\rho_n^{(k)}(I)\}
	=
	\tau_{k,n}^2\Theta_D^{(k)}+\kappa_{k,n}\Theta_D'^{(k)}+\sigma_{\chi,k,n}^2\Psi_D^{(k)}+2\xi_{k,n}\Psi_D'^{(k)},
	\]
	which is nonnegative as an average of conditional variances. The weights of order \(k\) are zero for \(k>m+1\), since a split has \(m\) training observations.
	
	\noindent\emph{Step 9: The first-order term and the series.}\quad
	At \(k=1\) the set has one member, so \eqref{eq:component_second_moment} has no \(\kappa\) term, and by Step~5
	\[
	\mathbb E(h_i^2\mid I)=\sigma_V^2v_i^2+\sigma_T^2w_i^2+2\gamma_0v_iw_i.
	\]
	Summed over \(i\) this is the first-order part of \eqref{eq:cond_var_roles}, whose average over \(I\) is \(\sigma_V^2\Gamma_D+\sigma_T^2\Lambda_D+2\gamma_0\Omega_D\). Summing over \(k\) gives the series of Theorem~\ref{thm:variance_identity}.
	
\end{proof}

\subsection{Proof of Theorem~\ref{thm:pairwise_identity}}
\label{appendix:proof_auc_identity}

The argument does not use class stratification. It holds for any fixed-size design whose inclusion structure is independent of the covariates given the labels, with every training set of size \(m\). Steps~1--5 prove the representation of the components stated in Section~\ref{subsec:pairwise_metrics}, Steps~6--7 prove Theorem~\ref{thm:pairwise_identity}(a), and Step~8 specializes both to class-stratified designs, which gives part (b).

\paragraph{Training compositions, kernels and weights.}
Condition on \(Y\). For split \(r\), write \(V_r^y\) for its validated observations of class \(y\), \(\mathcal P_r=V_r^1\times V_r^0\) for its validated case--control pairs, and
\[
m_{1r}:=\sum_{l\in T_r}Y_l
\]
for the number of cases in its training set, its training composition. A design is class-stratified when \(m_{1r}\) is the same for every split. For a case \(i\) and a control \(j\) validated in split \(r\), write \(\ell_{ij}^{(r)}:=\mathcal C\{\hat f_{T_r}(X_i),\hat f_{T_r}(X_j)\}\), let
\[
\bar\ell^{(r)}_{i\cdot}:=\mathbb E(\ell_{ij}^{(r)}\mid X_i,X_{T_r}),
\qquad
\bar\ell^{(r)}_{\cdot j}:=\mathbb E(\ell_{ij}^{(r)}\mid X_j,X_{T_r})
\]
be the contribution with the partner averaged over its class, and let \(R_r:=\mathrm{AUC}(\hat f_{T_r})\), the expected value of the metric for the fitted model. For a split \(r\) of composition \(m_1\) and sets \(\mathcal J\subseteq T_r\) and \(\mathcal S\subseteq T_r\) of training observations, define the kernels
\[
\begin{aligned}
	\zeta^{(k)}_{m_1}(X_i,X_j;X_{\mathcal J})&:=m^{k-2}\big(\ell_{ij}^{(r)}\big)_{\{i,j\}\cup\mathcal J},
	&
	\chi^{(k)}_{A,m_1}(X_{\mathcal S})&:=m^{k}\big(R_r\big)_{\mathcal S},\\
	\phi^{(k)}_{1,m_1}(X_i;X_{\mathcal J})&:=m^{k-1}\big(\bar\ell^{(r)}_{i\cdot}\big)_{\{i\}\cup\mathcal J},
	&
	\phi^{(k)}_{0,m_1}(X_j;X_{\mathcal J})&:=m^{k-1}\big(\bar\ell^{(r)}_{\cdot j}\big)_{\{j\}\cup\mathcal J},
\end{aligned}
\]
with \(|\mathcal J|=k-2\) for \(\zeta^{(k)}_{m_1}\), \(k\ge2\), with \(|\mathcal J|=k-1\) for \(\phi^{(k)}_{1,m_1}\) and \(\phi^{(k)}_{0,m_1}\), and with \(|\mathcal S|=k\), \(k\ge1\). Step~2 shows that they are the same for every split of composition \(m_1\). With \(\mu_m(m_1/m)\) the expected value of the metric for a model fitted to \(m_1\) cases and \(m-m_1\) controls, as in Section~\ref{subsec:pairwise_metrics}, the kernels at \(k=1\) are \(\phi^{(1)}_{y,m_1}=g_{y,m_1}\) and \(\chi^{(1)}_{A,m_1}(x)=\bar\phi_{y,m_1}(x)\), where
\[
\begin{aligned}
	g_{1,m_1}(x)&:=\mathbb E(\bar\ell^{(r)}_{i\cdot}\mid X_i=x)-\mu_m(m_1/m),\\
	g_{0,m_1}(x)&:=\mathbb E(\bar\ell^{(r)}_{\cdot j}\mid X_j=x)-\mu_m(m_1/m),\\
	\bar\phi_{y,m_1}(x)&:=m\big\{\mathbb E(R_r\mid X_l=x)-\mu_m(m_1/m)\big\},
\end{aligned}
\]
the first two for a validated case and a validated control, with the partner and the training set averaged out, and the third for a training observation \(l\) of class \(y\). All three have mean zero within their class. At \(k=2\), \(\phi^{(2)}_{1,m_1}\) and \(\phi^{(2)}_{0,m_1}\) are the validation--training interactions of a case and of a control, \(\chi^{(2)}_{A,m_1}\) is the training--training interaction, and \(\zeta^{(2)}_{m_1}(X_i,X_j)\) is the validation-pair interaction, at composition \(m_1\). With \(M_A:=\sum_r|V_r^1||V_r^0|\) the number of validated pairs, define the weights
\[
\begin{aligned}
	e^{(k)}_{ij;\mathcal J,m_1}&:=\frac{1}{M_Am^{k-2}}\sum_{r:\,m_{1r}=m_1}\mathbb I(i,j\in V_r)\prod_{l\in\mathcal J}I_{lr},\\
	c^{(k)}_{i;\mathcal J,m_1}&:=\frac{1}{M_Am^{k-1}}\sum_{r:\,m_{1r}=m_1}\mathbb I(i\in V_r)\,|V_r^{1-Y_i}|\prod_{l\in\mathcal J}I_{lr},\\
	d^{(k)}_{\mathcal S,m_1}&:=\frac{1}{m^{k}}\sum_{r:\,m_{1r}=m_1}\frac{|V_r^1||V_r^0|}{M_A}\prod_{l\in\mathcal S}I_{lr},
\end{aligned}
\qquad
s_{m_1}:=\sum_{r:\,m_{1r}=m_1}\frac{|V_r^1||V_r^0|}{M_A}.
\]
Among the splits of composition \(m_1\), they count those in which a case and a control of the set are validated together, those in which one member is validated, and those in which the whole set trains, and \(s_{m_1}\) is the share of the validated pairs that come from splits of composition \(m_1\), so that \(\sum_{m_1} s_{m_1}=1\). Summed over \(m_1\) they are the weights \(e^{(k)}\), \(c^{(k)}\), \(d^{(k)}\) of Section~\ref{subsec:pairwise_metrics}. At \(k=1\) write \(\tilde v_{i,m_1}:=c^{(1)}_{i;\varnothing,m_1}\) and \(w_{l,m_1}:=d^{(1)}_{\{l\},m_1}\) for the validation and training weights that an observation receives from splits of composition \(m_1\), so that \(\sum_{m_1}\tilde v_{i,m_1}=\tilde v_i\) and \(\sum_{m_1} w_{l,m_1}=w_l\). At \(k=2\), \(\sum_{m_1} e^{(2)}_{ij,m_1}=v_{(i,j)}\) is the validation weight of the pair. The covariances of the first-order kernels across two compositions are
\[
\begin{aligned}
	\sigma_{V\mid y}(m_1,m_1')&:=\mathbb E\{g_{y,m_1}(X)\,g_{y,m_1'}(X)\mid Y=y\},\\
	\sigma_{T\mid y}(m_1,m_1')&:=\mathbb E\{\bar\phi_{y,m_1}(X)\,\bar\phi_{y,m_1'}(X)\mid Y=y\},\\
	\gamma_{0\mid y}(m_1,m_1')&:=\mathbb E\{g_{y,m_1}(X)\,\bar\phi_{y,m_1'}(X)\mid Y=y\},
\end{aligned}
\]
and on the diagonal \(m_1=m_1'\) they are \(\sigma_{V\mid y}^2\), \(\sigma_{T\mid y}^2\), and \(\gamma_{0\mid y}\) at composition \(m_1\). The proof shows that, for every set \(\mathcal S\) of \(k\ge2\) observations,
\begin{equation}
	\label{eq:auc_components_general}
	\begin{aligned}
		h_{\mathcal S}
		=\sum_{m_1}\Big\{&
		\sum_{\substack{i,j\in\mathcal S\\ Y_i=1,\,Y_j=0}}e^{(k)}_{ij;\mathcal S\setminus\{i,j\},m_1}\,\zeta^{(k)}_{m_1}(X_i,X_j;X_{\mathcal S\setminus\{i,j\}})\\
		&+
		\sum_{i\in\mathcal S}c^{(k)}_{i;\mathcal S\setminus\{i\},m_1}\,\phi^{(k)}_{Y_i,m_1}(X_i;X_{\mathcal S\setminus\{i\}})
		+
		d^{(k)}_{\mathcal S,m_1}\,\chi^{(k)}_{A,m_1}(X_{\mathcal S})\Big\},
	\end{aligned}
\end{equation}
that
\begin{equation}
	\label{eq:auc_variance_general}
	\begin{aligned}
		\mathbb E(\widehat R_D\mid I,Y)={}&\sum_{m_1} s_{m_1}\,\mu_m(m_1/m),\\
		\Var(\widehat R_D\mid I,Y)
		={}&\sum_i\sum_{m_1,m_1'}\big\{\tilde v_{i,m_1}\tilde v_{i,m_1'}\,\sigma_{V\mid Y_i}(m_1,m_1')
		+w_{i,m_1}w_{i,m_1'}\,\sigma_{T\mid Y_i}(m_1,m_1')\\
		&\qquad\qquad+2\,\tilde v_{i,m_1}w_{i,m_1'}\,\gamma_{0\mid Y_i}(m_1,m_1')\big\}
		+\sum_{k\ge2}\rho_n^{(k)}(I,Y),
	\end{aligned}
\end{equation}
with \(\rho_n^{(k)}(I,Y)\) a quadratic form in the weights of order \(k\) whose coefficients are indexed by a pair of compositions and the class pattern of \(\mathcal S\), and that
\begin{equation}
	\label{eq:auc_label_lotv_general}
	\Var(\widehat R_D\mid Y)=\mathbb E_I\big\{\Var(\widehat R_D\mid I,Y)\,\big|\,Y\big\}+\Var_I\Big\{\sum_{m_1} s_{m_1}\,\mu_m(m_1/m)\,\Big|\,Y\Big\}.
\end{equation}
The first display is the representation stated in Section~\ref{subsec:pairwise_metrics} and the second is Theorem~\ref{thm:pairwise_identity}(a), with \(\chi^{(k)}_{A,m_1}\) written \(\chi^{(k)}_{m_1}\) and \(g_{y,m_1}\) written \(\ell^\star_{i,m_1}-\mu_m(m_1/m)\) there. For a class-stratified design there is one composition, the sums over \(m_1\) have one term, the last term of \eqref{eq:auc_label_lotv_general} is zero, and the displays reduce to Theorem~\ref{thm:pairwise_identity}(b) (Step~8).

\begin{proof}
	Condition on \(Y\) and \(I\) throughout. The covariates are independent given \(Y\), with \(X_l\sim P_{Y_l}\), so the components of Step~1 of Section~\ref{appendix:proof_variance_identity}, taken in the covariates under this product law, have the same properties, and \eqref{eq:support_rule} holds. Write the estimate as
	\[
	\widehat R_D=\sum_r\sum_{(i,j)\in\mathcal P_r}a_{ijr}\ell_{ij}^{(r)},
	\]
	where \(a_{ijr}\ge0\) are aggregation weights summing to one, pooled aggregation being \(a_{ijr}=1/M_A\). By linearity,
	\[
	h_{\mathcal S}=\sum_r\sum_{(i,j)\in\mathcal P_r}a_{ijr}\big(\ell_{ij}^{(r)}\big)_{\mathcal S}.
	\]
	
	\noindent\emph{Step 1: Which terms contribute.}\quad
	The contribution \(\ell_{ij}^{(r)}\) is a function of \(X_i\), \(X_j\), and \(X_{T_r}\), so by \eqref{eq:support_rule} its component on \(\mathcal S\) is zero unless \(\mathcal S\subseteq\{i,j\}\cup T_r\). Since \(V_r\cap T_r=\varnothing\), there are three cases. Both \(i\) and \(j\) are in \(\mathcal S\), with \(\mathcal S\setminus\{i,j\}\subseteq T_r\). Exactly one of them is in \(\mathcal S\), with the rest of \(\mathcal S\) in \(T_r\). Neither is in \(\mathcal S\), and \(\mathcal S\subseteq T_r\).
	
	\noindent\emph{Step 2: Both validated members in \(\mathcal S\).}\quad
	Let \(\mathcal J:=\mathcal S\setminus\{i,j\}\subseteq T_r\), \(|\mathcal J|=k-2\). The component \((\ell_{ij}^{(r)})_{\{i,j\}\cup\mathcal J}\) is an alternating sum of the conditional means \(\mathbb E(\ell_{ij}^{(r)}\mid X_{\mathcal A})\), \(\mathcal A\subseteq\{i,j\}\cup\mathcal J\), each of which integrates the training covariates outside \(\mathcal A\) against their class laws. The law of the fit is unchanged by permuting its training observations and the covariates of a class are i.i.d., so this component depends on the split and on the choice of \(\mathcal J\) only through the classes of the members of \(\mathcal J\) and the numbers of cases and controls among the training observations outside \(\mathcal J\). Every training set has \(m\) observations, so given the classes of the members of \(\mathcal J\) these numbers are fixed by \(m_{1r}\). The component is therefore the same function of \((X_i,X_j,X_{\mathcal J})\) for every split of the same composition, and it is symmetric in the training arguments of each class. By definition it is \(m^{-(k-2)}\zeta^{(k)}_{m_{1r}}(X_i,X_j;X_{\mathcal J})\). Summing \(a_{ijr}\) over the splits of composition \(m_1\) in which \(i\) and \(j\) are validated and \(\mathcal J\) trains gives
	\[
	e^{(k)}_{ij;\mathcal J,m_1}\,\zeta^{(k)}_{m_1}(X_i,X_j;X_{\mathcal J}),
	\qquad
	e^{(k)}_{ij;\mathcal J,m_1}=m^{-(k-2)}\sum_{r:\,m_{1r}=m_1}a_{ijr}\prod_{l\in\mathcal J}I_{lr},
	\]
	which is the weight defined above under pooled aggregation. Summing over \(m_1\) gives the first term of \eqref{eq:auc_components_general}.
	
	\noindent\emph{Step 3: One validated member in \(\mathcal S\).}\quad
	Let \(i\in\mathcal S\), \(j\notin\mathcal S\), and \(\mathcal J:=\mathcal S\setminus\{i\}\subseteq T_r\), \(|\mathcal J|=k-1\). For \(\mathcal A\subseteq\mathcal S\), \(X_j\) is independent of \((X_i,X_{T_r})\) given \(Y\), so
	\[
	\mathbb E(\ell_{ij}^{(r)}\mid X_{\mathcal A})=\mathbb E(\bar\ell^{(r)}_{i\cdot}\mid X_{\mathcal A}).
	\]
	Hence
	\[
	\big(\ell_{ij}^{(r)}\big)_{\mathcal S}=\big(\bar\ell^{(r)}_{i\cdot}\big)_{\{i\}\cup\mathcal J}=m^{-(k-1)}\phi^{(k)}_{1,m_{1r}}(X_i;X_{\mathcal J}),
	\]
	the same for every control partner \(j\in V_r^0\) and, by the argument of Step~2, for every split of the same composition and every \(\mathcal J\) with the same class pattern. Summing over the partners gives \(\sum_{j\in V_r^0}a_{ijr}\), which is \(|V_r^0|/M_A\) under pooled aggregation, and summing over the splits of composition \(m_1\) gives \(c^{(k)}_{i;\mathcal J,m_1}\,\phi^{(k)}_{1,m_1}(X_i;X_{\mathcal J})\). A validated control in \(\mathcal S\) with its case partner outside contributes \(c^{(k)}_{j;\mathcal J,m_1}\,\phi^{(k)}_{0,m_1}(X_j;X_{\mathcal J})\) in the same way. Summing over \(m_1\) gives the second term of \eqref{eq:auc_components_general}.
	
	\noindent\emph{Step 4: All of \(\mathcal S\) trains.}\quad
	Let \(i,j\notin\mathcal S\subseteq T_r\). Since \((X_i,X_j)\) is independent of \(X_{T_r}\) given \(Y\),
	\[
	\mathbb E(\ell_{ij}^{(r)}\mid X_{T_r})=R_r,
	\qquad
	\mathbb E(\ell_{ij}^{(r)}\mid X_{\mathcal A})=\mathbb E(R_r\mid X_{\mathcal A}),\quad\mathcal A\subseteq\mathcal S.
	\]
	Hence \((\ell_{ij}^{(r)})_{\mathcal S}=(R_r)_{\mathcal S}=m^{-k}\chi^{(k)}_{A,m_{1r}}(X_{\mathcal S})\) for every pair of \(\mathcal P_r\), by the argument of Step~2 the same function for every split of the same composition and every \(\mathcal S\) with the same class pattern. Summing \(a_{ijr}\) over the pairs of \(\mathcal P_r\) gives \(s_r:=\sum_{(i,j)\in\mathcal P_r}a_{ijr}\), which is \(|V_r^1||V_r^0|/M_A\) under pooled aggregation, and summing over the splits of composition \(m_1\) gives \(d^{(k)}_{\mathcal S,m_1}\,\chi^{(k)}_{A,m_1}(X_{\mathcal S})\). Summing over \(m_1\) gives the third term, and Steps 1--4 give \eqref{eq:auc_components_general}.
	
	\noindent\emph{Step 5: Order one.}\quad
	Steps 1--4 hold at \(k=1\) as well, where Step~2 is empty. In Step~3, the component of \(\ell_{ij}^{(r)}\) on \(\{i\}\) is \(\mathbb E(\ell_{ij}^{(r)}\mid X_i)-\mathbb E(\ell_{ij}^{(r)})=g_{Y_i,m_{1r}}(X_i)\). In Step~4, the component on \(\{l\}\) for \(l\in T_r\) is \(\mathbb E(R_r\mid X_l)-\mathbb E(R_r)=\bar\phi_{Y_l,m_{1r}}(X_l)/m\). Summing as there,
	\[
	h_{\{i\}}=\sum_{m_1}\big\{\tilde v_{i,m_1}\,g_{Y_i,m_1}(X_i)+w_{i,m_1}\,\bar\phi_{Y_i,m_1}(X_i)\big\}.
	\]
	
	\noindent\emph{Step 6: Second moments and the series.}\quad
	By orthogonality of the components,
	\[
	\Var(\widehat R_D\mid I,Y)=\sum_{\mathcal S}\mathbb E(h_{\mathcal S}^2\mid I,Y),
	\]
	and the sum over the sets of size \(k\) is nonnegative as a sum of conditional variances. Given \((I,Y)\) the covariate \(X_i\) has law \(P_{Y_i}\), and \(g_{y,m_1}\) and \(\bar\phi_{y,m_1}\) have mean zero within class \(y\), so by Step~5
	\[
	\begin{aligned}
		\mathbb E(h_{\{i\}}^2\mid I,Y)=\sum_{m_1,m_1'}\big\{&\tilde v_{i,m_1}\tilde v_{i,m_1'}\,\sigma_{V\mid Y_i}(m_1,m_1')+w_{i,m_1}w_{i,m_1'}\,\sigma_{T\mid Y_i}(m_1,m_1')\\
		&+\tilde v_{i,m_1}w_{i,m_1'}\,\gamma_{0\mid Y_i}(m_1,m_1')+w_{i,m_1}\tilde v_{i,m_1'}\,\gamma_{0\mid Y_i}(m_1',m_1)\big\}.
	\end{aligned}
	\]
	Exchanging the labels \(m_1\) and \(m_1'\) in the last sum shows that it equals the third, which gives the factor \(2\) in \eqref{eq:auc_variance_general}. For \(k\ge2\) the kernels of a set \(\mathcal S\) are functions of \(X_{\mathcal S}\) with the symmetries of Step~2, and the covariates of a class are i.i.d., so the second moments and cross moments of the kernels of \(\mathcal S\) at two compositions, over all assignments of roles to its members, depend on \(\mathcal S\) only through its numbers of cases and controls. Squaring \eqref{eq:auc_components_general} and taking the expectation given \((I,Y)\) therefore gives a quadratic form in the weights of \(\mathcal S\), with coefficients indexed by the pair of compositions and the class pattern of \(\mathcal S\). This gives the variance in \eqref{eq:auc_variance_general}.
	
	\noindent\emph{Step 7: The conditional mean and the variance given the labels.}\quad
	A validated pair is independent of its training set given \(Y\), and a pair validated in a split of composition \(m_1\) has expected contribution \(\mu_m(m_1/m)\). Hence
	\[
	\mathbb E(\widehat R_D\mid I,Y)=\sum_rs_r\,\mu_m(m_{1r}/m)=\sum_{m_1} s_{m_1}\,\mu_m(m_1/m),
	\]
	the mean in \eqref{eq:auc_variance_general}. The law of total variance over \(I\) given \(Y\) gives \eqref{eq:auc_label_lotv_general}. When relabeling the observations carries the design's law given \(Y\) to its law given any other label vector with the same class counts, as for a design independent of the labels whose law is unchanged by permuting the observations, both terms of \eqref{eq:auc_label_lotv_general} depend on \(Y\) only through the class counts.
	
	\noindent\emph{Step 8: Class-stratified designs.}\quad
	Let every training set have \(m_1\) cases and \(m_0\) controls. There is then one composition, \(m_{1r}\equiv m_1\), with \(s_{m_1}=1\), and the weights of Steps~2--4 are \(e^{(k)}\), \(c^{(k)}\), and \(d^{(k)}\). Write the kernels without the composition index, \(\zeta^{(k)}\), \(\phi^{(k)}_{Y_i}\), and \(\chi^{(k)}_A\), and at \(k=1\) write \(g_{Y_i,m}\) and \(\bar\phi_{Y,m}(X_l)\) for \(g_{Y_i,m_1}\) and \(\bar\phi_{Y_l,m_1}(X_l)\) at this common composition. These are the first-order objects of Section~\ref{subsec:assumptions} with the training size marked, \(g_{Y_i,m}(X_i)=\ell_i^\star-R_{p,n}(Y)\) and \(\bar\phi_{Y,m}=\bar\phi_Y\), because \(\mu_m(m_1/m)=R_{p,n}(Y)\). The representation \eqref{eq:auc_components_general} is
	\begin{equation}
		\label{eq:auc_components}
		\begin{aligned}
			h_{\mathcal S}
			={}&
			\sum_{\substack{i,j\in\mathcal S\\ Y_i=1,\,Y_j=0}}e^{(k)}_{ij;\mathcal S\setminus\{i,j\}}\,\zeta^{(k)}(X_i,X_j;X_{\mathcal S\setminus\{i,j\}})\\
			&+
			\sum_{i\in\mathcal S}c^{(k)}_{i;\mathcal S\setminus\{i\}}\,\phi^{(k)}_{Y_i}(X_i;X_{\mathcal S\setminus\{i\}})
			+
			d^{(k)}_{\mathcal S}\,\chi^{(k)}_A(X_{\mathcal S}),
		\end{aligned}
	\end{equation}
	which is the class-stratified case of the representation in Section~\ref{subsec:pairwise_metrics}, with \(h_{\{i\}}=\tilde v_i\,g_{Y_i,m}(X_i)+w_i\,\bar\phi_{Y,m}(X_i)\) at order one. Each double sum over compositions in Step~6 has one term, with the constants \(\sigma_{V\mid y}^2\), \(\sigma_{T\mid y}^2\), and \(\gamma_{0\mid y}\) at the training composition \((m_1,m_0)\), so
	\[
	\mathbb E(h_{\{i\}}^2\mid I,Y)=\tilde v_i^2\sigma^2_{V\mid Y_i}+w_i^2\sigma^2_{T\mid Y_i}+2\tilde v_iw_i\gamma_{0\mid Y_i},
	\]
	and the sum over \(i\) is the three first-order sums of Theorem~\ref{thm:pairwise_identity}(b). The conditional mean is \(R_{p,n}(Y)\) for every \(I\), so the last term of \eqref{eq:auc_label_lotv_general} is zero, and averaging the conditional variances over \(I\) given \(Y\) gives \(\Var(\widehat R_D\mid Y)\). It depends on \(Y\) only through the class counts, since the covariates of a class are i.i.d., the training composition is a function of the class counts, and relabeling the observations carries the design's law given \(Y\) to its law given any other label vector with the same class counts. Since \(\mathbb E(\widehat R_D\mid Y)=R_{p,n}(Y)\) is also a function of the class counts, \(\Var(\widehat R_D\mid n_1)=\Var(\widehat R_D\mid Y)\), which is Theorem~\ref{thm:pairwise_identity}(b). The law of total variance over \(Y\) then adds \(\Var\{R_{p,n}(Y)\}\), which gives the variance over the population stated after the theorem.
\end{proof}

\paragraph{Other pair-average metrics.}
This proves the statement on pair averages without labels in Section~\ref{subsec:pairwise_metrics}.

\begin{proof}
	\noindent\emph{Step 1: Symmetric metrics.}\quad
	For a symmetric pair-average metric in which every pair of validated observations enters, Steps 1--4 of Section~\ref{appendix:proof_auc_identity} hold without conditioning on the labels, with the components taken in the observations, one kernel in place of \(\phi^{(k)}_1\) and \(\phi^{(k)}_0\), and \(|V_r|-1\) in place of \(|V_r^{1-Y_i}|\). At \(k=1\) the validation influence is \(\ell_i^\star-R_{p,n}\), with \(\ell_i^\star:=\mathbb E(\ell_{ij}^{(r)}\mid Z_i)\) for any validated partner \(j\).
	
	\noindent\emph{Step 2: The weights.}\quad
	Under pooled aggregation at a fixed-size design every split has \((n-m)(n-m-1)/2\) pairs, so
	\[
	\tilde v_i=\frac{N_i(n-m-1)}{M_A}=2v_i,
	\]
	and every split carries the same share of the pairs, so \(w_i\) is the unit-level training weight. Hence
	\[
	h_i=v_i\{2(\ell_i^\star-R_{p,n})\}+w_i\bar\phi(Z_i),
	\]
	and \eqref{eq:cond_var_roles} holds with \(2(\ell_i^\star-R_{p,n})\) in place of \(\ell_i^\star-R_{p,n}\), its remainder including the validation-pair interactions.
\end{proof}

\subsection{The bootstrap identity}
\label{appendix:proof_bootstrap_identity}

This states the bootstrap condition A3 named in Section~\ref{subsec:variance_decomposition} and proves the first-order identity for bootstrap cross-validation stated there.

\smallskip\noindent\textbf{A3 (Bootstrap).} \emph{For the bootstrap, \(\ell_i^\star\), \(\phi_i\), and \(\bar\phi\) are defined with the training set drawn as a resample and the influence of a training observation counted per unit of multiplicity. (i) Given the resamples, \(\widehat R_D-\mathbb E(\widehat R_D\mid I)=\sum_iv_i\{\ell_i^\star-\mathbb E(\ell_i^\star)\}+\sum_iw_i\bar\phi(Z_i)+r_D\), with \(\mathbb E(r_D^2)=o(n^{-1})\). (ii) The mean of the estimate given the resamples has variance \(\Var_I\{\mathbb{E}(\widehat R_D\mid I)\}=o(n^{-1})\). (iii) For comparisons with fixed-size designs at \(p=1-e^{-1}\), the model's limiting constants are the same under weighted and fixed-size training.}

\smallskip\noindent Under A3(ii) the second term of \eqref{eq:lotv} is \(o(n^{-1})\), and under A3(i)--(ii) the first-order part of Theorem~\ref{thm:variance_identity} holds up to \(o(n^{-1})\).

\begin{proof}
	\noindent\emph{Step 1: The conditional mean.}\quad
	By the law of total variance,
	\[
	\Var(\widehat R_D)=\mathbb E_I\{\Var(\widehat R_D\mid I)\}+\Var_I\{\mathbb E(\widehat R_D\mid I)\},
	\]
	and the second term is \(o(n^{-1})\) by A3(ii).
	
	\noindent\emph{Step 2: The linear part.}\quad
	Let
	\[
	A_D:=\sum_iv_i\{\ell_i^\star-\mathbb E(\ell_i^\star)\}+\sum_iw_i\bar\phi(Z_i).
	\]
	Given \(I\), its summands are independent across \(i\) with mean zero, so
	\[
	\Var(A_D\mid I)=\sigma_V^2\sum_iv_i^2+\sigma_T^2\sum_iw_i^2+2\gamma_0\sum_iv_iw_i .
	\]
	
	\noindent\emph{Step 3: The remainder.}\quad
	By A3(i), \(r_D\) has mean zero given \(I\), and
	\[
	\Var(\widehat R_D\mid I)=\Var(A_D\mid I)+2\Cov(A_D,r_D\mid I)+\mathbb E(r_D^2\mid I).
	\]
	The bootstrap role moments are \(1/n+O\{1/(nS)\}+o(n^{-1})\) by Step~3 of Section~\ref{appendix:proof_role_moments}, and the constants are bounded by A1, so \(\mathbb E_I\{\Var(A_D\mid I)\}=O(n^{-1})\). By the Cauchy--Schwarz inequality,
	\[
	\big|\mathbb E_I\{\Cov(A_D,r_D\mid I)\}\big|\le\big[\mathbb E_I\{\Var(A_D\mid I)\}\big]^{1/2}\{\mathbb E(r_D^2)\}^{1/2}=O(n^{-1/2})\,o(n^{-1/2}).
	\]
	Averaging over \(I\) and adding Step~1,
	\[
	\Var(\widehat R_D)=\sigma_V^2\Gamma_D+\sigma_T^2\Lambda_D+2\gamma_0\Omega_D+o(n^{-1}).
	\]
\end{proof}

\subsection{The training terms when the model is fit by minimizing the reported loss}
\label{appendix:proof_refit_second_order}

This derives the statement of Section~\ref{subsec:refitting}, for a fixed-size design with \(\max_iv_i\le c/n\) (Section~\ref{appendix:stability_assumptions}). Let \(\ell_i(f)=L(Y_i,f(X_i))\) with \(L\) twice continuously differentiable in its second argument and with bounded second derivative, as squared error, the Brier score, and log-loss on predictions bounded away from \(0\) and \(1\) are. Let the fitted model have the influence-function expansion of Section~\ref{appendix:stability_assumptions}, in which a training observation at \(z\) moves the prediction at \(x\) by \(\psi(z;x)/m\) to first order, and write \(f_p^\star\) for the limit of the fitted model and \(R(f)=\mathbb E[L\{Y,f(X)\}]\). The model is fit by minimizing the reported loss when the derivative of \(R\) at \(f_p^\star\) is zero in the direction \(\psi(z;\cdot)\) for every \(z\). This holds when the model minimizes the average of \(L\) over its class, and when \(f_p^\star\) minimizes \(R\) over all functions, as for a correctly specified model evaluated by a strictly proper scoring rule.

\begin{proof}
	\noindent\emph{Step 1: The limit of the training influence.}\quad
	By Steps~1 and~4 of the influence-function case in Section~\ref{appendix:stability_assumptions}, $\bar\phi_m\to\bar\phi_\infty$ in $L_2(P)$.
	
	\noindent\emph{Step 2: The limit is a directional derivative.}\quad
	For $R(f)=\mathbb E[L\{Y,f(X)\}]$ and a direction $h\in L_2(P_X)$, the bound on the second derivative of $L$ bounds its second-order remainder by a constant times $t^2h(X)^2$, so
	\[
	\frac{d}{dt}R(f_p^\star+th)\Big|_{t=0}=\mathbb E\big[\partial_2L\{Y,f_p^\star(X)\}\,h(X)\big].
	\]
	At $h=\psi(z;\cdot)$, which is in $L_2(P_X)$ for almost every $z$, this is $\bar\phi_\infty(z)$.
	
	\noindent\emph{Step 3: Zero when the model is fit by minimizing the reported loss.}\quad
	If the derivative of Step 2 is zero in the direction $\psi(z;\cdot)$ for every $z$, then $\bar\phi_\infty\equiv0$.
	
	\noindent\emph{Step 4: The constants.}\quad
	By Step 1,
	\[
	\sigma_{T,m}^2=\Var\{\bar\phi_m(Z)\}\to\Var\{\bar\phi_\infty(Z)\}=0,
	\]
	and by the Cauchy--Schwarz inequality and the moment condition of A1,
	\[
	|\gamma_{0,m}|=\big|\Cov\{\ell_i^\star,\bar\phi_m(Z_i)\}\big|
	\le\{\Var(\ell_i^\star)\,\sigma_{T,m}^2\}^{1/2}\to0.
	\]
	
	\noindent\emph{Step 5: The variance.}\quad
	For a fixed-size design $w_i\le1/m$, so $\sum_iw_i^2\le1/m$ and $|\sum_iv_iw_i|\le\max_iw_i\sum_iv_i\le1/m$. Hence
	\[
	\big|\sigma_{T,m}^2\Lambda_D+2\gamma_{0,m}\Omega_D\big|
	\le\frac{\sigma_{T,m}^2+2|\gamma_{0,m}|}{m}=o(n^{-1}).
	\]
	Step~3 of the influence-function case in Section~\ref{appendix:stability_assumptions} gives (S1)--(S2), so A2(i) holds by (b) of Section~\ref{appendix:stability_assumptions}, since $\max_iv_i\le c/n$, and Theorem~\ref{thm:variance_identity} gives $\Var(\widehat R_D)=\sigma_V^2\Gamma_D+o(n^{-1})$.
\end{proof}

\subsection{Derivation of \eqref{eq:pair_cov}, \eqref{eq:pairwise_form}, and the constants of Bengio and Grandvalet}
\label{appendix:proof_pairwise_form}

Throughout, the design is fixed-size with \(\max_iv_i\le c/n\), A1 and (S1)--(S2) of Section~\ref{appendix:stability_assumptions} hold, and \eqref{eq:app_if_expansion} writes a contribution of unit $i$ validated on split $r$ as
\[
\ell_i^{(r)}
=
\ell_i^\star+\frac{1}{m}\sum_{k\in T_r}\phi_i(Z_k)+\varrho_{ir}.
\]

\begin{proof}
	\noindent\emph{Step 1: $\sigma_\ell^2$.}\quad
	At fixed $Z_i$ the three terms are the parts of $\ell_i^{(r)}$ of order zero, one, and two and higher in the training observations, so they are uncorrelated, and the terms $\phi_i(Z_k)$ are uncorrelated across $k$. Hence
	\[
	\sigma_\ell^2
	=
	\sigma_V^2+\frac{\mathbb E\{\phi_i(Z)^2\}}{m}+\mathbb E(\varrho_{ir}^2).
	\]
	The second term is $o(1)$ by (S1). The third is $o(n^{-1})$ by (S2).
	
	\noindent\emph{Step 2: \eqref{eq:pair_cov}.}\quad
	Let units $i\ne i'$ be validated on splits $r$ and $r'$. Expanding both contributions, the covariances of order $n^{-1}$ are
	\begin{itemize}
		\item[(1)] between $\ell_i^\star$ and the term $k=i$ of the expansion of $\ell_{i'}^{(r')}$, present when $i\in T_{r'}$, and symmetrically;
		\item[(2)] between the two sums through their common training observations $k\in T_r\cap T_{r'}$.
	\end{itemize}
	Averaging over $Z_{i'}$ first, since $\mathbb E\{\phi_{i'}(z)\}=\bar\phi(z)$,
	\[
	\mathbb E\{\ell_i^\star\,\phi_{i'}(Z_i)\}=\mathbb E\{\ell_i^\star\,\bar\phi(Z_i)\}=\gamma_0 .
	\]
	Averaging over $Z_i$ and $Z_{i'}$ separately,
	\[
	\mathbb E\{\phi_i(Z_k)\,\phi_{i'}(Z_k)\}=\mathbb E\{\bar\phi(Z_k)^2\}=\sigma_T^2 .
	\]
	The remaining covariances are a $\kappa_n$ term, $\xi_n$ terms, and the covariance of the two remainders, of sizes at most
	\[
	\frac{\mathbb I(i\in T_{r'})\mathbb I(i'\in T_r)\,|\kappa_n|}{m^2},
	\qquad
	\frac{2\,|T_r\cap T_{r'}|\,|\xi_n|}{m^3},
	\qquad
	\sup\mathbb E(\varrho_{ir}^2).
	\] By (S1), $|\kappa_n|\le\tau_n^2\le\mathbb E\{\phi_i(Z)^2\}=o(n)$. Averaged over $Z_i$, the component of $\varrho_{ir}$ on two training observations is the component of $R_r$ on them, so
	\[
	\mathbb E(\varrho_{ir}^2)\ge\binom m2\frac{\sigma_{\chi,n}^2}{m^4},
	\]
	and (S2) gives $\sigma_{\chi,n}^2=o(n)$ and $|\xi_n|\le\tau_n\sigma_{\chi,n}=o(n)$. Each remaining covariance is therefore $o(n^{-1})$, and
	\[
	\Cov\big(\ell_i^{(r)},\ell_{i'}^{(r')}\,\big|\,I\big)
	=
	\frac{|T_r\cap T_{r'}|}{m^2}\,\sigma_T^2
	+
	\frac{\mathbb I(i\in T_{r'})+\mathbb I(i'\in T_r)}{m}\,\gamma_0
	+
	o(n^{-1}).
	\]
	
	\noindent\emph{Step 3: $C_\ell$.}\quad
	For training sets drawn independently, each uniform among the subsets of size $m$,
	\[
	\mathbb E(|T_r\cap T_{r'}|)=\frac{m^2}{n}+O(1),
	\qquad
	\mathbb P(i\in T_{r'})=\frac mn+O(n^{-1}),
	\]
	so averaging the display of Step 2 over the two training sets gives
	\[
	C_\ell=\frac{\sigma_T^2+2\gamma_0}{n}+o(n^{-1}).
	\]
	
	\noindent\emph{Step 4: \eqref{eq:pairwise_form}.}\quad
	With $\max_iv_i\le c/n$, (S1)--(S2) give A2(i) by (b) of Section~\ref{appendix:stability_assumptions}, and Theorem~\ref{thm:variance_identity} gives $\Var(\widehat R_D)=\sigma_V^2\Gamma_D+\sigma_T^2\Lambda_D+2\gamma_0\Omega_D+o(n^{-1})$. Adding and subtracting $(\sigma_T^2+2\gamma_0)/n$,
	\[
	\Var(\widehat R_D)
	=
	\sigma_V^2\,\Gamma_D+\frac{\sigma_T^2+2\gamma_0}{n}+\sigma_T^2\Big(\Lambda_D-\frac1n\Big)+2\gamma_0\Big(\Omega_D-\frac1n\Big)+o(n^{-1}).
	\]
	By Step 1 and $\Gamma_D=O(n^{-1})$, and by Step 3,
	\[
	\sigma_V^2\Gamma_D=\sigma_\ell^2\Gamma_D+o(n^{-1}),
	\qquad
	\frac{\sigma_T^2+2\gamma_0}{n}=C_\ell+o(n^{-1}).
	\]
	
	\noindent\emph{Step 5: The constants of Bengio and Grandvalet.}\quad
	Let $K$ divide $n$, so that $m=pn$ and $q=1/K$. Two units validated in the same fold have $|T_r\cap T_{r'}|=m$, and neither is in the other's training set, so by Step 2
	\[
	\omega=\frac{m}{m^2}\,\sigma_T^2+o(n^{-1})=\frac{\sigma_T^2}{pn}+o(n^{-1}).
	\]
	Two units validated in different folds have $|T_r\cap T_{r'}|=(1-2q)n$, and each is in the other's training set, so
	\[
	\gamma^{\mathrm{BG}}=\frac{(1-2q)n}{m^2}\,\sigma_T^2+\frac{2}{m}\,\gamma_0+o(n^{-1})
	=\frac{(1-2q)\sigma_T^2/p+2\gamma_0}{pn}+o(n^{-1}).
	\]
	With $p+q=1$,
	\[
	q\,\omega+p\,\gamma^{\mathrm{BG}}
	=
	\frac{q+1-2q}{pn}\,\sigma_T^2+\frac2n\,\gamma_0+o(n^{-1})
	=
	C_\ell+o(n^{-1}).
	\]
\end{proof}

\subsection{The label part of \(\sigma_T^2\)}
\label{appendix:auc_label_part}

This proves the display for \(\sigma_T^2\) in Section~\ref{subsec:pairwise_metrics}. Let \(T\) be a training set of \(m\) independent draws from \(P\), \(R_T:=\mathrm{AUC}(\hat f_T)\), and
\[
\bar\phi(z):=m\{\mathbb E(R_T\mid Z_l=z)-\mathbb E(R_T)\},\qquad l\in T,
\]
the training influence of Section~\ref{subsec:assumptions}, with the label of the training observation random. Write \(\bar\phi_{y,m,m_1}\) for the training influence \(\bar\phi_Y\) of an observation of class \(y\) in a training set of \(m\) observations with \(m_1\) cases, and \(\bar\phi_y^\circ\) for its \(L_2\) limit as \(m_1/m\to\pi_1\) under A4(ii). Let \(\delta_m:=m^{-1/3}\).

\begin{proof}
	\noindent\emph{Step 1: Split by the label.}\quad
	For \(z=(x,y)\),
	\[
	\bar\phi(z)
	=
	m\{\mathbb E(R_T\mid Z_l=z)-\mathbb E(R_T\mid Y_l=y)\}
	+
	m\{\mathbb E(R_T\mid Y_l=y)-\mathbb E(R_T)\}.
	\]
	
	\noindent\emph{Step 2: The label term.}\quad
	Let \(N\sim\mathrm{Bin}(m-1,\pi_1)\) be the number of cases among the other training observations. Then
	\[
	m\{\mathbb E(R_T\mid Y_l=1)-\mathbb E(R_T)\}
	=
	(1-\pi_1)\,\mathbb E\Big[m\Big\{\mu_m\Big(\frac{N+1}m\Big)-\mu_m\Big(\frac Nm\Big)\Big\}\Big].
	\]
	On the event \(|N/m-\pi_1|\le\delta_m\), the bracket is \(\mu_m'\) at a point within \(\delta_m+1/m\) of \(\pi_1\), which tends to \(\mu_\infty'(\pi_1)\) by A4(iii). Off the event the bracket is at most \(m\), and by Hoeffding's inequality the event fails with probability \(o(m^{-j})\) for every \(j\). So the label term tends to \((1-\pi_1)\mu_\infty'(\pi_1)\) for \(y=1\), and in the same way to \(-\pi_1\mu_\infty'(\pi_1)\) for \(y=0\),
	\[
	m\{\mathbb E(R_T\mid Y_l=y)-\mathbb E(R_T)\}\to(y-\pi_1)\,\mu_\infty'(\pi_1).
	\]
	
	\noindent\emph{Step 3: The within-class term.}\quad
	The number \(N\) is independent of \(Z_l\), and given \(N\) and \(Y_l=y\) the training set has \(N+y\) cases, so the first term of Step~1 is
	\[
	\mathbb E_N\big\{\bar\phi_{y,m,N+y}(x)\big\}.
	\]
	On the event of Step~2, A4(ii) and the continuity of the limits at \(\pi_1\) put \(\bar\phi_{y,m,N+y}\) within \(o(1)\) of \(\bar\phi_y^\circ\) in \(L_2(P_y)\). Off it, \(|\bar\phi_{y,m,m_1}|\le m\). By Jensen's inequality the average over \(N\) converges in \(L_2(P_y)\) to \(\bar\phi_y^\circ\).
	
	\noindent\emph{Step 4: The variance.}\quad
	By Steps 1--3, \(\bar\phi(Z)\) converges in \(L_2(P)\) to
	\[
	\bar\phi_Y^\circ(X)+(Y-\pi_1)\,\mu_\infty'(\pi_1).
	\]
	The two parts are uncorrelated, since \(\bar\phi_y^\circ\) has mean zero within each class, so
	\[
	\Var\{\bar\phi(Z)\}\to\pi_1\sigma^2_{T\mid1}+\pi_0\sigma^2_{T\mid0}+\pi_1\pi_0\,\mu_\infty'(\pi_1)^2=\sigma_T^2.
	\]
\end{proof}

\subsection{Proof of Corollary~\ref{prop:pairwise_reduction} and of the statements after it}
\label{appendix:pairwise_kernel_mapping}

\paragraph{Setting.}
Write \(\hat\pi_y:=n_y/n\) for the class proportions of the sample and \(u_i:=\hat\pi_{Y_i}\tilde v_i\) for the validation weights rescaled to the unit-level scale, so that \(\sum_iu_i=1\). Let \(\mu_m\) be as in Section~\ref{subsec:pairwise_metrics}, a function of the case fraction of the training set. For a split \(r\) with \(m_{1r}\) cases in its training set, write \(g_{y,r}\) for the centered validation influence \(\ell_i^\star-\mu_m(m_{1r}/m)\) of a validated observation of class \(y\), and \(\bar\phi_{y,r}\) for the training influence \(\bar\phi_Y\) of a training observation of class \(y\), both at the training composition of split \(r\). Write \(g_y\) for the \(L_2\) limit of \(g_{y,r}\) at \(\pi_1\) under A4(ii), and \(\bar\phi_y^\circ\) for that of \(\bar\phi_{y,r}\), as in Section~\ref{appendix:auc_label_part}. Below, as in \(\sigma_A^2\) and \(\gamma_A\), the constants \(\sigma^2_{V\mid y}\), \(\sigma^2_{T\mid y}\), and \(\gamma_{0\mid y}\) are taken at these limits, the variances and covariance of \(g_y\) and \(\bar\phi_y^\circ\) under \(P_y\). Let \(\delta_n:=n^{-1/3}\). By A4(ii) and (iii) and the continuity of the limits at \(\pi_1\),
\[
\bar\varepsilon_n:=\sup_{\substack{y,\ m'\ge pn-2\\ |m_1'/m'-\pi_1|\le2\delta_n}}\big(\|g_{y,m',m_1'}-g_y\|_2+\|\bar\phi_{y,m',m_1'}-\bar\phi_y^\circ\|_2+|\mu_{m'}'(m_1'/m')-\mu_\infty'(\pi_1)|\big)
\]
tends to zero, with the subscript \(m',m_1'\) marking a training set of \(m'\) observations with \(m_1'\) cases. All statements below are on an event \(\mathcal E_n\) whose complement has probability \(o(n^{-j})\) for every \(j\). The estimate and its first-order terms are bounded, so the complement contributes \(o(n^{-1})\) to every variance below.

\begin{proof}[Proof of Corollary~\ref{prop:pairwise_reduction}]
	The design is fixed-size with \(p<1\) and independent of the labels, so given \(I\) the labels are i.i.d.\ Bernoulli\((\pi_1)\). Let \(\mathcal E_n\) be the event that every split has \(|m_{1r}/m-\pi_1|\le\delta_n\) and \(|n_{1r}/(n-m)-\pi_1|\le\delta_n\), with \(n_{yr}\) the number of validated observations of class \(y\) in split \(r\). By Hoeffding's inequality and a union bound over at most polynomially many splits, its complement has probability at most \(4S\exp\{-2\min(m,n-m)\delta_n^2\}\), uniformly in \(I\).
	
	\noindent\emph{Step 1: The first-order representation given \((I,Y)\).}\quad
	The proof of Theorem~\ref{thm:pairwise_identity} in Section~\ref{appendix:proof_auc_identity} does not use class stratification. Its single-observation component (Step~5 there), written split by split, is
	\[
	h_i(X_i)=\sum_ra_{ir}\,g_{Y_i,r}(X_i)+\sum_r\frac{s_rI_{ir}}{m}\,\bar\phi_{Y_i,r}(X_i),
	\qquad
	a_{ir}:=\frac{\mathbb I(i\in V_r)\,n_{1-Y_i,r}}{M_A},
	\]
	with \(s_r=n_{1r}n_{0r}/M_A\) the share of the validated pairs in split \(r\), so that \(\tilde v_i=\sum_ra_{ir}\) and \(w_i=\sum_rs_rI_{ir}/m\). By orthogonality of the components,
	\[
	\Var(\widehat R_D\mid I,Y)=\sum_i\Var(h_i\mid I,Y)+\sum_{k\ge2}\rho_n^{(k)}(I,Y).
	\]
	
	\noindent\emph{Step 2: The kernels at their limits.}\quad
	Let \(\tilde h_i:=\tilde v_i\,g_{Y_i}(X_i)+w_i\,\bar\phi^\circ_{Y_i}(X_i)\). On \(\mathcal E_n\), every split's kernels are within \(\bar\varepsilon_n\) of the limits, so
	\[
	\|h_i-\tilde h_i\|_2\le\bar\varepsilon_n(\tilde v_i+w_i).
	\]
	On \(\mathcal E_n\), \(n_{yr}\) and \(M_A/S\) are of order \(n\) and \(n^2\), so \(\tilde v_i\le C/n\) and \(w_i\le1/m\). The kernels have bounded norms, so
	\[
	\Big|\sum_i\Var(h_i\mid I,Y)-\sum_i\Var(\tilde h_i\mid I,Y)\Big|
	\le
	C\bar\varepsilon_n\sum_i(\tilde v_i+w_i)^2
	=
	o(n^{-1}),
	\]
	and
	\[
	\sum_i\Var(\tilde h_i\mid I,Y)
	=
	\sum_i\tilde v_i^2\sigma^2_{V\mid Y_i}+\sum_iw_i^2\sigma^2_{T\mid Y_i}+2\sum_i\tilde v_iw_i\gamma_{0\mid Y_i}.
	\]
	
	\noindent\emph{Step 3: The weights.}\quad
	Let \(v_i=N_i/M\) and \(w_i^{\mathrm u}=\sum_r|V_r|I_{ir}/(Mm)\) be the unit-level weights, which are functions of \(I\) alone. The design is fixed-size, so \(|V_r|=n-m\) and \(w_i^{\mathrm u}=\sum_rI_{ir}/(Sm)\). On \(\mathcal E_n\),
	\[
	n_{1r}n_{0r}=(n-m)^2\pi_1\pi_0\{1+O(\delta_n)\},
	\qquad
	M_A=S(n-m)^2\pi_1\pi_0\{1+O(\delta_n)\},
	\]
	uniformly over splits, so
	\[
	s_r=\frac1S\{1+O(\delta_n)\},
	\qquad
	w_i=w_i^{\mathrm u}\{1+O(\delta_n)\},
	\qquad
	\tilde v_i=\frac{v_i}{\pi_{Y_i}}\{1+O(\delta_n)\}.
	\]
	Since \(\sum_iv_i^2\), \(\sum_i(w_i^{\mathrm u})^2\), and \(\sum_iv_iw_i^{\mathrm u}\) are \(O(n^{-1})\), Step~2 gives, on \(\mathcal E_n\),
	\[
	\sum_i\Var(h_i\mid I,Y)
	=
	\sum_i\frac{v_i^2\sigma^2_{V\mid Y_i}}{\pi_{Y_i}^2}+\sum_i(w_i^{\mathrm u})^2\sigma^2_{T\mid Y_i}+2\sum_i\frac{v_iw_i^{\mathrm u}\gamma_{0\mid Y_i}}{\pi_{Y_i}}+o(n^{-1}).
	\]
	
	\noindent\emph{Step 4: The conditional mean.}\quad
	A validated pair of split \(r\) has expectation \(\mu_m(m_{1r}/m)\) given \((I,Y)\), so
	\[
	\mathbb E(\widehat R_D\mid I,Y)=\sum_rs_r\,\mu_m(m_{1r}/m).
	\]
	On \(\mathcal E_n\), by the mean value theorem,
	\[
	\mu_m(m_{1r}/m)=\mu_m(\pi_1)+\mu_\infty'(\pi_1)(m_{1r}/m-\pi_1)+e_r,
	\qquad
	|e_r|\le\bar\varepsilon_n|m_{1r}/m-\pi_1|.
	\]
	With \(\sum_rs_r=1\) and \(m_{1r}/m-\pi_1=m^{-1}\sum_{l\in T_r}(Y_l-\pi_1)\),
	\[
	\mathbb E(\widehat R_D\mid I,Y)
	=
	\mu_m(\pi_1)
	+\mu_\infty'(\pi_1)\sum_lw_l^{\mathrm u}(Y_l-\pi_1)
	+\mu_\infty'(\pi_1)\sum_r\Big(s_r-\frac1S\Big)(m_{1r}/m-\pi_1)
	+\sum_rs_re_r.
	\]
	On \(\mathcal E_n\) the third term is at most a constant times \(\delta_n^2=n^{-2/3}\). By Jensen's inequality,
	\[
	\Big(\sum_rs_re_r\Big)^2\le\bar\varepsilon_n^2\sum_rs_r(m_{1r}/m-\pi_1)^2,
	\qquad
	\mathbb E\Big\{\sum_rs_r(m_{1r}/m-\pi_1)^2\,\Big|\,I\Big\}\le\{1+O(\delta_n)\}\frac{\pi_1\pi_0}{m},
	\]
	so the last term has second moment \(o(n^{-1})\). The second term has variance \(\mu_\infty'(\pi_1)^2\pi_1\pi_0\sum_l(w_l^{\mathrm u})^2\) given \(I\), which is \(O(n^{-1})\). Hence
	\[
	\Var_Y\{\mathbb E(\widehat R_D\mid I,Y)\mid I\}=\pi_1\pi_0\,\mu_\infty'(\pi_1)^2\sum_l(w_l^{\mathrm u})^2+o(n^{-1}).
	\]
	
	\noindent\emph{Step 5: Average over the labels.}\quad
	Given \(I\) the labels are i.i.d., and \(v_i\) and \(w_i^{\mathrm u}\) do not depend on them, so averaging Step~3 over \(Y\),
	\[
	\mathbb E_Y\Big\{\sum_i\Var(h_i\mid I,Y)\,\Big|\,I\Big\}
	=
	\sigma_A^2\sum_iv_i^2+(\pi_1\sigma^2_{T\mid1}+\pi_0\sigma^2_{T\mid0})\sum_i(w_i^{\mathrm u})^2+2\gamma_A\sum_iv_iw_i^{\mathrm u}+o(n^{-1}).
	\]
	By the law of total variance given \(I\), Steps 1 and 4, and the display for \(\sigma_T^2\) in Section~\ref{subsec:pairwise_metrics} (Section~\ref{appendix:auc_label_part}),
	\[
	\Var(\widehat R_D\mid I)
	=
	\sigma_A^2\sum_iv_i^2+\sigma_T^2\sum_i(w_i^{\mathrm u})^2+2\gamma_A\sum_iv_iw_i^{\mathrm u}
	+\mathbb E_Y\Big\{\sum_{k\ge2}\rho_n^{(k)}(I,Y)\,\Big|\,I\Big\}+o(n^{-1}),
	\]
	with the \(o(n^{-1})\) uniform in \(I\). By A4(i) the average over \(I\) of the fourth term is \(o(n^{-1})\). The fourth term is nonnegative, and on \(\mathcal E_n\) the \(o(n^{-1})\) is uniform, so the remainder of the first display of the corollary has expected absolute value \(o(n^{-1})\).
	
	\noindent\emph{Step 6: The conditional mean given \(I\).}\quad
	Averaging Step~4 over \(Y\), the second term has mean zero, the third is \(O(n^{-2/3})\), and the fourth is \(o(n^{-1/2})\), all uniformly in \(I\). So \(\mathbb E(\widehat R_D\mid I)=\mu_m(\pi_1)+o(n^{-1/2})\) uniformly, and
	\[
	\Var_I\{\mathbb E(\widehat R_D\mid I)\}=o(n^{-1}).
	\]
\end{proof}

This proves the statement of Section~\ref{subsec:design_comparison} on the penalty of a design independent of the labels. For such a design, the matrix with diagonal \((\sigma_A^2,\sigma_T^2)\) and off-diagonal \(\gamma_A\) is the covariance matrix of \((\{\ell_i^\star-\mu_m(\pi_1)\}/\pi_{Y_i},\bar\phi(Z_i))\), with \(\bar\phi\) not conditioned on the labels, because the first coordinate has mean zero within each class. It is therefore positive semidefinite, and by the first display of the corollary \eqref{eq:imbalance_decomposition} holds with it in place of \(\Sigma\), up to a remainder of expected absolute value \(o(n^{-1})\). For \(K\)-fold, \(v_i=w_i=1/n\), so the penalty is zero.

\paragraph{Stratified \(K\)-fold.}
This proves the stratified \(K\)-fold statement of Corollary~\ref{prop:pairwise_reduction}. Each fold contains \(n_{yr}\) observations of class \(y\), with \(n_{yr}\) equal to \(\lfloor n_y/K\rfloor\) or \(\lceil n_y/K\rceil\). The folds are assigned at random within each class, and the labels are i.i.d. Let \(\mathcal E_n:=\{|\hat\pi_1-\pi_1|\le\delta_n\}\), whose complement has probability \(o(n^{-j})\) for every \(j\). The one-observation condition of the corollary is that adding one case or one control to a training set of size at least \(pn-2\) with class fraction within \(2\delta_n\) of \(\pi_1\) changes the expected AUC by at most \(\theta_nn^{-1/2}\), with \(\theta_n\to0\).

\begin{proof}
	\noindent\emph{Step 1: The weights.}\quad
	Write \(n_{yr}=n_y/K+\delta_{yr}\), with \(|\delta_{yr}|<1\) and \(\sum_r\delta_{yr}=0\). Then
	\[
	M_A=\sum_rn_{1r}n_{0r}=\frac{n_1n_0}{K}+\sum_r\delta_{1r}\delta_{0r}=\frac{n_1n_0}{K}\{1+O(n^{-2})\}.
	\]
	For \(i\) of class \(y\) in fold \(r\), \(\tilde v_i=n_{1-y,r}/M_A=1/n_y+O(n^{-2})\), so \(u_i=1/n+O(n^{-2})\). Every split carries share \(s_r=n_{1r}n_{0r}/M_A=1/K+O(n^{-1})\) of the pairs and has \(|T_r|=pn+O(1)\), so
	\[
	w_i=\sum_{r'\ne r}\frac{s_{r'}}{|T_{r'}|}=\frac1n+O(n^{-2}).
	\]
	When \(K\) divides both class counts, every \(\delta_{yr}\) is zero and \(\tilde v_i=1/n_{Y_i}\), \(u_i=w_i=1/n\) exactly.
	
	\noindent\emph{Step 2: The conditional variance.}\quad
	Fold \(r\) trains on \(n_y-n_{yr}\) observations of class \(y\), with class fraction \(\hat\pi_1+O(n^{-1})\) and size \(pn+O(1)\). On \(\mathcal E_n\) its kernels are within \(\bar\varepsilon_n\) of the limits, as in Step~2 of the proof of Corollary~\ref{prop:pairwise_reduction}, with the same bounds on the weights. Steps 1--2 of that proof and Step~1 here give, on \(\mathcal E_n\),
	\[
	\sum_i\Var(h_i\mid I,Y)
	=
	\frac1n\Big\{\frac{\sigma^2_{V\mid1}}{\hat\pi_1}+\frac{\sigma^2_{V\mid0}}{\hat\pi_0}+\hat\pi_1\sigma^2_{T\mid1}+\hat\pi_0\sigma^2_{T\mid0}+2\gamma_{0\mid1}+2\gamma_{0\mid0}\Big\}+o(n^{-1}).
	\]
	Since \(\hat\pi_y\) is within \(\delta_n\) of \(\pi_y\) on \(\mathcal E_n\),
	\[
	\mathbb E_{I,Y}\Big\{\sum_i\Var(h_i\mid I,Y)\Big\}
	=
	\frac{\sigma_A^2+\pi_1\sigma^2_{T\mid1}+\pi_0\sigma^2_{T\mid0}+2\gamma_A}{n}+o(n^{-1}).
	\]
	
	\noindent\emph{Step 3: The conditional mean.}\quad
	Write \(\mu_{(r)}\) for the expected AUC of a model fitted to the training composition of fold \(r\), so that
	\[
	\mathbb E(\widehat R_D\mid I,Y)=\sum_rs_r\mu_{(r)}.
	\]
	When \(K\) divides both class counts, every fold trains on \(m=pn\) observations with class fraction \(\hat\pi_1\), and this is \(\mu_m(\hat\pi_1)\) exactly. In general, let \(m_\star:=n-\lceil n/K\rceil-1\). The composition \((n_1-\lceil n_1/K\rceil,\,n_0-\lceil n_0/K\rceil)\) has size \(m_\star\) or \(m_\star+1\), and every fold's composition differs from it by at most one case and one control. Removing one observation where needed gives a composition of size \(m_\star\) with \(\tilde m_1\) cases, where \(\tilde m_1/m_\star=\hat\pi_1+O(n^{-1})\). By the one-observation condition,
	\[
	\mathbb E(\widehat R_D\mid I,Y)=\mu_{m_\star}(\tilde m_1/m_\star)+b_n,
	\qquad
	|b_n|\le3\theta_nn^{-1/2}.
	\]
	The training size \(m_\star\) does not depend on \(Y\). On \(\mathcal E_n\), as in Step~4 of the proof of Corollary~\ref{prop:pairwise_reduction},
	\[
	\mu_{m_\star}(\tilde m_1/m_\star)=\mu_{m_\star}(\pi_1)+\mu_\infty'(\pi_1)(\tilde m_1/m_\star-\pi_1)+e,
	\qquad
	|e|\le\bar\varepsilon_n|\tilde m_1/m_\star-\pi_1|,
	\]
	and \(\Var(\tilde m_1/m_\star)=\Var(\hat\pi_1)+O(n^{-3/2})=\pi_1\pi_0/n+o(n^{-1})\). Since \(\mu_{m_\star}(\tilde m_1/m_\star)\) has standard deviation \(O(n^{-1/2})\) and \(b_n\) and \(e\) are \(o(n^{-1/2})\) in \(L_2\),
	\[
	\Var_{I,Y}\{\mathbb E(\widehat R_D\mid I,Y)\}=\frac{\pi_1\pi_0\,\mu_\infty'(\pi_1)^2}{n}+o(n^{-1}).
	\]
	
	\noindent\emph{Step 4: The total.}\quad
	Adding Steps 2 and 3, with A4(i) for the components on two or more observations and the display for \(\sigma_T^2\) of Section~\ref{appendix:auc_label_part},
	\[
	\Var(\widehat R_D)=\frac{\sigma_A^2+\sigma_T^2+2\gamma_A}{n}+o(n^{-1}).
	\]
\end{proof}

\section{Proofs and derivations for Section~\ref{sec:implications}}
\label{appendix:proofs_sec3}

\subsection{Derivation of the role moments \eqref{eq:role_moments}}
\label{appendix:proof_role_moments}

\begin{proof}
	\noindent\emph{Step 1: $K$-fold with $R$ repetitions.}\quad
	With equal folds, every observation is validated $R$ times and appears in $R(K-1)$ training sets of size $n(K-1)/K$, with $M=Rn$ and $s_r=1/(RK)$. Hence
	\[
	v_i=\frac{R}{Rn}=\frac1n,
	\qquad
	w_i=R(K-1)\cdot\frac{1}{RK}\cdot\frac{K}{n(K-1)}=\frac1n,
	\]
	and the three moments equal $1/n$ exactly.
	
	\noindent\emph{Step 2: Monte Carlo cross-validation.}\quad
	With independent splits of validation size $nq$, each split validates observation $i$ with probability $q$ independently of the others, so $N_i\sim\mathrm{Binomial}(S,q)$ exactly, and $M=Snq$ is constant. With $v_i=N_i/(Snq)$ and $w_i=(S-N_i)/(Snp)$,
	\[
	\Gamma_D=\frac{n\,\mathbb E(N_i^2)}{S^2n^2q^2}=\frac{Spq+S^2q^2}{S^2nq^2}=\frac1n\Big(1+\frac{p}{qS}\Big),
	\qquad
	\Lambda_D=\frac{n\,\mathbb E\{(S-N_i)^2\}}{S^2n^2p^2}=\frac1n\Big(1+\frac{q}{pS}\Big),
	\]
	\[
	\Omega_D=\frac{n\,\mathbb E\{N_i(S-N_i)\}}{S^2n^2pq}=\frac{S^2q-S^2q^2-Spq}{S^2npq}=\frac1n\Big(1-\frac1S\Big).
	\]
	
	\noindent\emph{Step 3: Bootstrap cross-validation.}\quad
	With independent resamples, $\tilde I_{ir}\sim\mathrm{Binomial}(n,1/n)$, independent across $r$, with
	\[
	\mathbb E(\tilde I_{ir})=1,\qquad \mathbb E(\tilde I_{ir}^2)=2-\frac1n.
	\]
	The count $N_i=\sum_r\mathbb I(\tilde I_{ir}=0)$ is $\mathrm{Binomial}(S,q_n)$ with $q_n=(1-1/n)^n\to e^{-1}$, and $M=\sum_r|V_r|$ with $\Var(|V_r|)=O(n)$, so
	\[
	M=Snq_n\{1+O_P(n^{-1/2})\},
	\qquad
	\frac{|V_r|}{M}=\frac1S\{1+O_P(n^{-1/2})\}.
	\]
	The relative errors are bounded, so they converge in $L_1$ and contribute $o(n^{-1})$ to moments of order $n^{-1}$. Then
	\[
	\Gamma_D=\frac{n\,\mathbb E(N_i^2)}{(Snq_n)^2}+o(n^{-1})=\frac1n\Big(1+\frac{p}{qS}\Big)+o(n^{-1}),
	\]
	\[
	\Lambda_D=\frac{n\,\mathbb E\big\{(\sum_r\tilde I_{ir})^2\big\}}{S^2n^2}+o(n^{-1})=\frac{S(2-1/n)+S(S-1)}{S^2n}+o(n^{-1})=\frac1n\Big(1+\frac1S\Big)+o(n^{-1}),
	\]
	\[
	\Omega_D=\frac{n\,\mathbb E\big(N_i\sum_{r'}\tilde I_{ir'}\big)}{Snq_n\cdot Sn}+o(n^{-1})=\frac{S(S-1)q_n}{S^2nq_n}+o(n^{-1})=\frac1n\Big(1-\frac1S\Big)+o(n^{-1}),
	\]
	using $\mathbb E\{\mathbb I(\tilde I_{ir}=0)\tilde I_{ir'}\}=q_n$ for $r\ne r'$ and $0$ for $r=r'$.
\end{proof}

\subsection{Derivation of \eqref{eq:nadeau_bengio}}
\label{appendix:proof_nadeau_bengio}

Let $D$ be Monte Carlo cross-validation with $S$ splits at training fraction $p$, under A1 and A2(i), and write $q=1-p$.

\begin{proof}
	\noindent\emph{Step 1: What $\hat\sigma^2$ estimates.}\quad
	Given the data the $S$ split estimates are independent and identically distributed, so $\mathbb E(\hat\sigma^2\mid\mathcal D_n)=\Var_I(\widehat R_1\mid\mathcal D_n)$, and $\mathbb E(\hat\sigma^2)$ is the split noise of a single split. By \eqref{eq:floor_split} and Step~4 of Section~\ref{appendix:proof_split_lotv}, its parts above first order are $o(n^{-1})$ under A2(i), and its first-order part is $Q_D$ at $S=1$, so
	\[
	\mathbb E(\hat\sigma^2)=\frac1n\Big(\frac pq\sigma_V^2+\frac qp\sigma_T^2-2\gamma_0\Big)+o(n^{-1}).
	\]
	Multiplying by $1/S+q/p$ gives the expected value of the estimator.
	
	\noindent\emph{Step 2: The error of the target.}\quad
	The variance of Monte Carlo cross-validation is $\mathcal V/n+Q_D+o(n^{-1})$, where $Q_D$ is its term in $1/(nS)$. The target therefore exceeds it by
	\[
	\frac1n\Big(\frac pq\sigma_V^2+\frac qp\sigma_T^2-2\gamma_0\Big)\frac qp-\frac{\mathcal V}{n}
	=
	\frac1n\Big\{\sigma_T^2\Big(\frac{q^2}{p^2}-1\Big)-2\gamma_0\Big(\frac qp+1\Big)\Big\}.
	\]
	
	\noindent\emph{Step 3: Simplify with $p+q=1$.}\quad
	The two coefficients are
	\[
	\frac{q^2}{p^2}-1=\frac{(q-p)(q+p)}{p^2}=\frac{q-p}{p^2},
	\qquad
	\frac qp+1=\frac1p,
	\]
	so the error is
	\[
	-\frac{(2p-1)\sigma_T^2+2p\gamma_0}{p^2n},
	\]
	which is zero if and only if $2p\gamma_0=(q-p)\sigma_T^2$. At $p=q$ the error is $-4\gamma_0/n$.
	
	\noindent\emph{Step 4: The same condition as a correlation.}\quad
	By \eqref{eq:lotv_data}, two independent splits have covariance equal to the sampling variance, which is $\mathcal V/n+o(n^{-1})$ under A2(i). One split has variance $\mathcal V/n$ plus $Q_D$ at $S=1$,
	\[
	\frac{\sigma_V^2+\sigma_T^2+2\gamma_0}{n}+\frac1n\Big(\frac pq\sigma_V^2+\frac qp\sigma_T^2-2\gamma_0\Big)=\frac1n\Big(\frac{\sigma_V^2}{q}+\frac{\sigma_T^2}{p}\Big),
	\]
	up to $o(n^{-1})$. So at first order the correlation of two split estimates is
	\[
	\rho_S=\frac{\sigma_V^2+\sigma_T^2+2\gamma_0}{\sigma_V^2/q+\sigma_T^2/p}.
	\]
	For split estimates with common correlation $\rho_S$,
	\[
	\hat\sigma^2\Big(\frac1S+\frac{\rho_S}{1-\rho_S}\Big)
	\]
	is unbiased for the variance of their average, and the factor $q/p$ of \citet{nadeau2003} is $\rho_S/(1-\rho_S)$ at $\rho_S=q$. Setting $\rho_S=q$ gives
	\[
	\sigma_T^2\Big(1-\frac qp\Big)+2\gamma_0=0,
	\]
	which is the condition of Step~3.
\end{proof}

\subsection{Derivation of the sampling variance and the split noise by order}
\label{appendix:proof_split_lotv}

This derives \eqref{eq:floor_split}, the split of Section~\ref{subsec:split_noise} into the sampling variance and the split noise, order by order.

\begin{proof}
	
	\noindent\emph{Step 1: Decompose the variance order by order.}\quad
	Let the design be fixed-size, satisfy condition (iii), and use pooled aggregation. Fix a realized inclusion structure $I$, and write the conditional Hoeffding decomposition as
	\[
	\widehat R_D-R_{p,n}
	=
	\sum_{\varnothing\ne\mathcal S\subseteq\{1,\ldots,n\}}
	h_{\mathcal S,I}.
	\]
	The components are mutually orthogonal. Since $I$ is independent of the data, averaging over $I$ commutes with conditioning on any subset of the observations. Therefore
	\[
	\bar h_{\mathcal S}
	:=
	\mathbb E_I(h_{\mathcal S,I})
	\]
	is the corresponding Hoeffding component of
	$\mathbb E_I(\widehat R_D\mid\mathcal D_n)-R_{p,n}$.
	
	It follows that
	\[
	\Var_{\mathcal D}\!\left\{
	\mathbb E_I(\widehat R_D\mid\mathcal D_n)
	\right\}
	=
	\sum_{\mathcal S}\|\bar h_{\mathcal S}\|_2^2,
	\]
	whereas
	\[
	\Var(\widehat R_D)
	=
	\sum_{\mathcal S}
	\mathbb E_I(\|h_{\mathcal S,I}\|_2^2).
	\]
	Subtracting gives the split noise:
	\begin{equation}
		\label{eq:split_noise_orderwise}
		\mathbb E_{\mathcal D}\!\left\{
		\Var_I(\widehat R_D\mid\mathcal D_n)
		\right\}
		=
		\sum_{\mathcal S}
		\left[
		\mathbb E_I(\|h_{\mathcal S,I}\|_2^2)
		-
		\|\bar h_{\mathcal S}\|_2^2
		\right].
	\end{equation}
	Each bracket is nonnegative by Jensen's inequality. Thus split noise can be separated exactly by Hoeffding order.
	
	\noindent\emph{Step 2: Single-observation terms.}\quad
	By Theorem~\ref{thm:dual_role}, $h_{\{i\},I}=v_i(\ell_i^\star-R_{p,n})+w_i\bar\phi(Z_i)$. Exchangeability over observation indices and $\sum_i v_i=\sum_iw_i=1$ give
	\[
	\mathbb E_I(v_i)
	=
	\mathbb E_I(w_i)
	=
	\frac1n,
	\qquad\text{so}\qquad
	\bar h_{\{i\}}
	=
	\frac1nH(Z_i),
	\]
	and the first-order part of the sampling variance is
	\[
	\sum_i\|\bar h_{\{i\}}\|_2^2
	=
	\frac{\Var\{H(Z)\}}{n}
	=
	\frac{\mathcal V}{n}.
	\]
	The first-order part of \eqref{eq:split_noise_orderwise} is the first-order term of Theorem~\ref{thm:variance_identity} minus $\mathcal V/n$, which is $Q_D$ by \eqref{eq:imbalance_decomposition}.
	
	\noindent\emph{Step 3: Order $k$.}\quad
	By \eqref{eq:order_k_kernel}, for a set $\mathcal S$ of $k$ observations,
	\[
	h_{\mathcal S,I}=\sum_{i\in\mathcal S}c^{(k)}_{i;\mathcal S\setminus\{i\}}\,\phi^{(k)}(Z_i;Z_{\mathcal S\setminus\{i\}})+d^{(k)}_{\mathcal S}\,\chi^{(k)}(Z_{\mathcal S}).
	\]
	Under condition (iii) the distribution of $I$ is invariant under permutations of the observation indices, so the marginal distribution of each training set $T_r$ is invariant under permutations of the subsets of size $m$, and $T_r$ is uniformly distributed over them. Under pooled aggregation, $M=S(n-m)$, so
	\[
	\begin{aligned}
		\mathbb E_I\big(c^{(k)}_{i;\mathcal J}\big)
		&=
		\frac{1}{M m^{k-1}}\sum_r\mathbb P_D(i\in V_r,\ \mathcal J\subseteq T_r)
		=
		\frac{1}{(n-m)m^{k-1}}\binom{n-k}{m-k+1}\Big/\binom nm,
	\end{aligned}
	\]
	and, since $\sum_r|V_r|/M=1$,
	\[
	\mathbb E_I\big(d^{(k)}_{\mathcal S}\big)
	=
	\frac{1}{m^{k}}\sum_r\frac{|V_r|}{M}\,\mathbb P_D(\mathcal S\subseteq T_r)
	=
	\frac{1}{m^{k}}\binom{n-k}{m-k}\Big/\binom nm,
	\]
	the same for every index set. Write $\bar c^{(k)}$ and $\bar d^{(k)}$ for these averages. Then $\bar h_{\mathcal S}$ is $h_{\mathcal S,I}$ with $\bar c^{(k)}$ and $\bar d^{(k)}$ in place of the weights, so the order-$k$ part of the sampling variance, $\sum_{|\mathcal S|=k}\|\bar h_{\mathcal S}\|_2^2$, is the order-$k$ term of Theorem~\ref{thm:variance_identity} with every weight replaced by its average. It depends on the design only through $n$ and $m$. The order-$k$ part of \eqref{eq:split_noise_orderwise} is the order-$k$ term of Theorem~\ref{thm:variance_identity} minus this quantity. It is nonnegative by Jensen's inequality, and it is zero when the weights of order $k$ do not vary with $I$, since then $h_{\mathcal S,I}=\bar h_{\mathcal S}$ for every $\mathcal S$.
	
	At $k=1$, $\bar c^{(1)}=\bar d^{(1)}=1/n$, which is Step 2. At $k=2$, $\bar c:=\bar c^{(2)}=1/\{n(n-1)\}$ and $\bar d:=\bar d^{(2)}=(1-1/m)/\{n(n-1)\}$. With
	\[
	\Theta^\circ:=n(n-1)\bar c^2,
	\qquad
	\Psi^\circ:=\binom n2\bar d^2,
	\qquad
	\Psi'^{\circ}:=n(n-1)\bar c\,\bar d,
	\]
	the pair part of the sampling variance is
	\[
	\Pi_n
	=
	\sum_{i<j}\|\bar h_{\{i,j\}}\|_2^2
	=
	(\tau_n^2+\kappa_n)\Theta^\circ
	+
	\sigma_{\chi,n}^2\Psi^\circ
	+
	2\xi_n\Psi'^{\circ},
	\]
	the pair term of Theorem~\ref{thm:variance_identity} is $\tau_n^2\Theta_D+\kappa_n\Theta_D'+\sigma_{\chi,n}^2\Psi_D+2\xi_n\Psi_D'$, and their difference is the second-order split noise
	\[
	\mathcal N_D^{(2)}
	=
	\tau_n^2(\Theta_D-\Theta^\circ)
	+
	\kappa_n(\Theta_D'-\Theta^\circ)
	+
	\sigma_{\chi,n}^2(\Psi_D-\Psi^\circ)
	+
	2\xi_n(\Psi_D'-\Psi'^{\circ}),
	\]
	nonnegative as a whole.
	
	\noindent\emph{Step 4: Higher-order terms.}\quad
	By Step~3 and \eqref{eq:split_noise_orderwise}, the parts of order three and higher are
	\[
	\sum_{k\ge3}\Pi_n^{(k)}
	=
	\sum_{|\mathcal S|\ge3}\|\bar h_{\mathcal S}\|_2^2,
	\qquad
	\sum_{k\ge3}\mathcal N_D^{(k)}
	=
	\sum_{|\mathcal S|\ge3}
	\left[
	\mathbb E_I(\|h_{\mathcal S,I}\|_2^2)
	-
	\|\bar h_{\mathcal S}\|_2^2
	\right].
	\]
	Both are nonnegative, and they sum to \(\mathbb E_I\{\rho_n^{(3+)}(I)\}\). Combining Steps 2--4 gives
	\[
	\Var_{\mathcal D}\!\left\{
	\mathbb E_I(\widehat R_D\mid\mathcal D_n)
	\right\}
	=
	\frac{\mathcal V}{n}
	+
	\Pi_n
	+
	\sum_{k\ge3}\Pi_n^{(k)},
	\qquad
	\mathbb E_{\mathcal D}\!\left\{
	\Var_I(\widehat R_D\mid\mathcal D_n)
	\right\}
	=
	Q_D
	+
	\mathcal N_D^{(2)}
	+
	\sum_{k\ge3}\mathcal N_D^{(k)},
	\]
	which are the two lines of \eqref{eq:floor_split}. By Step~3, \(\Pi_n^{(k)}\) and \(\mathcal N_D^{(k)}\) are nonnegative and add to \(\mathbb E_I\{\rho_n^{(k)}(I)\}\) at every \(k\ge2\), so the parts of the sampling variance and of the split noise above first order are each at most \(\mathbb E_I\{\rho_n^{(2+)}(I)\}\). Under A2(i) both are therefore \(o(n^{-1})\). For \(K\)-fold, \(Q_D=0\) by \eqref{eq:QD_designs}, so its sampling variance is \(\mathcal V/n+o(n^{-1})\) and the split noise of one partition is \(o(n^{-1})\).
	
	\noindent\emph{Step 5: Leave-one-out.}\quad
	Leave-one-out has a single inclusion structure, so $h_{\mathcal S,I}=\bar h_{\mathcal S}$ for every $\mathcal S$, every bracket in \eqref{eq:split_noise_orderwise} is zero, and the split noise is zero at every order.
	
\end{proof}

\subsection{The sampling variance and the split noise for the within-fold AUC}
\label{appendix:proof_split_lotv_auc}

This section proves the decomposition of Section~\ref{subsec:split_noise} for the within-fold AUC. The labels \(Y\) have their law conditioned on an event of the class counts on which every class has at least two observations, for example \(n_y\ge2\) or \(\{n_1\ge K,\,n_0\ge K\}\). The design is class-stratified and fixed-size. Every training set has \(m_1\) cases and \(m_0\) controls, \(m=m_1+m_0\), and \((m_1,m_0)\) may depend on the class counts \((n_1,n_0)\). The map from \((n_1,n_0)\) to \((m_1,m_0)\) is the design's \emph{training-composition rule}. Every validation set is the complement of its training set and contains at least one case and one control, so \(1\le n_y-m_y\). Given the labels \(Y\), the inclusion structure \(I\) is independent of the covariates, and permuting the cases among themselves, or the controls among themselves, leaves its distribution unchanged. The model is fit symmetrically in its training observations, and aggregation is pooled. We write \(R_{p,n}(Y)\), as in Section~\ref{subsec:cv_design}, for the expected AUC of a model fitted to \(m_1\) cases and \(m_0\) controls.

\begin{proof}
	\noindent\emph{Step 1: Separate the labels.}\quad
	For a class-stratified fixed-size design, \(\mathbb E(\widehat R_D\mid I,Y)=R_{p,n}(Y)\) for every realized \(I\) (Section~\ref{subsec:pairwise_metrics}). Write \(\mathbf X=(X_1,\dots,X_n)\), so that the data are \((\mathbf X,Y)\). Since
	\[
	\mathbb E_{\mathbf X}\{\mathbb E_I(\widehat R_D\mid\mathbf X,Y)\mid Y\}=\mathbb E(\widehat R_D\mid Y)=R_{p,n}(Y),
	\]
	the two terms of \eqref{eq:lotv_data} are
	\[
	\Var_{\mathbf X,Y}\{\mathbb E_I(\widehat R_D\mid\mathbf X,Y)\}
	=
	\Var\{R_{p,n}(Y)\}
	+
	\mathbb E_Y\big[\Var_{\mathbf X}\{\mathbb E_I(\widehat R_D\mid\mathbf X,Y)\mid Y\}\big]
	\]
	and
	\[
	\mathbb E_{\mathbf X,Y}\{\Var_I(\widehat R_D\mid\mathbf X,Y)\}
	=
	\mathbb E_Y\big[\mathbb E_{\mathbf X}\{\Var_I(\widehat R_D\mid\mathbf X,Y)\mid Y\}\big].
	\]
	The term \(\Var\{R_{p,n}(Y)\}\) is the variance across label vectors of the expected AUC at the training composition. It depends on the design only through its training-composition rule.
	
	\noindent\emph{Step 2: Decompose given the labels.}\quad
	Fix \(Y\). Given \((I,Y)\), \(\widehat R_D-R_{p,n}(Y)=\sum_{\varnothing\ne\mathcal S}h_{\mathcal S,I}\), with the components \eqref{eq:auc_components} in the covariates under their product law given \(Y\), and with
	\[
	h_{\{i\},I}=\tilde v_i\,g_{Y_i,m}(X_i)+w_i\,\bar\phi_{Y,m}(X_i)
	\]
	at first order (Section~\ref{appendix:proof_auc_identity}). The components are mutually orthogonal given \((I,Y)\). Write \(\|\cdot\|\) for the norm of \(L_2\) under the law of \(\mathbf X\) given \(Y\). Since \(I\) is independent of \(\mathbf X\) given \(Y\), averaging over \(I\) given \(Y\) commutes with conditioning on any subset of the covariates. Therefore
	\[
	\bar h_{\mathcal S}:=\mathbb E_I(h_{\mathcal S,I}\mid Y)
	\]
	is the Hoeffding component of \(\mathbb E_I(\widehat R_D\mid\mathbf X,Y)-R_{p,n}(Y)\) on \(\mathcal S\). Because \(\mathbb E(\widehat R_D\mid I,Y)\) does not vary with \(I\),
	\[
	\Var(\widehat R_D\mid Y)=\mathbb E_I\{\Var(\widehat R_D\mid I,Y)\mid Y\}=\sum_{\mathcal S}\mathbb E_I\big(\|h_{\mathcal S,I}\|^2\mid Y\big).
	\]
	As in Step~1 of Section~\ref{appendix:proof_split_lotv}, it follows that
	\[
	\Var_{\mathbf X}\{\mathbb E_I(\widehat R_D\mid\mathbf X,Y)\mid Y\}=\sum_{\mathcal S}\|\bar h_{\mathcal S}\|^2
	\]
	and
	\[
	\mathbb E_{\mathbf X}\{\Var_I(\widehat R_D\mid\mathbf X,Y)\mid Y\}=\sum_{\mathcal S}\big[\mathbb E_I(\|h_{\mathcal S,I}\|^2\mid Y)-\|\bar h_{\mathcal S}\|^2\big].
	\]
	
	\noindent\emph{Step 3: The averaged weights.}\quad
	The marginal law of a training set \(T_r\) given \(Y\) is invariant under permutations of the cases among themselves and of the controls among themselves. These permutations act transitively on the sets with \(m_1\) cases and \(m_0\) controls, so \(T_r\) is uniformly distributed over them. Its cases are therefore a uniformly drawn \(m_1\)-subset of the \(n_1\) cases and, independently, its controls a uniformly drawn \(m_0\)-subset of the \(n_0\) controls. Every validation set is the complement of its training set, so \(|V_r^y|=n_y-m_y\) and \(M_A=S(n_1-m_1)(n_0-m_0)\) for a design with \(S\) splits. Each weight of order \(k\) of Section~\ref{appendix:proof_auc_identity} counts the splits in which some members of its set are validated and the others train. With \(|V_r^y|\) and \(M_A\) fixed, its average over \(I\) given \(Y\) is the probability of that event under this law of \(T_r\), divided by \((n_1-m_1)(n_0-m_0)m^{k-2}\) for \(e^{(k)}\), by \((n_y-m_y)m^{k-1}\) for \(c^{(k)}_{i;\mathcal J}\) with \(i\) of class \(y\), and by \(m^k\) for \(d^{(k)}\). Each average therefore depends on its set only through its numbers of cases and controls, and on the design only through \((n_1,n_0,m_1,m_0)\). At \(k=1\), for \(i\) of class \(y\), this gives
	\[
	\mathbb E_I(\tilde v_i\mid Y)=\frac1{n_y},
	\qquad
	\mathbb E_I(w_i\mid Y)=\frac{m_y}{n_y\,m}.
	\]
	
	\noindent\emph{Step 4: Sampling variance and split noise by order.}\quad
	By Steps 2 and 3, \(\bar h_{\mathcal S}\) is \eqref{eq:auc_components} with every weight replaced by its average given \(Y\). The kernels of \(\mathcal S\) are the same in every split, and they depend on \(\mathcal S\) only through its numbers of cases and controls and on the design only through \((m_1,m_0)\) (Section~\ref{appendix:proof_auc_identity}). Hence \(\|\bar h_{\mathcal S}\|^2\) depends on \(\mathcal S\) only through its numbers of cases and controls, and on the design only through \((n_1,n_0,m_1,m_0)\). With Step 1, the sampling variance
	\begin{equation}
		\label{eq:auc_floor}
		\Var_{\mathrm{samp}}(D):=\Var\{R_{p,n}(Y)\}+\mathbb E_Y\Big(\sum_{\mathcal S}\|\bar h_{\mathcal S}\|^2\Big)
	\end{equation}
	depends on the design only through \(n\) and its training-composition rule. The order-\(k\) part of the split noise is
	\[
	\mathbb E_Y\Big\{\sum_{|\mathcal S|=k}\big[\mathbb E_I(\|h_{\mathcal S,I}\|^2\mid Y)-\|\bar h_{\mathcal S}\|^2\big]\Big\}.
	\]
	Given \(Y\), \(h_{\mathcal S,I}\) is linear in the weights of \(\mathcal S\) with kernels that do not depend on \(I\), so \(\|h_{\mathcal S,I}\|^2\) is a positive semidefinite quadratic form in those weights. Each bracket is therefore nonnegative by Jensen's inequality, and it is zero when the weights of \(\mathcal S\) do not vary with \(I\) given \(Y\).
	
	At first order, Step 3 gives
	\[
	\bar h_{\{i\}}=\frac{g_{y,m}(X_i)+(m_y/m)\bar\phi_{y,m}(X_i)}{n_y}
	\]
	for \(i\) of class \(y\), with \(g_{y,m}\) and \(\bar\phi_{y,m}\) the first-order kernels at the training composition. Let \(\mathcal Q_{y,m}(m_y/m)\) be the variance under \(P_y\) of \(g_{y,m}+(m_y/m)\bar\phi_{y,m}\), which is \(\sigma_{V\mid y}^2+2(m_y/m)\gamma_{0\mid y}+(m_y/m)^2\sigma_{T\mid y}^2\). The \(n_y\) observations of class \(y\) contribute \(\mathcal Q_{y,m}(m_y/m)/n_y\), so the first-order part of the sampling variance is
	\[
	\Var\{R_{p,n}(Y)\}+\mathbb E_Y\Big\{\frac{\mathcal Q_{1,m}(m_1/m)}{n_1}+\frac{\mathcal Q_{0,m}(m_0/m)}{n_0}\Big\}
	=\mathcal W_{1,n}(D)
	\]
	of \eqref{eq:auc_first_floor}. When the weights of order one do not vary with \(I\) given \(Y\), as for stratified \(K\)-fold with \(K\) dividing both class counts and for leave-pair-out, \(\mathcal W_{1,n}(D)\) is also the first-order part of the variance.
	
	\noindent\emph{Step 5: Designs whose weights do not vary.}\quad
	If the weights of every order do not vary with \(I\) given \(Y\), then \(h_{\mathcal S,I}=\bar h_{\mathcal S}\) for every \(\mathcal S\), the split noise is zero at every order, and the variance is the sampling variance. Leave-pair-out is such a design, with \(m_y=n_y-1\), when its splits are taken in random order, which changes no weight. So is exhaustive validation at any training-composition rule, which uses every training set with the prescribed numbers of cases and controls once, in random order. Its variance is therefore the sampling variance of every design with that rule.
\end{proof}

\subsection{Proof of Theorem~\ref{thm:balanced_optimal}, \eqref{eq:split_noise_counts}, and the cost of pair balance}
\label{appendix:proof_order_balance}

\begin{proof}
	\noindent\emph{Step 1: Balance at order $k$ implies balance at order $k-1$, for $2\le k\le m+1$.}\quad
	Fix $i$ and a set $\mathcal J'$ of $k-2$ observations not containing $i$. In a split with $i\in V_r$ and $\mathcal J'\subseteq T_r$, the number of training observations outside $\mathcal J'$ is $m-k+2$, so
	\[
	\sum_{j\notin\mathcal J'\cup\{i\}}c^{(k)}_{i;\mathcal J'\cup\{j\}}
	=
	\frac{1}{m^{k-1}}\sum_r a_{ir}\prod_{j'\in\mathcal J'}I_{j'r}\sum_{j\notin\mathcal J'\cup\{i\}}I_{jr}
	=
	\frac{m-k+2}{m}\,c^{(k-1)}_{i;\mathcal J'},
	\]
	with a nonzero multiplier for $k\le m+1$. For a set $\mathcal J$ of $k-1$ observations, a split with $\mathcal J\subseteq T_r$ has $\sum_{i\notin\mathcal J}a_{ir}=\sum_{i\in V_r}a_{ir}=s_r$, since $a_{ir}=0$ unless $i\in V_r$ and $V_r$ is disjoint from $\mathcal J$, so
	\[
	\sum_{i\notin\mathcal J}c^{(k)}_{i;\mathcal J}
	=
	\frac{1}{m^{k-1}}\sum_r s_r\prod_{j\in\mathcal J}I_{jr}
	=
	d^{(k-1)}_{\mathcal J}.
	\]
	If the weights of order $k$ do not vary with $I$, neither do the two left-hand sides, so the weights of order $k-1$ do not vary with $I$. Induction gives balance at every lower order. The restriction to $k\le m+1$ is needed. For $k>m+1$ every weight of order $k$ is zero, and the first multiplier is zero at $k=m+2$.
	
	\noindent\emph{Step 2: Variance of a balanced design and the lower bound.}\quad
	By Section~\ref{subsec:split_noise}, the variance of any fixed-size ordinary design is the sum over orders of the sampling variance and the split noise, the sampling variance is $\mathcal V/n$ at order one and $\Pi_n^{(l)}$ at order $l\ge2$ for every such design at the same $n$ and $m$, and the split noise at every order is nonnegative. Hence every such design has variance
	\[
	\Var(\widehat R_D)
	=\frac{\mathcal V}{n}+\sum_{l\ge2}\Pi_n^{(l)}+Q_D+\sum_{l\ge2}\mathcal N_D^{(l)}
	\ge\frac{\mathcal V}{n}+\sum_{l=2}^{k}\Pi_n^{(l)}+Q_D+\sum_{l=2}^{k}\mathcal N_D^{(l)},
	\]
	since every \(\Pi_n^{(l)}\) and every \(\mathcal N_D^{(l)}\) is nonnegative. By \eqref{eq:split_noise_orderwise},
	\[
	\mathcal N_D^{(l)}=\sum_{|\mathcal S|=l}\Big\{\mathbb E_I\big(\|h_{\mathcal S,I}\|_2^2\big)-\|\bar h_{\mathcal S}\|_2^2\Big\},
	\]
	and since \(\bar h_{\mathcal S}=\mathbb E_I(h_{\mathcal S,I})\) with \(I\) independent of the data,
	\[
	\mathbb E_I\big(\|h_{\mathcal S,I}\|_2^2\big)-\|\bar h_{\mathcal S}\|_2^2=\mathbb E_I\big(\|h_{\mathcal S,I}-\bar h_{\mathcal S}\|_2^2\big).
	\]
	So \(\mathcal N_D^{(l)}=0\) only if \(h_{\mathcal S,I}=\bar h_{\mathcal S}\) for almost every \(I\) and every set \(\mathcal S\) of size \(l\), that is, only if the components of order \(l\) do not vary with \(I\). For a design balanced at order $k$ the split noise is zero at orders $1,\dots,k$ by Step 1 and Section~\ref{subsec:split_noise}, and the sampling variance and split noise at orders above $k$ sum to $\sum_{l>k}\mathbb E_I\{\rho_n^{(l)}(I)\}=\mathbb E_I\{\rho_n^{((k+1)+)}(I)\}$ by Theorem~\ref{thm:variance_identity}. This gives the displayed identity. For any fixed-size design the parts above order \(k\) sum to \(\sum_{l>k}\{\Pi_n^{(l)}+\mathcal N_D^{(l)}\}=\mathbb E_I\{\rho_n^{((k+1)+)}(I)\}\), which is \(o(n^{-k})\) under A2 at order \(k\). Then
	\[
	\Var(\widehat R_D)=\frac{\mathcal V}{n}+\sum_{l=2}^{k}\Pi_n^{(l)}+Q_D+\sum_{l=2}^{k}\mathcal N_D^{(l)}+o(n^{-k}),
	\]
	so the design attains the lower bound up to \(o(n^{-k})\) if and only if \(Q_D+\sum_{l=2}^{k}\mathcal N_D^{(l)}=o(n^{-k})\), and a design balanced at order \(k\) has \(Q_D+\sum_{l=2}^{k}\mathcal N_D^{(l)}=0\).
	
	\noindent\emph{Step 3: The within-fold AUC.}\quad
	Condition on $Y$ and take a class-stratified fixed-size design as in Section~\ref{appendix:proof_split_lotv_auc}, with the weights $e^{(k)}$, $c^{(k)}$, $d^{(k)}$ of Section~\ref{appendix:proof_auc_identity}. For $2\le k\le m+1$, sets $\mathcal J''$, $\mathcal J'$, $\mathcal J$ of $k-3$, $k-2$, and $k-1$ observations, and $i$, $j$ outside them, summing the indicators of the training set over a split as in Step 1 gives
	\[
	\sum_{l\notin\mathcal J''\cup\{i,j\}}e^{(k)}_{ij;\mathcal J''\cup\{l\}}=\frac{m-k+3}{m}\,e^{(k-1)}_{ij;\mathcal J''},
	\qquad
	\sum_{j\notin\mathcal J'\cup\{i\}}c^{(k)}_{i;\mathcal J'\cup\{j\}}=\frac{m-k+2}{m}\,c^{(k-1)}_{i;\mathcal J'},
	\]
	the first for $k\ge3$, and summing over the validated observations of a split gives
	\[
	\sum_{i\notin\mathcal J}c^{(k)}_{i;\mathcal J}=2\,d^{(k-1)}_{\mathcal J},
	\]
	since \(\sum_{i\in V_r}|V_r^{1-Y_i}|=2|V_r^1||V_r^0|\). As in Step 1, weights of order $k$ that do not vary with $I$ given $Y$ make those of every lower order constant too. By Steps 4 and 5 of Section~\ref{appendix:proof_split_lotv_auc}, given $Y$ the variance is the sampling variance plus a nonnegative split noise at every order, the sampling variance depends on the design only through its training-composition rule, and the split noise of an order is zero when its weights do not vary. Step 2 then holds given $Y$, and averaging over $Y$ gives Theorem~\ref{thm:balanced_optimal} for the within-fold AUC, with the class-stratified designs of a common training-composition rule in place of the fixed-size designs of a common training size.
	
	\noindent\emph{Step 4: Repeated $K$-fold designs and the split noise at order two.}\quad
	For $K$-fold with $R$ partitions into equal folds, pooled, $a_{ir}=1/(Rn)$ for $i\in V_r$, $s_r=1/(RK)$, and $m=n-n/K$. Every observation is validated once and trains in $K-1$ folds per partition, so every $K$-fold design is balanced at order one. At $k=2$, with $\lambda_{ij}$ the number of partitions in which $i$ and $j$ share a fold, $i$ is validated while $j$ trains in $R-\lambda_{ij}$ partitions, and the number of folds containing neither is $(R-\lambda_{ij})(K-2)+\lambda_{ij}(K-1)$, so
	\[
	c_{ij}=c_{ji}=\frac{R-\lambda_{ij}}{Rnm},
	\qquad
	d_{ij}=\frac{R(K-2)+\lambda_{ij}}{RKm^2}.
	\]
	Each partition gives every observation $n/K-1$ fold-mates, so $\sum_{j\ne i}\lambda_{ij}=R(n/K-1)$ for every $i$, and the average of $\lambda_{ij}$ over ordered pairs is $\bar\lambda=R(n-K)/\{K(n-1)\}$. With the observations in random order, condition (iii) holds, $\mathbb E_I(c_{ij})=\bar c:=(R-\bar\lambda)/(Rnm)$, and $\mathbb E_I(d_{ij})=\bar d:=\{R(K-2)+\bar\lambda\}/(RKm^2)$. Since these hold for every pair, each second-order excess is an expected sum of products of deviations,
	\[
	\Theta_D-\Theta^\circ=\Theta_D'-\Theta^\circ=\mathbb E_I\Big\{\sum_{i\ne j}(c_{ij}-\bar c)^2\Big\},
	\quad
	\Psi_D-\Psi^\circ=\mathbb E_I\Big\{\sum_{i<j}(d_{ij}-\bar d)^2\Big\},
	\]
	\[
	\Psi_D'-\Psi'^{\circ}=\mathbb E_I\Big\{\sum_{i\ne j}(c_{ij}-\bar c)(d_{ij}-\bar d)\Big\},
	\]
	where the first equality uses $c_{ij}=c_{ji}$. Reordering the observations permutes the multiset of co-occurrence counts, so the sums do not vary with $I$ and the expectations can be dropped. With $c_{ij}-\bar c=-(\lambda_{ij}-\bar\lambda)/(Rnm)$ and $d_{ij}-\bar d=(\lambda_{ij}-\bar\lambda)/(RKm^2)$, writing $\Delta:=\sum_{i\ne j}(\lambda_{ij}-\bar\lambda)^2$,
	\[
	\mathcal N_D^{(2)}
	=
	\frac{\Delta}{R^2}\Big\{\frac{\tau_n^2+\kappa_n}{n^2m^2}+\frac{\sigma_{\chi,n}^2}{2K^2m^4}-\frac{2\xi_n}{nKm^3}\Big\}
	=
	\frac{K^2\,G_n(K)}{(K-1)\,n^4R^2}\,\Delta,
	\]
	the second form by $m=n(K-1)/K$ and the definition of $G$, with $G_n$ the function $G$ of Section~\ref{subsec:loo} at the pair constants of training size $m$, which is \eqref{eq:split_noise_counts}. It is zero when $\Delta=0$, and only then when $G_n(K)>0$. $\Delta=0$ holds if and only if every pair shares a fold in the same number of partitions, which is the constancy of $c_{ij}$ and $d_{ij}$ across pairs and realizations. For independently drawn partitions, $\lambda_{ij}$ is binomial with $R$ trials and success probability $(n/K-1)/(n-1)$, so
	\[
	\mathbb E(\Delta)=n(n-1)\Var(\lambda_{ij})=\frac{nmR(n-K)}{K(n-1)}.
	\]
	The second-order split noise of independently repeated \(K\)-fold is therefore
	\[
	\mathcal N_D^{(2)}=\frac{G_n(K)\,(n-K)}{Rn^2(n-1)},
	\]
	and its excess role moments satisfy \(n^2(\Theta_D-\Theta^\circ)=n^2(\Theta_D'-\Theta^\circ)\to1/\{R(K-1)\}\), \(n^2(\Psi_D-\Psi^\circ)\to1/\{2R(K-1)^3\}\), and \(n^2(\Psi_D'-\Psi'^{\circ})\to-1/\{R(K-1)^2\}\), with \(n^2\Theta^\circ\to1\), \(n^2\Psi^\circ\to\frac12\), and \(n^2\Psi'^{\circ}\to1\).
	
	\noindent\emph{Step 5: The number of fits exact pair balance needs.}\quad
	Take $R$ partitions of $n$ observations into $K$ folds of size $g=n/K$, and let $\mathbf A$ be the co-occurrence matrix, with $\mathbf A_{ii}=R$ and $\mathbf A_{ij}=\lambda_{ij}$. With $\mathbf P_\rho$ the orthogonal projection onto the span of the fold indicators of partition $\rho$,
	\[
	\mathbf A=g\sum_{\rho=1}^R\mathbf P_\rho,
	\qquad
	\mathbf P_\rho=\frac{\mathbf 1\mathbf 1^\top}{n}+\mathbf P^\circ_\rho,
	\]
	with $\mathbf P^\circ_\rho$ a projection of rank $K-1$ orthogonal to $\mathbf 1$. On the orthogonal complement of $\mathbf 1$, $\mathbf A$ is $g\sum_\rho\mathbf P^\circ_\rho$, of rank at most $R(K-1)$. If every $\lambda_{ij}$ equals $\bar\lambda$, then on that complement
	\[
	\mathbf A=(R-\bar\lambda)\,\mathbf I,
	\qquad
	R-\bar\lambda=\frac{Rg(K-1)}{n-1}>0,
	\]
	which has rank $n-1$. Hence exact pair balance needs $R(K-1)\ge n-1$, that is, at least $K(n-1)/(K-1)$ fits.
	
	\noindent\emph{Step 6: Balance at order three.}\quad
	For $K$-fold with folds of size $g=n/K$, each partition puts a given set of three observations in one of three patterns: all three in one fold, one of them apart from the other two, which share a fold, or, when $K\ge3$, all three in different folds. Over the $\binom n3$ triples, one partition has $K\binom g3$ of the first, $K\binom g2(n-g)$ of the second, and $\binom K3g^3$ of the third. For a triple \(\{i,j,l\}\) let \(B_i\) and \(C\) count the partitions with \(i\) apart and with all three apart. Up to factors common to all triples, the order-three weight with \(i\) validated and \(j,l\) training is \(B_i+C\), and the weight with all three training is \((K-1)R-(B_i+B_j+B_l)-2C\). Both are the same for every triple only if every \(B_i\) and \(C\) are, and then so is the number \(R-B_i-B_j-B_l-C\) of partitions with the triple together. Balance at order three therefore requires every triple to be in each pattern, and every member to be the one apart, in the same number of partitions, so
	\[
	\frac{RK\binom g3}{\binom n3},
	\qquad
	\frac{RK\binom g2(n-g)}{3\binom n3},
	\qquad
	\frac{R\binom K3g^3}{\binom n3}
	\]
	must be integers, together with $\bar\lambda=R(g-1)/(n-1)$. At $n=27$ and $K=3$ this requires $R$ to be a multiple of $325$, against $13$ partitions for pair balance.
\end{proof}

\subsection{Derivation of \eqref{eq:K_second_corrected} and \eqref{eq:loo_endpoint_variance_gap}}
\label{app:endpoint_variance_optimality}

\begin{proof}
	\noindent\emph{Step 1: The first-order part.}\quad
	By \eqref{eq:floor_split}, the variance of $K$-fold is its sampling variance plus its split noise. By Step~2 of Section~\ref{appendix:proof_split_lotv}, the single-observation part of the sampling variance of a fixed-size design at training size \(m\) is
	\[
	\frac{\Var\{H_m(Z)\}}{n}=\frac{\mathcal V_m}{n},
	\]
	and \(K\)-fold has no first-order split noise, by \eqref{eq:QD_designs}.
	
	\noindent\emph{Step 2: The pair part.}\quad
	By Step~3 of Section~\ref{appendix:proof_split_lotv}, the pair part of the sampling variance at training size \(m\) is
	\[
	\Pi_n(m)=\frac{1}{n(n-1)}\Big\{\tau_n^2+\kappa_n+\tfrac12\sigma_{\chi,n}^2\big(1-\tfrac1m\big)^2+2\xi_n\big(1-\tfrac1m\big)\Big\},
	\]
	with the pair constants at training size \(m\), so \(n^2\Pi_n(m)\to\tau^2+\kappa+\tfrac12\sigma_\chi^2+2\xi\) at both \(m=pn\) and \(m=n-1\) when they converge to common limits. By Step~4 of Section~\ref{appendix:proof_order_balance}, the pair part of the split noise of repeated \(K\)-fold is \(G_n(K)(n-K)/\{Rn^2(n-1)\}=G(K)/(Rn^2)+o(n^{-2})\).
	
	\noindent\emph{Step 3: Higher orders.}\quad
	The parts of the sampling variance and of the split noise of order three and higher are bounded by \(\mathbb E_I\{\rho_n^{(3+)}(I)\}\) (Step~4 of Section~\ref{appendix:proof_split_lotv}), which is \(o(n^{-2})\) under A2(ii).
	
	\noindent\emph{Step 4: The two variances.}\quad
	Adding Steps 1--3, with \(p=(K-1)/K\),
	\[
	\begin{split}
		\Var(\widehat R_{K,R})
		&=\frac{\mathcal V_{pn}}{n}+\frac{\tau^2+\kappa+\tfrac12\sigma_\chi^2+2\xi}{n^2}+\frac{G(K)}{Rn^2}+o(n^{-2}),\\
		\Var(\widehat R_{\mathrm{LOO}})
		&=\frac{\mathcal V_{n-1}}{n}+\frac{\tau^2+\kappa+\tfrac12\sigma_\chi^2+2\xi}{n^2}+o(n^{-2}).
	\end{split}
	\]
	The second line holds because leave-one-out has no split noise at any order (Step~5 of Section~\ref{appendix:proof_split_lotv}). Subtraction proves \eqref{eq:loo_endpoint_variance_gap}.
\end{proof}

\subsection{The sign of the finite-training-size correction}
\label{app:sign_training_correction}

\paragraph{A general covariance interpretation.}
This derives \eqref{eq:rate_gap} and writes \(b\) as a covariance.
\begin{proof}
	\noindent\emph{Step 1: Expand the variance.}\quad
	Let \(H_m(Z)=\ell_m^\star(Z)-\mathbb E(\ell_m^\star)+\bar\phi_m(Z)\) be the combined influence at training size \(m\). Suppose that
	\[
	H_m=H_\infty+\Delta_H m^{-\gamma}+r_m,
	\qquad
	\|r_m\|_{L_2(P)}=o(m^{-\gamma}),
	\qquad
	H_\infty,\Delta_H\in L_2(P),
	\]
	for some \(0<\gamma\le1\). Then \(\mathbb E(H_\infty)=\mathbb E(\Delta_H)=0\), since \(\mathbb E(H_m)=0\) for every \(m\). Expanding,
	\[
	\Var(H_m)=\Var(H_\infty)+2m^{-\gamma}\Cov(H_\infty,\Delta_H)+m^{-2\gamma}\Var(\Delta_H)+2\Cov(H_\infty+\Delta_H m^{-\gamma},r_m)+\Var(r_m).
	\]
	
	\noindent\emph{Step 2: The coefficient.}\quad
	The third term is \(O(m^{-2\gamma})\), which is \(o(m^{-\gamma})\) because \(\gamma>0\). By Cauchy--Schwarz the fourth is \(o(m^{-\gamma})\), and the fifth is \(o(m^{-2\gamma})\). Hence \(\mathcal V_m=\mathcal V_\infty+bm^{-\gamma}+o(m^{-\gamma})\) with
	\[
	b=2\Cov(H_\infty,\Delta_H).
	\]
	
	\noindent\emph{Step 3: The gap between two training sizes.}\quad
	For a fixed \(p\in(0,1)\),
	\[
	(pn)^{-\gamma}-(n-1)^{-\gamma}=n^{-\gamma}(p^{-\gamma}-1)+O(n^{-1-\gamma}),
	\]
	and under the scalar expansion of Step~2 the remainders at the two sizes are \(o(n^{-\gamma})\). Dividing by \(n\) gives \eqref{eq:rate_gap}, and at \(\gamma=1\), \(p^{-1}-1=1/(K-1)\).
\end{proof}

\paragraph{An exact fixed-shrinkage family.}
This gives the shrinkage example of Section~\ref{subsec:loo}, with the sample mean at \(\alpha=1\). Let $Y$ have mean $\mu$, variance $\sigma^2$, and finite fourth central moment. Write $\epsilon=Y-\mu$, $\mu_3=\mathbb E(\epsilon^3)$, and $\mu_4=\mathbb E(\epsilon^4)$. For a fixed $0<\alpha\le1$ and fixed target $t$, fit the constant prediction
\[
\widehat f_T=\alpha\bar Y_T+(1-\alpha)t,
\]
and evaluate by squared error.

\begin{proof}
	\noindent\emph{Step 1: The validation influence.}\quad
	Put $d=(1-\alpha)(\mu-t)$. An independent validation error is $\epsilon+d-\alpha\bar\epsilon_T$. Taking the training expectation gives
	\[
	\ell_m^\star(Y)=(\epsilon+d)^2+\frac{\alpha^2\sigma^2}{m},
	\qquad
	\mathbb E(\ell_m^\star)=\sigma^2+d^2+\frac{\alpha^2\sigma^2}{m}.
	\]
	
	\noindent\emph{Step 2: The training influence.}\quad
	The population risk of a fitted constant is $\sigma^2+(d-\alpha\bar\epsilon_T)^2$. Conditioning on one training observation and multiplying its centered effect by $m$ gives exactly
	\[
	\bar\phi_m(Y)=-2\alpha d\epsilon
	+\frac{\alpha^2}{m}(\epsilon^2-\sigma^2).
	\]
	
	\noindent\emph{Step 3: The combined influence.}\quad
	Adding Steps 1 and 2,
	\[
	H_m(Y)=\left(1+\frac{\alpha^2}{m}\right)(\epsilon^2-\sigma^2)
	+2(1-\alpha)^2(\mu-t)\epsilon.
	\]
	
	\noindent\emph{Step 4: Its variance and \(b\).}\quad
	With $\Var(\epsilon^2)=\mu_4-\sigma^4$,
	\[
	\begin{split}
		\Var(H_m)
		&=\left(1+\frac{\alpha^2}{m}\right)^2\Var(\epsilon^2)
		+4\left(1+\frac{\alpha^2}{m}\right)(1-\alpha)^2(\mu-t)\mu_3+4(1-\alpha)^4(\mu-t)^2\sigma^2\\
		&=\mathcal V_\infty+\frac{b}{m}+\frac{\alpha^4\,\Var(\epsilon^2)}{m^2},
	\end{split}
	\]
	where
	\begin{equation}
		\label{eq:shrinkage_training_correction}
		b=2\alpha^2\left\{\mu_4-\sigma^4
		+2(1-\alpha)^2(\mu-t)\mu_3\right\}.
	\end{equation}
	
	\noindent\emph{Step 5: The sign of \(b\).}\quad
	At $\alpha=1$ the rule is the sample mean and $b=2(\mu_4-\sigma^4)\ge0$. At $\mu_3=0$, or at $t=\mu$, $b=2\alpha^2(\mu_4-\sigma^4)\ge0$ for every $\alpha$. For $0<\alpha<1$, $b<0$ exactly when
	\[
	2(1-\alpha)^2(\mu-t)\mu_3<-(\mu_4-\sigma^4).
	\]
	This happens for a Bernoulli outcome. For \(Y\sim\mathrm{Bernoulli}(\theta)\), where squared error is the Brier score, \(\mu_3=\theta(1-\theta)(1-2\theta)\) and \(\mu_4-\sigma^4=\theta(1-\theta)(1-2\theta)^2\). At \(\alpha=1/2\) and \(t=3/4\), so that \(\widehat f_T=\bar Y_T/2+3/8\) lies in \([3/8,7/8]\),
	\[
	b=\frac12\theta(1-\theta)(1-2\theta)\left(\frac58-\frac32\theta\right)<0
	\qquad\text{for }\frac5{12}<\theta<\frac12.
	\]
\end{proof}

\subsection{The sign of \(\xi\)}
\label{app:sign_xi}

\paragraph{A distorted risk.}
This gives the example of Section~\ref{subsec:loo}. Let \(Y\sim\mathrm{Bernoulli}(\theta)\) with \(0<\theta<1\), and write \(\epsilon=Y-\theta\) and \(\sigma^2=\theta(1-\theta)\). For a fixed function \(g\) that is bounded on \([0,1]\) and three times continuously differentiable near \(\theta\), fit the risk
\[
\widehat f_T=g(\bar Y_T),
\]
and evaluate by the Brier score. At \(g(y)=y\) the rule is the sample mean. Write \(g\), \(g'\), \(g''\) for the values at \(\theta\).

\begin{proof}
	\noindent\emph{Step 1: The fit.}\quad
	With \(\bar\epsilon_T=\bar Y_T-\theta\),
	\[
	g(\bar Y_T)=g+g'\bar\epsilon_T+\tfrac12g''\bar\epsilon_T^2+r_T .
	\]
	On \(|\bar\epsilon_T|\le\delta\) the remainder is at most a constant times \(|\bar\epsilon_T|^3\), and \(\mathbb P(|\bar\epsilon_T|>\delta)\) decreases exponentially in \(m\) while \(g\) is bounded, so \(\|r_T\|_{L_4}=O(m^{-3/2})\).
	
	\noindent\emph{Step 2: The validation--training kernel.}\quad
	The contribution of a validated observation \(i\) is \(Y_i-2Y_ig(\bar Y_T)+g(\bar Y_T)^2\). Only the middle term depends jointly on \(Y_i\) and a training observation \(j\), through the part of \(g(\bar Y_T)\) that depends on \(Y_j\) alone, which is \(g'\epsilon_j/m+O(m^{-2})\) by Step~1. Hence
	\[
	\phi^{(2)}(Z_i;Z_j)=-2g'\epsilon_i\epsilon_j+O(m^{-1}).
	\]
	
	\noindent\emph{Step 3: The training--training kernel.}\quad
	The expected Brier score of a fitted risk \(f\) is \(\sigma^2+(\theta-f)^2\), and
	\[
	\{\theta-g(\bar Y_T)\}^2=(\theta-g)^2-2(\theta-g)\{g(\bar Y_T)-g\}+\{g(\bar Y_T)-g\}^2 .
	\]
	By Step~1, the part of \(g(\bar Y_T)-g\) that depends jointly on two training observations \(j\) and \(k\) is \(g''\epsilon_j\epsilon_k/m^2\), and that of \(\{g(\bar Y_T)-g\}^2\) is \(2g'^2\epsilon_j\epsilon_k/m^2\), each up to \(O(m^{-5/2})\). Hence
	\[
	\chi(Z_j,Z_k)=2\big\{g'^2-(\theta-g)g''\big\}\epsilon_j\epsilon_k+O(m^{-1/2}).
	\]
	
	\noindent\emph{Step 4: \(\xi\).}\quad
	Since \(\mathbb E(\epsilon_1^2\epsilon_2^2)=\sigma^4\), Steps 2 and 3 give
	\[
	\xi=-4\sigma^4g'\big\{g'^2-(\theta-g)g''\big\}.
	\]
	At \(g(y)=y\), \(\xi=-4\sigma^4\), the value for the sample mean.
	
	\noindent\emph{Step 5: The sign.}\quad
	For increasing \(g\), \(\xi>0\) exactly when
	\[
	(\theta-g)g''>g'^2 .
	\]
	The left side is zero when the fitted risk is calibrated, \(g=\theta\), and when \(g\) is linear, and then \(\xi<0\).
	
	\noindent\emph{Step 6: Risks pushed toward 0 and 1.}\quad
	Take
	\[
	g(y)=\frac{y^\alpha}{y^\alpha+(1-y)^\alpha},
	\qquad \alpha>1,
	\]
	which multiplies the log-odds by \(\alpha\). Then \(g(1/2)=1/2\), and for \(\theta\ne1/2\) the fitted risk lies between \(\theta\) and the nearer of 0 and 1. With
	\[
	g'=\frac{\alpha g(1-g)}{\theta(1-\theta)},
	\qquad
	g''=\frac{\alpha g(1-g)\{\alpha(1-2g)+2\theta-1\}}{\theta^2(1-\theta)^2},
	\]
	the condition of Step~5 becomes
	\[
	(\theta-g)\{\alpha(1-2g)+2\theta-1\}>\alpha g(1-g).
	\]
	At \(\theta=1/2\) the left side is zero, so \(\xi<0\). At \(\theta=0.1\) the condition holds exactly when \(\alpha>1.571\), where the fitted risk is \(0.0307\).
\end{proof}

\subsection{Proof of Corollary~\ref{cor:loo_limit} and exhaustive designs}
\label{app:loo_corollaries}

\paragraph{Corollary~\ref{cor:loo_limit}.}
\begin{proof}
	\noindent\emph{Step 1: The two terms.}\quad
	By \eqref{eq:loo_endpoint_variance_gap}, with \(p=(K-1)/K\), the difference of the two variances is the sampling variance term \((\mathcal V_{pn}-\mathcal V_{n-1})/n\) plus the split-noise term \(G(K)/(Rn^2)\), up to \(o(n^{-2})\).
	
	\noindent\emph{Step 2: The sampling variance term.}\quad
	The training size \(pn=n-n/K\) increases in \(K\) and is at most \(n-1\). When \(\mathcal V_m\) does not increase in \(m\), \(\mathcal V_{pn}\ge\mathcal V_{n-1}\), and \(\mathcal V_{pn}\) does not increase in \(K\). So the first term is nonnegative and does not increase in \(K\).
	
	\noindent\emph{Step 3: The split-noise term is nonnegative.}\quad
	At training size \(m\), write \(\phi^{(2)}_1\) and \(\phi^{(2)}_2\) for the validation--training kernel of a pair in its two orientations, as in Section~\ref{subsec:variance_decomposition}, and \(\chi\) for its training--training kernel. The two orientations have the same law and \(\chi\) is symmetric, so
	\[
	\mathbb E\{(\phi^{(2)}_1+\phi^{(2)}_2)^2\}=2(\tau_n^2+\kappa_n),
	\qquad
	\mathbb E\{(\phi^{(2)}_1+\phi^{(2)}_2)\chi\}=2\xi_n,
	\qquad
	\mathbb E(\chi^2)=\sigma_{\chi,n}^2.
	\]
	Hence
	\[
	\frac{\tau_n^2+\kappa_n}{K-1}-\frac{2\xi_n}{(K-1)^2}+\frac{\sigma_{\chi,n}^2}{2(K-1)^3}
	=\frac{1}{2(K-1)}\,\mathbb E\Big\{\Big(\phi^{(2)}_1+\phi^{(2)}_2-\frac{\chi}{K-1}\Big)^2\Big\}\ge0,
	\]
	and in the limit
	\[
	G(K)=\frac{\tau^2+\kappa}{K-1}-\frac{2\xi}{(K-1)^2}+\frac{\sigma_\chi^2}{2(K-1)^3}\ge0
	\]
	for every \(K\ge2\), whatever the sign of \(\xi\). With Step~2, up to \(o(n^{-2})\), the variance of repeated \(K\)-fold is at least that of leave-one-out. At \(K=n\), \(K\)-fold cross-validation is leave-one-out. This proves (a).
	
	\noindent\emph{Step 4: The split-noise term does not increase in \(K\) when \(\xi\le0\).}\quad
	By Cauchy--Schwarz, \(|\kappa|\le\tau^2\), so \(\tau^2+\kappa\ge0\). Also \(\sigma_\chi^2\ge0\), and \(-2\xi\ge0\) when \(\xi\le0\). Each of the three terms of \(G(K)\) is then nonnegative and does not increase in \(K\). With Step~2, up to \(o(n^{-2})\), the variance of repeated \(K\)-fold does not increase in \(K\).
	
	\noindent\emph{Step 5: The variance at a fixed number of fits.}\quad
	With \(R=S/K\), \eqref{eq:K_second_corrected} gives
	\[
	\Var(\widehat R_{K,S/K})=\frac{\mathcal V_{pn}}{n}+\frac{\tau^2+\kappa+\tfrac12\sigma_\chi^2+2\xi}{n^2}+\frac{KG(K)}{Sn^2}+o(n^{-2}),
	\]
	with
	\[
	KG(K)=\frac{K}{K-1}\Big\{\tau^2+\kappa-\frac{2\xi}{K-1}+\frac{\sigma_\chi^2}{2(K-1)^2}\Big\}.
	\]
	
	\noindent\emph{Step 6: The split noise per fit does not increase in \(K\).}\quad
	By Step~4, both factors of \(KG(K)\) in Step~5 are nonnegative and do not increase in \(K\), so neither does \(KG(K)\).
	
	\noindent\emph{Step 7: The sampling variance.}\quad
	For \(K_1>K_2\) dividing \(S\), the training fractions satisfy \(p_1>p_2\), so \(\mathcal V_{p_1n}\le\mathcal V_{p_2n}\) by Step~2. With Step~6,
	\[
	\Var(\widehat R_{K_1,S/K_1})\le\Var(\widehat R_{K_2,S/K_2})+o(n^{-2}),
	\]
	and the largest \(K\), one run of \(S\)-fold, has the smallest variance up to \(o(n^{-2})\). This proves (b).
\end{proof}

\paragraph{Exhaustive designs.}
Leave-one-out is the case \(m=n-1\) of exhaustive leave-\((n-m)\)-out, which uses every training set of size \(m\) once, so it has no split noise and attains the minimum, \(\Var_{\mathrm{samp}}(m)\). Among fixed-size designs whose training sizes lie in a given set containing \(n-1\), exhaustive designs included, leave-one-out is therefore smallest exactly when \(\Var_{\mathrm{samp}}(n-1)\le\Var_{\mathrm{samp}}(m)\) at every \(m\) in the set (below). Under A2(ii) and convergence of the pair constants this turns only on \(\mathcal V_m\), since \(\Var_{\mathrm{samp}}(pn)-\Var_{\mathrm{samp}}(n-1)=(\mathcal V_{pn}-\mathcal V_{n-1})/n+o(n^{-2})\). When \(\Var_{\mathrm{samp}}(pn)\) is lower, as when \(b<0\), more than \(\mathcal N_{K,1}/\{\Var_{\mathrm{samp}}(n-1)-\Var_{\mathrm{samp}}(pn)\}\) repetitions give \(K\)-fold the smaller variance, by \eqref{eq:repeated_floor}.
\begin{proof}
	\noindent\emph{Step 1: Every design is its sampling variance plus its split noise.}\quad
	Fix \(n\). Section~\ref{appendix:proof_split_lotv} uses of the design only that it is fixed-size, that each validation set is the complement of its training set, condition (iii), pooled aggregation, and independence of \(I\) from the data, so its conclusions hold for every design compared. Each such design has variance equal to its sampling variance plus its split noise, the split noise is nonnegative, and the sampling variance depends on the design only through \(n\) and its training size \(m\).
	
	\noindent\emph{Step 2: Exhaustive designs have no split noise.}\quad
	Exhaustive leave-\((n-m)\)-out uses every training set of size \(m\) once, in random order, so it is a fixed-size design that satisfies condition (iii), and its weights of every order do not vary with \(I\). By the argument of Step~5 of Section~\ref{appendix:proof_split_lotv}, its split noise is zero at every order, so its variance is \(\Var_{\mathrm{samp}}(m)\). In particular \(\Var(\widehat R_{\mathrm{LOO}})=\Var_{\mathrm{samp}}(n-1)\).
	
	\noindent\emph{Step 3: The two directions.}\quad
	If \(\Var_{\mathrm{samp}}(n-1)\le\Var_{\mathrm{samp}}(m)\) for every training size \(m\) in the set, every design \(D\) compared, at training size \(m_D\), has
	\[
	\Var(\widehat R_D)\ge\Var_{\mathrm{samp}}(m_D)\ge\Var_{\mathrm{samp}}(n-1)=\Var(\widehat R_{\mathrm{LOO}}).
	\]
	If \(\Var_{\mathrm{samp}}(m)<\Var_{\mathrm{samp}}(n-1)\) for some training size \(m\) in the set, exhaustive leave-\((n-m)\)-out is among the designs compared and has variance \(\Var_{\mathrm{samp}}(m)<\Var(\widehat R_{\mathrm{LOO}})\).
	
	\noindent\emph{Step 4: The second-order form.}\quad
	By Steps 1--4 of Section~\ref{appendix:proof_split_lotv},
	\[
	\Var_{\mathrm{samp}}(m)=\frac{\mathcal V_m}{n}+\Pi_n(m)+\sum_{k\ge3}\Pi_n^{(k)}(m),
	\]
	where \(0\le\sum_{k\ge3}\Pi_n^{(k)}(m)\le\mathbb E_I\{\rho_n^{(3+)}(I)\}\) for every design at training size \(m\). Under A2(ii) and convergence of the pair constants, this sum is \(o(n^{-2})\) at \(m=pn\) and at \(m=n-1\), and \(n^2\{\Pi_n(pn)-\Pi_n(n-1)\}\to0\). Hence
	\[
	\Var_{\mathrm{samp}}(pn)-\Var_{\mathrm{samp}}(n-1)=\frac{\mathcal V_{pn}-\mathcal V_{n-1}}{n}+o(n^{-2}).
	\]
	
	\noindent\emph{Step 5: The sign when \(b<0\).}\quad
	If \(\mathcal V_m=\mathcal V_\infty+bm^{-\gamma}+o(m^{-\gamma})\) with \(b<0\) and \(0<\gamma\le1\), the difference in Step~4 is negative and of exact order \(n^{-1-\gamma}\) by \eqref{eq:rate_gap}, so the sampling variance at \(pn\) is the lower one for all large \(n\).
\end{proof}

\subsection{Derivation of \eqref{eq:lpo_endpoint_variance_gap} and \eqref{eq:paired_gap}}
\label{appendix:proof_pairwise_second}

The conditions of \eqref{eq:lpo_endpoint_variance_gap} and \eqref{eq:paired_gap}, for each design compared and for leave-pair-out, are A4, A2(ii) under the law given the labels, which is \eqref{eq:auc_third_remainder} below, and convergence in \(L_2\) of the pair kernels as the training size grows, uniformly for training class fractions near \(\pi_1\), to limits continuous at \(\pi_1\).

Condition on the label vector $Y$ and on a class-stratified inclusion structure $I$. Conditional on $Y$, the covariates are independent under the product law $\bigotimes_i P_{Y_i}$. Write $n_1=\sum_iY_i$, $n_0=n-n_1$, $\hat\pi=n_1/n$, and $p=(K-1)/K$, $m=pn$. For fixed $K$ and population class proportions bounded away from zero, the event $\mathcal E_{n,K}:=\{n_1\ge K,n_0\ge K\}$ has probability $1-O(e^{-cn})$ for some $c>0$. All AUC statements are understood on this event, whose complement is negligible at every polynomial order considered below. Throughout, \(K\) divides both class counts, so that every training set has $m_y=pn_y$ observations of class $y$.

\begin{proof}
	\noindent\emph{Step 1: The decomposition.}\quad
	Write $\mu_{I,Y}:=\mathbb E(\widehat R_D\mid I,Y)$ and $h_{\mathcal S,I,Y}$ for the components \eqref{eq:auc_components} under the conditional product law. By Theorem~\ref{thm:pairwise_identity}, with the norms conditional on $(I,Y)$,
	\[
	\begin{aligned}
		\Var(\widehat R_D)={}&\Var_{I,Y}(\mu_{I,Y})+\mathbb E_{I,Y}\Big\{\sum_i\|h_{\{i\},I,Y}\|_2^2\Big\}\\
		&+\mathbb E_{I,Y}\Big\{\sum_{a<b}\|h_{\{a,b\},I,Y}\|_2^2\Big\}+\mathbb E_{I,Y}\{\rho_n^{(3+)}(I,Y)\},
	\end{aligned}
	\]
	the first-order part, the pair part, and the rest. The remainder condition is A2(ii) under the conditional law,
	\begin{equation}
		\label{eq:auc_third_remainder}
		\mathbb E_{I,Y}\{\rho_n^{(3+)}(I,Y)\}=o(n^{-2}),
	\end{equation}
	uniformly over the designs compared, leave-pair-out included.
	
	\noindent\emph{Step 2: The exact kernels at a class fraction.}\quad
	The expansion over the random class count below differentiates in the training class fraction, so the first-order influences and the kernels at $k=2$ of Section~\ref{appendix:proof_auc_identity} are indexed here by the fixed-count training law $\mathbb P_{m,t}$, with $mt$ class-1 observations from $P_1$ and $m(1-t)$ class-0 observations from $P_0$, $t\in\mathcal R_m:=\{j/m:j=1,\ldots,m-1\}$. The model is fit symmetrically, so their ordering is immaterial. With the subscript $m,t$ marking the law, write
	\[
	\begin{aligned}
		\mu_m(t)&:=\mathbb E_{m,t}\{\mathrm{AUC}(\widehat f_T)\},
		&g_{y,m,t}&:=g_{y,m}\ \text{for a validated observation of class }y,\\
		\zeta_{m,t}&:=\zeta^{(2)},
		&\bar\phi_{c,m,t}&:=\bar\phi_{Y,m}\ \text{for a training observation of class }c,\\
		\phi^{\circ}_{y,c;m,t}&:=\phi^{(2)}_y,
		&\chi_{cc';m,t}&:=\chi^{(2)}_A,
	\end{aligned}
	\]
	with the training observation of \(\phi^{\circ}\) of class $c$ and those of \(\chi\) of classes $c$ and $c'$. Each has mean zero in every argument under the product law. The class-normalized form
	\[
	\phi^{u\circ}_{y,c;m,t}(x,z):=\frac{\phi^{\circ}_{y,c;m,t}(x,z)}{\pi_y(t)},
	\qquad \pi_1(t)=t,\quad \pi_0(t)=1-t,
	\]
	carries the AUC weights onto the unit-level pair weights. For $R$ exactly class-balanced partitions, $M_A=Rn_1n_0/K$, and $|V_r^{1-y}|=n_{1-y}/K$, so the weight $c^{(2)}_{a;b}$ of Section~\ref{appendix:proof_auc_identity} equals $c_{ab}/\hat\pi_{Y_a}$, with $c_{ab}$ the unit-level pair weight of the same partitions, and
	\[
	c^{(2)}_{a;b}\,\phi^{\circ}_{Y_a,Y_b;m,\hat\pi}=c_{ab}\,\phi^{u\circ}_{Y_a,Y_b;m,\hat\pi}.
	\]
	The training share $|V_r^1||V_r^0|/M_A=1/(RK)$ equals $|V_r|/M$, so $d^{(2)}_{ab}=d_{ab}$. The pair weight $e^{(2)}_{ab}$ is the validation weight $v^A_{ab}$ of the pair in its case--control orientation, zero for a same-class pair. With $(X_{ab}^{(1)},X_{ab}^{(0)})$ the covariates of a pair in case--control order, \eqref{eq:auc_components} at $k=2$ (Section~\ref{appendix:proof_auc_identity}) reads
	\begin{equation}
		\label{eq:auc_exact_pair_kernel}
		\begin{aligned}
			h_{\{a,b\},I,Y}
			={}&v^A_{ab}\,\zeta_{m,\hat\pi}(X_{ab}^{(1)},X_{ab}^{(0)})
			+c_{ab}\,\phi^{u\circ}_{Y_a,Y_b;m,\hat\pi}(X_a,X_b)\\
			&+c_{ba}\,\phi^{u\circ}_{Y_b,Y_a;m,\hat\pi}(X_b,X_a)
			+d_{ab}\,\chi_{Y_aY_b;m,\hat\pi}(X_a,X_b).
		\end{aligned}
	\end{equation}
	No asymptotic approximation is used.
	
	\noindent\emph{Step 3: Reduction to the limiting kernels.}\quad
	Assume both population class proportions are bounded away from zero and, uniformly for $t$ in a neighbourhood of $\pi_1$ and over the finitely many class combinations,
	\begin{equation}
		\label{eq:auc_kernel_L2}
		\zeta_{m,t}\to\zeta_t,\qquad
		\phi^{u\circ}_{y,c;m,t}\to\phi^{u\circ}_{y,c;t},\qquad
		\chi_{cc';m,t}\to\chi_{cc';t}
		\quad\text{in }L_2,
	\end{equation}
	as $m\to\infty$, and suppose the three limiting kernel maps are $L_2$-continuous at $t=\pi_1$. Write
	\[
	\zeta:=\zeta_{\pi_1},
	\qquad
	\phi^{u\circ}_{y,c}:=\phi^{u\circ}_{y,c;\pi_1},
	\qquad
	\chi_{cc'}:=\chi_{cc';\pi_1}.
	\]
	At fixed $K$ and $R$, $v^A_{ab}$, $c_{ab}$, and $d_{ab}$ are all $O(n^{-2})$, and there are $O(n^2)$ observation pairs. Hence \eqref{eq:auc_kernel_L2}, its continuity condition, and $\hat\pi-\pi_1=O_P(n^{-1/2})$ imply
	\[
	\mathbb E_{I,Y}\Big\{\sum_{a<b}\|h_{\{a,b\},I,Y}-q_{ab,I,Y}\|_2^2\Big\}=o(n^{-2}),
	\]
	where $q_{ab,I,Y}$ is \eqref{eq:auc_exact_pair_kernel} with the limiting kernels and population class proportions. Let
	\[
	W_2:=\sum_{a<b}q_{ab,I,Y},
	\qquad
	\Delta_2:=\sum_{a<b}(h_{\{a,b\},I,Y}-q_{ab,I,Y}).
	\]
	Orthogonality of distinct second-order Hoeffding kernels gives $\mathbb E(W_2^2)=O(n^{-2})$ and $\mathbb E(\Delta_2^2)=o(n^{-2})$. Therefore
	\[
	|\Var(W_2+\Delta_2)-\Var(W_2)|
	\le 2\{\mathbb E(W_2^2)\}^{1/2}\{\mathbb E(\Delta_2^2)\}^{1/2}+\mathbb E(\Delta_2^2)
	=o(n^{-2}),
	\]
	which justifies replacing the exact pair component by the limiting one. Under stratification a pair shares a fold with probability \(1/K+O(n^{-1})\), and since the pair part is \(O(n^{-2})\), this changes it only by \(O(n^{-3})\).
	
	\noindent\emph{Step 4: The constants.}\quad
	For independent observations $Z_a$ and $Z_b$, with $\phi^{u\circ}_{Y_a,Y_b}$ and $\chi_{Y_aY_b}$ the limiting class-normalized validation--training and training--training kernels, the four pair constants are the mixture averages over the class pair
	\[
	\begin{aligned}
		\tau_A^2&:=\mathbb E\!\left[\{\phi^{u\circ}_{Y_a,Y_b}(X_a,X_b)\}^2\right],
		&\kappa_A&:=\mathbb E\!\left[\phi^{u\circ}_{Y_a,Y_b}(X_a,X_b)\phi^{u\circ}_{Y_b,Y_a}(X_b,X_a)\right],\\
		\sigma_{\chi,A}^2&:=\mathbb E\!\left[\{\chi_{Y_aY_b}(X_a,X_b)\}^2\right],
		&\xi_A&:=\mathbb E\!\left[\phi^{u\circ}_{Y_a,Y_b}(X_a,X_b)\chi_{Y_aY_b}(X_a,X_b)\right].
	\end{aligned}
	\]
	For a case--control pair $(a,b)$ the limiting pair-specific kernel and its variance are
	\[
	\zeta_{ab} := \ell^\star_{(a,b)} - \mathrm{AUC}_p - g_1(Z_a) - g_0(Z_b),
	\qquad
	\varsigma^2 = \Var\big(\ell^\star_{(a,b)}\big) - \sigma_{V\mid1}^2 - \sigma_{V\mid0}^2,
	\]
	and its oriented covariances with the training kernels are
	\[
	\omega_A := \mathbb{E}\{\zeta_{ab}\phi^{u\circ}_{1,0}(X_a,X_b)\},
	\qquad
	\omega_A' := \mathbb{E}\{\zeta_{ab}\phi^{u\circ}_{0,1}(X_b,X_a)\},
	\qquad
	\omega_A'' := \mathbb{E}\{\zeta_{ab}\chi_{10}(X_a,X_b)\}.
	\]
	
	\noindent\emph{Step 5: The first-order part.}\quad
	By Step~4 of Section~\ref{appendix:proof_split_lotv_auc}, the first-order part of stratified \(K\)-fold is its first-order sampling variance \(\mathcal W_{1,n}(K)\), with no first-order split noise, since \(\tilde v_i=1/n_{Y_i}\) and \(w_i=1/n\) in every partition. With $t_1=t$, $t_0=1-t$, and $\mathcal Q_{y,m}$ of that step at training class fraction $t$, let
	\[
	\mathcal Q_m(t):=\frac{\mathcal Q_{1,m}(t_1)}{t}+\frac{\mathcal Q_{0,m}(t_0)}{1-t}.
	\]
	At the proportional composition, $R_{p,n}(Y)=\mu_m(\hat\pi)$ and $n_y=n\hat\pi_y$, so \eqref{eq:auc_first_floor} is
	\begin{equation}
		\label{eq:auc_W1_exact}
		\mathcal W_{1,n}(K)
		=\Var\{\mu_m(\hat\pi)\}+\frac1n\,\mathbb E\{\mathcal Q_m(\hat\pi)\},
	\end{equation}
	which does not depend on \(R\).
	
	\noindent\emph{Step 6: The pair part of repeated \(K\)-fold.}\quad
	For independent observations $Z=(X,Y)$ and $Z'=(X',Y')$, set
	\[
	F_A=\phi^{u\circ}(Z,Z')+\phi^{u\circ}(Z',Z),\qquad
	\zeta^u=\frac{\mathbb I(Y\ne Y')}{\pi_1\pi_0}
	\zeta(X_{\mathrm{case}},X_{\mathrm{control}}),
	\qquad \chi=\chi_{YY'}(X,X'),
	\]
	with $\phi^{u\circ}(Z,Z'):=\phi^{u\circ}_{Y,Y'}(X,X')$, which includes the factor $1/\pi_y$ for the validated observation's class. By Step~4,
	\[
	\begin{gathered}
		\mathbb E(F_A^2)=2(\tau_A^2+\kappa_A),\quad
		\mathbb E(\chi^2)=\sigma_{\chi,A}^2,\quad
		\mathbb E(F_A\chi)=2\xi_A,\\
		\mathbb E\{(\zeta^u)^2\}=\frac{2\varsigma^2}{\pi_1\pi_0},\quad
		\mathbb E(F_A\zeta^u)=2(\omega_A+\omega_A'),\quad
		\mathbb E(\chi\zeta^u)=2\omega_A''.
	\end{gathered}
	\]
	In one partition, a pair that shares a fold has \(v^A_{ab}=K/(n_1n_0)\) if it is a case--control pair, \(c_{ab}=c_{ba}=0\), and \(d_{ab}=(K-1)/(Km^2)\), and a pair that does not has \(v^A_{ab}=0\), \(c_{ab}=c_{ba}=1/(nm)\), and \(d_{ab}=(K-2)/(Km^2)\). Replacing \(n_1n_0\) by \(n^2\pi_1\pi_0\) changes the variance only by \(O(n^{-3})\), since \(\mathbb E\{1/(\hat\pi(1-\hat\pi))\mid\mathcal E_{n,K}\}=1/(\pi_1\pi_0)+O(n^{-1})\). By \eqref{eq:auc_exact_pair_kernel} and Step~3, \(n^2\) times the pair's component is then, up to a term negligible in mean square,
	\[
	K\zeta^u+\frac{K}{K-1}\chi
	\quad\text{if the pair shares a fold},
	\qquad
	\frac{K}{K-1}F_A+\{1-(K-1)^{-2}\}\chi
	\quad\text{otherwise},
	\]
	with limiting probabilities $1/K$ and $(K-1)/K$. The average of the two configurations is $F_A+\chi+\zeta^u$, and their variance is $(K-1)^{-1}\{F_A-\chi/(K-1)-(K-1)\zeta^u\}^2$. For \(R\) independent partitions the pair's weights average \(R\) independent configurations, so the average is unchanged and the variance is divided by \(R\). Summing over the \(n^2/2+O(n)\) pairs, the pair part is \(\{\tfrac12\mathbb E(F_A+\chi+\zeta^u)^2+G_A(K)/R\}/n^2+o(n^{-2})\), with
	\begin{equation}
		\label{eq:auc_split_expected_square_endpoint}
		G_A(K)=\frac{1}{2(K-1)}\mathbb E
		\left\{\left(F_A-\frac{\chi}{K-1}-(K-1)\zeta^u\right)^2\right\}\ge0.
	\end{equation}
	With \(\varsigma^2\) the limiting variance of the pair-specific part \(\zeta^{(2)}\), \(\omega_A\), \(\omega_A'\), \(\omega_A''\) its covariances with the training effects on the case, the control and the pair, and \(\tau_A^2\), \(\kappa_A\), \(\sigma_{\chi,A}^2\), \(\xi_A\) the pairwise-metric analogues of the pair constants, let
	\[
	\lambda_1:=\tau_A^2+\kappa_A+2\omega_A'',\qquad
	\lambda_2:=-2\xi_A,\qquad \lambda_3:=\frac{\sigma_{\chi,A}^2}{2},\qquad
	\lambda_4:=\frac{\varsigma^2}{\pi_1\pi_0}.
	\]
	Expanding the square with the moments above gives
	\[
	G_A(K)=\frac{\lambda_1}{K-1}+\frac{\lambda_2}{(K-1)^2}+\frac{\lambda_3}{(K-1)^3}+\lambda_4(K-1)-2(\omega_A+\omega_A').
	\]
	Setting \(\varsigma^2\), \(\omega_A\), \(\omega_A'\) and \(\omega_A''\) to zero leaves \(G(K)\) of Section~\ref{subsec:loo} at the pairwise-metric analogues of the pair constants; the rest comes from the pair-specific part, its variance contributing \(\lambda_4(K-1)\) and its covariances with the training effects the \(\omega\) terms. The first part, \(\tfrac12\mathbb E(F_A+\chi+\zeta^u)^2\), does not depend on \(K\).
	
	\noindent\emph{Step 7: The pair part of leave-pair-out.}\quad
	At leave-pair-out the pair coefficient is $1/(n_1n_0)$, and the normalized validation--training and training--training coefficients are $n^{-2}+O(n^{-3})$. So, by the argument of Step~3, $n^2$ times the pair's component tends to $F_A+\chi+\zeta^u$, and the pair part is $\tfrac12\mathbb E(F_A+\chi+\zeta^u)^2/n^2+o(n^{-2})$.
	
	\noindent\emph{Step 8: The difference.}\quad
	Leave-pair-out has no split noise, and its first-order part is its first-order sampling variance \(\mathcal W_{1,n}(\mathrm{LPO})\) (Steps 4 and 5 of Section~\ref{appendix:proof_split_lotv_auc}). The parts of order three and higher are \(o(n^{-2})\) for both designs by \eqref{eq:auc_third_remainder}. Adding the parts of Steps 5--7,
	\[
	\Var(\widehat R_{K,R})-\Var(\widehat R_{\mathrm{LPO}})=\mathcal W_{1,n}(K)-\mathcal W_{1,n}(\mathrm{LPO})+\frac{G_A(K)}{Rn^2}+o(n^{-2}),
	\]
	with no expansion of \(\mathcal W_{1,n}\) in \(m\), which is \eqref{eq:lpo_endpoint_variance_gap}.
\end{proof}

\paragraph{Paired \(K\)-fold.}
\begin{proof}
	\noindent\emph{Step 1: The weights.}\quad
	The cases are split at random into folds \(C_1,\dots,C_K\) of size \(n_1/K\) and the controls into folds \(D_1,\dots,D_K\) of size \(n_0/K\). Split \((a,b)\) holds out \(C_a\cup D_b\), so every training set has \(m_y=pn_y\) observations of class \(y\), as under stratified \(K\)-fold. Every case--control pair is evaluated in exactly one of the \(K^2\) splits, so \(M_A=n_1n_0\) and every case--control pair has \(v^A_{ab}=1/(n_1n_0)\). Every observation is validated in \(K\) splits, with \(\tilde v_i=1/n_{Y_i}\), and trains in \(K(K-1)\) of the \(K^2\) splits, with \(w_i=1/n\). For a case \(a\) and a control \(b\), \(b\) trains in \(K-1\) of the \(K\) splits that validate \(a\), \(a\) trains in \(K-1\) of the \(K\) splits that validate \(b\), and both train in \((K-1)^2\) of the \(K^2\) splits, so, in the normalization of Step~6,
	\[
	c_{ab}=c_{ba}=\frac{K-1}{K}\,\frac1{nm},
	\qquad
	d_{ab}=\frac{(K-1)^2}{K^2m^2}.
	\]
	Two observations of the same class either share a fold, with \(c_{ab}=c_{ba}=0\) and \(d_{ab}=(K-1)/(Km^2)\), or do not, with \(c_{ab}=c_{ba}=1/(nm)\) and \(d_{ab}=(K-2)/(Km^2)\), and they share a fold with probability \(1/K+O(n^{-1})\).
	
	\noindent\emph{Step 2: The case--control pairs add no split noise.}\quad
	Under stratified \(K\)-fold a case--control pair shares a fold with probability \(1/K+O(n^{-1})\), and averaging the weights of Step~6 over the two configurations gives
	\[
	\mathbb E_I(v^A_{ab})=\frac1{n_1n_0},
	\qquad
	\mathbb E_I(c_{ab})=\frac{K-1}{K}\,\frac1{nm},
	\qquad
	\mathbb E_I(d_{ab})=\frac{(K-1)^2}{K^2m^2},
	\]
	up to \(O(n^{-1})\) relative error. These are the weights of Step~1. Under paired \(K\)-fold the weights of a case--control pair are therefore fixed at their averages over \(I\), so its component does not vary with the split assignment and contributes to the sampling variance only. The first-order weights are those of stratified \(K\)-fold, and so are the averaged weights of every order, since both designs are class-stratified with the same training composition and do not distinguish observations within a class. The two designs therefore have the same sampling variance, with first-order part \(\mathcal W_{1,n}(K)\) by Step~5.
	
	\noindent\emph{Step 3: The same-class pairs.}\quad
	A same-class pair has no pair-specific validation term, so \(\zeta^u=0\) in Step~6. Its two configurations and their probabilities are those of one stratified partition, so by Step~6 its weighted component, times \(n^2\), is
	\[
	\frac{K}{K-1}\,\chi
	\quad\text{if the pair shares a fold},
	\qquad
	\frac{K}{K-1}\,F_A+\{1-(K-1)^{-2}\}\chi
	\quad\text{otherwise},
	\]
	with average \(F_A+\chi\) and variance \((K-1)^{-1}\{F_A-\chi/(K-1)\}^2\). With \(R\) independent assignments the variance is divided by \(R\). Summing over the pairs as in Step~6, the second-order split noise is \(G_s(K)/(Rn^2)+o(n^{-2})\), with
	\[
	G_s(K):=\frac{1}{2(K-1)}\,\mathbb E\Big[\Big\{F_A-\frac{\chi}{K-1}\Big\}^2\,\mathbb I(Y=Y')\Big]
	=\frac{\tau_s^2+\kappa_s}{K-1}-\frac{2\xi_s}{(K-1)^2}+\frac{\sigma_{\chi,s}^2}{2(K-1)^3},
	\]
	where \(\tau_s^2+\kappa_s:=\tfrac12\mathbb E\{F_A^2\,\mathbb I(Y=Y')\}\), \(\xi_s:=\tfrac12\mathbb E\{F_A\chi\,\mathbb I(Y=Y')\}\), and \(\sigma_{\chi,s}^2:=\mathbb E\{\chi^2\,\mathbb I(Y=Y')\}\) are the same-class parts of the constants of Step~4. This is the function \(G\) of Section~\ref{subsec:loo} at these constants.
	
	\noindent\emph{Step 4: The difference.}\quad
	The parts of order three and higher are \(o(n^{-2})\) by \eqref{eq:auc_third_remainder}. With Steps 2 and 3 and Step~7,
	\[
	\Var\big(\widehat R_{K,R}^{\mathrm{paired}}\big)-\Var(\widehat R_{\mathrm{LPO}})=\mathcal W_{1,n}(K)-\mathcal W_{1,n}(\mathrm{LPO})+\frac{G_s(K)}{Rn^2}+o(n^{-2}),
	\]
	which is \eqref{eq:paired_gap}.
	
	\noindent\emph{Step 5: Comparison with stratified \(K\)-fold.}\quad
	Splitting the expectation in \eqref{eq:auc_split_expected_square_endpoint} by whether the two observations share a class,
	\[
	G_A(K)-G_s(K)=\frac{1}{2(K-1)}\,\mathbb E\Big[\Big\{F_A-\frac{\chi}{K-1}-(K-1)\zeta^u\Big\}^2\,\mathbb I(Y\ne Y')\Big]\ge0,
	\]
	so with the same number of repetitions paired \(K\)-fold has the smaller variance. Expanding the square with the moments of Step~6,
	\[
	G_A(K)-G_s(K)=\lambda_4(K-1)-2(\omega_A+\omega_A')+O\{(K-1)^{-1}\},
	\]
	and
	\[
	(K-1)\,G_s(K)=\tau_s^2+\kappa_s+O\{(K-1)^{-1}\}.
	\]
	One paired run and stratified \(K\)-fold repeated \(K\) times both take \(K^2\) fits, and the first has the smaller variance when \(G_s(K)<G_A(K)/K\), that is, when \((K-1)\,G_s(K)<G_A(K)-G_s(K)\). When \(\lambda_4>0\) this holds for all large enough \(K\). When \(\xi_s\le0\), each term of \(G_s\) is nonnegative and does not increase in \(K\), as in Step~4 of Section~\ref{app:loo_corollaries}.
\end{proof}

\subsection{Derivation of \eqref{eq:auc_rate_one} and proofs of the fold-count statements of Section~\ref{subsec:folds_auc}}
\label{app:endpoint_auc}

Under expansions of the expected metric and the first-order constants in \(1/m\), stated below,
\begin{equation}
	\label{eq:auc_rate_one}
	\mathcal W_{1,n}(K)-\mathcal W_{1,n}(\mathrm{LPO})=\frac{b_A/(K-1)-\ell_A}{n^2}+o(n^{-2}),
\end{equation}
where \(b_A\) is the finite-training correction to the first-order sampling variance and \(\ell_A\) the correction for the class composition of leave-pair-out, which removes one case and one control where class-stratified \(K\)-fold removes the same fraction of each class. The correction \(\ell_A\) is a change in the sampling variance, not split noise. For the AUC, \(\ell_A\) and the part of \(b_A\) that comes from the variance of the target across class counts are zero when the limiting fitted ranking does not depend on the training class proportion, a property of the model, not of the metric. The variance of leave-pair-out does not depend on \(K\), so by \eqref{eq:lpo_endpoint_variance_gap} and \eqref{eq:auc_rate_one} repeated \(K\)-fold depends on \(K\) only through \(b_A/(K-1)+G_A(K)/R\); at a slower rate \(m^{-\gamma}\), \(0<\gamma<1\), the sign of \(b_A\) determines which fold count has the smaller variance, as that of \(b\) does for unit-level metrics.

\paragraph{The first-order sampling variance of \(K\)-fold.}
Write \(\pi:=\pi_1\), \(\sigma_\pi^2:=\pi(1-\pi)\), and use \(\mu_m\), \(\mathcal Q_{y,m}\), and \(\mathcal Q_m\) of Section~\ref{appendix:proof_pairwise_second}. Suppose that, for some \(0<\gamma\le1\) and uniformly for \(t\) near \(\pi\),
\[
\begin{gathered}
	\mu_m(t)=\mu_\infty(t)+\beta_A(t)m^{-\gamma}+o(m^{-\gamma}),\\
	\mathcal Q_{y,m}(t_y)=\mathcal Q_y(t)+b_y(t)m^{-\gamma}+o(m^{-\gamma}),
\end{gathered}
\]
where the remainders and their first derivatives in \(t\) are \(o(m^{-\gamma})\), and the grid-indexed functions extend to a neighbourhood of \(\pi\) as in Section~\ref{appendix:proof_pairwise_second}. Suppose also that \(\mu_\infty\), \(\beta_A\), \(\mathcal Q_y\), and \(b_y\) are continuously differentiable near \(\pi\), that the second derivative of \(\mu_m\) in \(t\) is bounded near \(\pi\) uniformly in \(m\), and that these functions have polynomial bounds away from \(\pi\), whose contribution is then negligible on \(\mathcal E_{n,K}\) because \(\hat\pi\) leaves the neighbourhood with exponentially small probability.

\begin{proof}
	\noindent\emph{Step 1: The within-class variance.}\quad
	By the expansions of \(\mathcal Q_{y,m}\),
	\[
	\mathcal Q_m(t)=\mathcal Q_\infty(t)+b_Q(t)m^{-\gamma}+o(m^{-\gamma}),
	\]
	with
	\[
	\mathcal Q_\infty(t)=\frac{\mathcal Q_1(t)}{t}+\frac{\mathcal Q_0(t)}{1-t},
	\qquad
	b_Q(t)=\frac{b_1(t)}{t}+\frac{b_0(t)}{1-t}.
	\]
	
	\noindent\emph{Step 2: The expansion.}\quad
	With Step~1, \eqref{eq:auc_W1_exact} gives
	\begin{equation}
		\label{eq:auc_first_component_expansion}
		\mathcal W_{1,n}(K)=\Var\{\mu_\infty(\hat\pi)\}+\frac1n\mathbb E\{\mathcal Q_\infty(\hat\pi)\}+\frac{b_A}{n\,m^{\gamma}}+o(n^{-1-\gamma}),
	\end{equation}
	where
	\begin{equation}
		\label{eq:auc_bA_formula}
		b_A:=b_Q(\pi)+2\sigma_\pi^2\mu_\infty'(\pi)\beta_A'(\pi).
	\end{equation}
	
	\noindent\emph{Step 3: The two terms of \(b_A\).}\quad
	The first term of \(b_A\) comes from \(n^{-1}\mathbb E\{\mathcal Q_m(\hat\pi)\}\), since \(b_Q\) is continuous. The second comes from \(2m^{-\gamma}\Cov\{\mu_\infty(\hat\pi),\beta_A(\hat\pi)\}\), with
	\[
	\Cov\{\mu_\infty(\hat\pi),\beta_A(\hat\pi)\}=\frac{\sigma_\pi^2\mu_\infty'(\pi)\beta_A'(\pi)}{n}+o(n^{-1}).
	\]
	The term \(m^{-2\gamma}\Var\{\beta_A(\hat\pi)\}\) is \(O(n^{-1-2\gamma})\), and the remainders contribute \(o(n^{-1-\gamma})\) by Cauchy--Schwarz and the derivative bounds.
	
	\noindent\emph{Step 4: Two fold counts.}\quad
	The first two terms of \eqref{eq:auc_first_component_expansion} do not depend on \(K\), so the first-order sampling variances of two fold counts, at training fractions \(p_1\) and \(p_2\), differ by
	\[
	\frac{b_A(p_1^{-\gamma}-p_2^{-\gamma})}{n^{1+\gamma}}+o(n^{-1-\gamma}).
	\]
\end{proof}

\paragraph{The first-order sampling variance of leave-pair-out.}
\begin{proof}
	\noindent\emph{Step 1: The exact form.}\quad
	Leave-pair-out trains at $m=n-2$ with class fraction
	\[
	\tilde t=\frac{n_1-1}{n-2}=\hat\pi+\frac{2\hat\pi-1}{n-2},
	\qquad \tilde t_1=\tilde t,\quad \tilde t_0=1-\tilde t,
	\]
	and validates a class-$y$ observation at weight $1/n_y$, so by Step~4 of Section~\ref{appendix:proof_split_lotv_auc}
	\begin{equation}
		\label{eq:lpo_W1_exact}
		\mathcal W_{1,n}(\mathrm{LPO})
		=\Var\{\mu_m(\tilde t)\}
		+\frac1n\mathbb E\left\{\frac{\mathcal Q_{1,m}(\tilde t_1)}{\hat\pi}+
		\frac{\mathcal Q_{0,m}(\tilde t_0)}{1-\hat\pi}\right\}.
	\end{equation}
	
	\noindent\emph{Step 2: The rate \(\gamma<1\).}\quad
	At \(\gamma<1\), putting \(\hat\pi\) for \(\tilde t\) changes this by \(O(n^{-2})\). In the conditional-variance term the change is \(n^{-1}\) times \(O(n^{-1})\), since \(\mathcal Q_{y,m}\) has a bounded derivative and \(\tilde t-\hat\pi=O(n^{-1})\). In the mean term, \(\mu_m(\tilde t)-\mu_m(\hat\pi)\), as a function of \(\hat\pi\), is \(O(n^{-1})\) with derivative \(O(n^{-1})\), by the bounded second derivative of \(\mu_m\). Its variance is therefore \(O(n^{-3})\), and \(\Var\{\mu_m(\tilde t)\}-\Var\{\mu_m(\hat\pi)\}=O(n^{-2})\) by Cauchy--Schwarz. With \(\hat\pi\) in place of \(\tilde t\), \(\mathcal W_{1,n}(\mathrm{LPO})\) is \eqref{eq:auc_W1_exact} at \(m=n-2\). With \((n-2)^{-\gamma}=n^{-\gamma}+O(n^{-1-\gamma})\), \eqref{eq:auc_first_component_expansion} then gives
	\[
	\mathcal W_{1,n}(K)-\mathcal W_{1,n}(\mathrm{LPO})=\frac{b_A\,(p^{-\gamma}-1)}{n^{1+\gamma}}+o(n^{-1-\gamma}),
	\]
	and the class-composition correction is of smaller order.
	
	\noindent\emph{Step 3: The rate \(\gamma=1\).}\quad
	At \(\gamma=1\), suppose also that \(\mu_\infty\) is twice and \(\mathcal Q_y\) once continuously differentiable near \(\pi\). The shift of the class fraction is then of the order of the training correction. Expanding in \(\tilde t-\hat\pi\),
	\[
	\begin{gathered}
		\mu_m(\tilde t)=\mu_\infty(\hat\pi)+\frac{\beta_A(\hat\pi)+(2\hat\pi-1)\mu_\infty'(\hat\pi)}{n}+o(n^{-1}),\\
		\mathcal Q_{y,m}(\tilde t_y)=\mathcal Q_y(\hat\pi)+\frac{b_y(\hat\pi)+(2\hat\pi-1)\mathcal Q_y'(\hat\pi)}{n}+o(n^{-1}).
	\end{gathered}
	\]
	The covariance of the two terms of the first, by the delta method, and the second, with the observed validation denominators \(\hat\pi\) and \(1-\hat\pi\) retained, give
	\begin{equation}
		\label{eq:lpo_W1_expansion}
		\mathcal W_{1,n}(\mathrm{LPO})
		=\Var\{\mu_\infty(\hat\pi)\}+\frac1n\mathbb E\{\mathcal Q_\infty(\hat\pi)\}+\frac{b_A+\ell_A}{n^2}+o(n^{-2}),
	\end{equation}
	where
	\begin{equation}
		\label{eq:lpo_class_composition_correction}
		\begin{split}
			\ell_A={}&(2\pi-1)
			\left\{\frac{\mathcal Q_1'(\pi)}\pi+\frac{\mathcal Q_0'(\pi)}{1-\pi}\right\}\\
			&+2\sigma_\pi^2\mu_\infty'(\pi)
			\left\{2\mu_\infty'(\pi)+(2\pi-1)\mu_\infty''(\pi)\right\}.
		\end{split}
	\end{equation}
	
	\noindent\emph{Step 4: The difference.}\quad
	Subtracting \eqref{eq:lpo_W1_expansion} from \eqref{eq:auc_first_component_expansion} at \(\gamma=1\) and \(m=pn\), with \(1/p-1=1/(K-1)\), gives \eqref{eq:auc_rate_one}.
\end{proof}

\paragraph{The interior optimum.}
\begin{proof}
	\noindent\emph{Step 1: The part that depends on \(K\).}\quad
	The variance of leave-pair-out does not depend on \(K\), so by \eqref{eq:lpo_endpoint_variance_gap} and \eqref{eq:auc_rate_one} the part of \(n^2\Var(\widehat R_{K,R})\) that depends on \(K\) is \(b_A/(K-1)+G_A(K)/R+o(1)\). Multiplying it by \(R\) and dropping the constant \(-2(\omega_A+\omega_A')\) gives
	\[
	F_R(K)=\frac{\lambda_{1,R}}{K-1}+\frac{\lambda_2}{(K-1)^2}+\frac{\lambda_3}{(K-1)^3}+\lambda_4(K-1),
	\qquad
	\lambda_{1,R}:=\lambda_1+Rb_A.
	\]
	
	\noindent\emph{Step 2: The stationary point.}\quad
	Its derivative in $K$ is zero exactly at $\lambda_4(K-1)^4 = \lambda_{1,R}(K-1)^2 + 2\lambda_2(K-1) + 3\lambda_3$. If \(\lambda_4>0\) and \(\lambda_{1,R}\), \(\lambda_2\), \(\lambda_3\) are nonnegative and not all zero, this has one root \(K_R^\ast>1\).
	
	\noindent\emph{Step 3: The minimum.}\quad
	The second derivative is
	\[
	F_R''(K)=\frac{2\lambda_{1,R}}{(K-1)^3}+\frac{6\lambda_2}{(K-1)^4}+\frac{12\lambda_3}{(K-1)^5}>0,
	\]
	so the minimum over $K \ge 2$ is at $\max\{2, K_R^\ast\}$, and $K_R^\ast > 2$ exactly when $\lambda_4 < \lambda_{1,R} + 2\lambda_2 + 3\lambda_3$. Admissible fold counts must leave both classes represented in every validation fold.
\end{proof}

\paragraph{Ranking invariance.}
Suppose the limiting ranking functional is independent of the training class proportion, locally in the two class-conditional training laws, that is, $f^\star_{t,P_0,P_1}=\mathcal M_{t,P_0,P_1}\circ \mathsf s_{P_0,P_1}$, with $\mathcal M_{t,P_0,P_1}$ strictly increasing and the ranking function $\mathsf s_{P_0,P_1}$ free of $t$, locally in $t,P_0,P_1$.

\begin{proof}
	\noindent\emph{Step 1: The influences do not depend on the class fraction.}\quad
	The validation influences $g_y$ and the limiting AUC $\mu_\infty$ then do not depend on $t$. If $\Delta_y$ denotes the within-class training effect on limiting AUC, it also does not depend on $t$, and
	\[
	\bar\phi_{y,\infty,t}=\Delta_y/t_y,
	\qquad
	g_{y,\infty,t}+t_y\bar\phi_{y,\infty,t}=g_y+\Delta_y.
	\]
	
	\noindent\emph{Step 2: The corrections are zero.}\quad
	Thus $\mathcal Q_y'(t)=0$ and $\mu_\infty'(t)=0$, every term in \eqref{eq:lpo_class_composition_correction} is zero, and so is the product term of \eqref{eq:auc_bA_formula}. This proves the ranking-invariance statement at the start of this section.
\end{proof}

\paragraph{The exhaustive endpoint given the labels.}
\begin{proof}
	\noindent\emph{Step 1: The argument carries over.}\quad
	The argument for exhaustive designs in Section~\ref{app:loo_corollaries} applies given the labels, with Section~\ref{appendix:proof_split_lotv_auc} in place of Section~\ref{appendix:proof_split_lotv}. The designs compared are class-stratified and fixed-size, each validation set is the complement of its training set, and permutations within each class leave the distribution of \(I\) given \(Y\) unchanged. Given \(Y\), the conditional sampling variance \(\sum_{\mathcal S}\|\bar h_{\mathcal S}\|^2\) depends on the design only through \((n_1,n_0,m_1,m_0)\), exhaustive validation at a training composition has that conditional sampling variance as its conditional variance, and so does leave-pair-out.
	
	\noindent\emph{Step 2: The comparison.}\quad
	Leave-pair-out therefore has the smallest conditional variance among the designs compared if and only if its conditional sampling variance is no larger than that of every training composition compared.
\end{proof}

\section{Proofs for Section~\ref{sec:normality}}
\label{appendix:proofs_sec5}

\subsection{Proof of Theorem~\ref{thm:clt}}
\label{appendix:proof_clt}

Throughout, \(I\) stands for the inclusion structure, and for the within-fold AUC for the pair \((I,Y)\). Given \(I\), by A1 the estimate is a function of independent inputs \(U_1,\dots,U_n\), the observations, or for the AUC the covariates, with \(X_j\) distributed as \(X\) given \(Y_j\). Let \(U'\) and \(U''\) be independent copies of \(U=(U_1,\dots,U_n)\) given \(I\). A point of evaluation takes each coordinate \(j\) from \(U_j\), \(U_j'\), or \(U_j''\), and at such a point \(\nabla_j\) replaces coordinate \(j\) by one of its other two copies. Steps 1--9 prove the general statement. Steps 10--13 show that A2(i) and A5 give its conditions, and identify \(V\).

\paragraph{The general statement.}
\begin{proof}
	\noindent\emph{Step 1: Setting aside the remainder.}\quad
	Let
	\[
	\Delta_n:=\big\{\widehat R_D-\mathbb E(\widehat R_D\mid I)\big\}-\big\{\widetilde R_D-\mathbb E(\widetilde R_D\mid I)\big\}.
	\]
	By the condition of (a), \(n\Var(\Delta_n\mid I)\to0\), so \(\sqrt n\,\Delta_n\to0\) in probability, given \(I\) and unconditionally, by Chebyshev's inequality. By the Cauchy--Schwarz inequality,
	\[
	\big|n\Var(\widetilde R_D\mid I)-n\Var(\widehat R_D\mid I)\big|\le n\Var(\Delta_n\mid I)+2\big\{n\Var(\Delta_n\mid I)\,n\Var(\widehat R_D\mid I)\big\}^{1/2}\to0,
	\]
	so \(n\Var(\widetilde R_D\mid I)\to V\). By Slutsky's lemma it suffices to show that \(\sqrt n\{\widetilde R_D-\mathbb E(\widetilde R_D\mid I)\}\Rightarrow N(0,V)\) given \(I\).
	
	\noindent\emph{Step 2: The degenerate limit.}\quad
	If \(V=0\), Chebyshev's inequality gives \(\sqrt n\{\widetilde R_D-\mathbb E(\widetilde R_D\mid I)\}\to0\), which is the limit \(N(0,0)\). From here on \(V>0\). Write
	\[
	f:=\widetilde R_D-\mathbb E(\widetilde R_D\mid I),\qquad \sigma^2:=\Var(f\mid I),
	\]
	so that \(n\sigma^2\to V\), and \(\sigma^2\ge V/(2n)\) outside an event whose probability tends to zero. The centering does not depend on the inputs, so \(\nabla_jf=\nabla_j\widetilde R_D\) and \(\nabla_j\nabla_kf=\nabla_j\nabla_k\widetilde R_D\).
	
	\noindent\emph{Step 3: Chatterjee's bound.}\quad
	For a set \(A\) of indices, let \(U^A\) be \(U\) with the coordinates in \(A\) taken from \(U'\). For \(A\) a proper subset of \(\{1,\dots,n\}\), let
	\[
	c_A:=\Big\{\binom n{|A|}(n-|A|)\Big\}^{-1},\qquad
	T:=\frac12\sum_{j=1}^n\sum_{A\not\ni j}c_A\,\nabla_jf(U)\,\nabla_jf(U^A),
	\]
	with the replacement in each \(\nabla_j\) taken from \(U'\). Theorem~2.2 of \citet{chatterjee2008} asks only that the inputs be independent and \(f\) square-integrable. It gives \(\mathbb E(T\mid I)=\sigma^2\) and
	\[
	d_W\Big(\frac f\sigma,\,N(0,1)\Big)\le\frac{\{\Var(T\mid I)\}^{1/2}}{\sigma^2}+\frac1{2\sigma^3}\sum_j\mathbb E\big\{|\nabla_jf(U)|^3\,\big|\,I\big\},
	\]
	with \(d_W\) the Wasserstein distance between the law of \(f/\sigma\) given \(I\) and \(N(0,1)\).
	
	\noindent\emph{Step 4: The weights.}\quad
	For each \(j\),
	\[
	\sum_{A\not\ni j}c_A=\sum_{a=0}^{n-1}\binom{n-1}a\binom na^{-1}\frac1{n-a}=\sum_{a=0}^{n-1}\frac1n=1,
	\]
	so the weights \(c_A\) over the sets \(A\) not containing \(j\) are a probability distribution.
	
	\noindent\emph{Step 5: Efron--Stein.}\quad
	\(T\) is a function of the \(2n\) independent coordinates of \((U,U')\). For a coordinate \(s\), let \(T_s\) be \(T\) with that coordinate taken from \(U''\). By the Efron--Stein inequality,
	\[
	\Var(T\mid I)\le\frac12\sum_s\mathbb E\big\{(T-T_s)^2\,\big|\,I\big\}.
	\]
	Each index \(k\) owns two coordinates, \(U_k\) and \(U_k'\).
	
	\noindent\emph{Step 6: The change from one coordinate.}\quad
	Fix a coordinate \(s\) owned by \(k\), write
	\[
	g_{j,A}:=\nabla_jf(U)\,\nabla_jf(U^A),
	\]
	and mark with a subscript \(s\) the value after the replacement. For \(j=k\) both factors can change, and
	\[
	|g_{k,A}-g_{k,A,s}|\le|g_{k,A}|+|g_{k,A,s}|,
	\]
	two products of first differences in \(k\). For \(j\ne k\),
	\[
	g_{j,A}-g_{j,A,s}=\big\{\nabla_jf(U)-\nabla_jf(U)_s\big\}\,\nabla_jf(U^A)+\nabla_jf(U)_s\,\big\{\nabla_jf(U^A)-\nabla_jf(U^A)_s\big\}.
	\]
	Each brace is zero when the coordinate does not enter the evaluation, and otherwise it is a second difference \(\nabla_j\nabla_kf\) at a point of evaluation. So each term is a second difference times a first difference, each at its own point.
	
	\noindent\emph{Step 7: Squaring.}\quad
	Let \(\alpha_k\) be the largest value of \(\mathbb E\{(\nabla_kf)^4\mid I\}\), and \(\beta_k\) the largest value of \(\mathbb E[\{\sum_{j\ne k}|\nabla_j\nabla_kf|\,|\nabla_jf|\}^2\mid I]\), over the points of evaluation, with each difference at its own point. These are the summands of the theorem's first and third conditions. Split \(T-T_s=\Delta_1+\Delta_2\), with \(\Delta_1\) the terms \(j=k\) and \(\Delta_2\) the rest. By Step 6, Jensen's inequality over the weights of Step 4, and the Cauchy--Schwarz inequality,
	\[
	\Delta_1^2\le\frac12\sum_{A\not\ni k}c_A\big(g_{k,A}^2+g_{k,A,s}^2\big),\qquad
	\mathbb E(\Delta_1^2\mid I)\le\alpha_k.
	\]
	For \(\Delta_2\), let \(A_j\), \(j\ne k\), be independent random sets, \(A_j\) with the law \(c_A\) of Step 4. Then \(\Delta_2\) is the expectation over the \(A_j\) of \(\frac12\sum_{j\ne k}(g_{j,A_j}-g_{j,A_j,s})\). By Step 6 this sum is at most \(\frac12(P+Q)\) in absolute value, with \(P\) and \(Q\) sums over \(j\ne k\) of a second difference times a first difference, each at its own point. By Jensen's inequality over the \(A_j\) and \((a+b)^2\le2a^2+2b^2\),
	\[
	\mathbb E(\Delta_2^2\mid I)\le\frac12\,\mathbb E\big(P^2+Q^2\,\big|\,I\big)\le\beta_k.
	\]
	Hence \(\mathbb E\{(T-T_s)^2\mid I\}\le2(\alpha_k+\beta_k)\), and summing over the two coordinates of each \(k\) in Step 5,
	\[
	\Var(T\mid I)\le2\sum_k(\alpha_k+\beta_k).
	\]
	
	\noindent\emph{Step 8: Normality given \(I\).}\quad
	By Step 2, \(\sigma^2\ge V/(2n)\). By the theorem's conditions,
	\begin{gather*}
		\frac{\{\Var(T\mid I)\}^{1/2}}{\sigma^2}\le\frac{2n}{V}\,\Big\{2\sum_k(\alpha_k+\beta_k)\Big\}^{1/2}=\frac{2n}{V}\,o(n^{-1})=o(1),\\
		\frac1{2\sigma^3}\sum_j\mathbb E\big\{|\nabla_jf|^3\,\big|\,I\big\}\le\frac{(2n)^{3/2}}{2V^{3/2}}\,o(n^{-3/2})=o(1),
	\end{gather*}
	in probability. So \(f/\sigma\Rightarrow N(0,1)\) given \(I\), and with \(n\sigma^2\to V\) and Step 1, \(\sqrt n\{\widehat R_D-\mathbb E(\widehat R_D\mid I)\}\Rightarrow N(0,V)\) given \(I\).
	
	\noindent\emph{Step 9: Removing the conditioning.}\quad
	Let \(W_1:=\sqrt n\{\widehat R_D-\mathbb E(\widehat R_D\mid I)\}\). The real and imaginary parts of \(x\mapsto e^{itx}\) are Lipschitz, so by Step 8
	\[
	\big|\mathbb E(e^{itW_1}\mid I)-e^{-t^2V/2}\big|\to0
	\]
	in probability. The left side is at most \(2\), so \(\mathbb E(e^{itW_1})\to e^{-t^2V/2}\) by dominated convergence, which is the unconditional statement. Let \(W_2:=\sqrt n\{\mathbb E(\widehat R_D\mid I)-R_{p,n}\}\), a function of \(I\). If \(W_2\Rightarrow N(0,V')\), then by the same bound
	\[
	\mathbb E\big\{e^{it(W_1+W_2)}\big\}=\mathbb E\big\{e^{itW_2}\,\mathbb E(e^{itW_1}\mid I)\big\}\to e^{-t^2V/2}\,e^{-t^2V'/2},
	\]
	and \(\sqrt n(\widehat R_D-R_{p,n})\Rightarrow N(0,V+V')\). For a unit-level metric under a fixed-size design, \(\mathbb E(\widehat R_D\mid I)=R_{p,n}\), so \(W_2=0\). For the bootstrap, centered at \(\mathbb E(\widehat R_D)\), A3(ii) and Chebyshev's inequality give \(W_2\to0\) in probability. Step~13 gives \(W_2\) for the AUC.
	
\end{proof}

\paragraph{Under A2(i) and A5.}
\begin{proof}
	\noindent\emph{Step 10: The first-order representation.}\quad
	In each case, given \(I\),
	\begin{equation}\label{eq:F1_rep}
		\widehat R_D-\mathbb E(\widehat R_D\mid I)=\sum_jL_j(U_j)+r_D,\qquad n\,\mathbb E(r_D^2\mid I)\to0\ \text{in probability},
	\end{equation}
	where the \(L_j\) have mean zero and \(|L_j|\le a_je_j\), with \(a_j\ge0\) a function of \(I\) and envelopes \(e_j\) whose squares are uniformly integrable over \(n\).
	
	For a unit-level metric and a fixed-size design, Theorem~\ref{thm:dual_role} gives \(L_j=v_j(\ell_j^\star-R_{p,n})+w_j\bar\phi(Z_j)\) and \(r_D=\varrho_D\), with \(\mathbb E(\varrho_D^2\mid I)=\rho_n^{(2+)}(I)\), which A2(i) and Markov's inequality make \(o(n^{-1})\) in probability. Take \(a_j:=\max(v_j,w_j)\) and \(e_j:=|\ell_j^\star-R_{p,n}|+|\bar\phi(Z_j)|\). By A5 the influences converge in \(L_2\), so the \(e_j^2\) are uniformly integrable. For the bootstrap, A3(i) gives the same, with \(\ell_j^\star-\mathbb E(\ell_j^\star)\) in place of \(\ell_j^\star-R_{p,n}\) and the influences of A3, to which A5 applies.
	
	For the within-fold AUC, given \((I,Y)\), Step~1 of the proof of Corollary~\ref{prop:pairwise_reduction} gives the components \(h_j(X_j)\) on one observation, with kernels at the training composition of each split. Let \(L_j:=\tilde v_j\,g_{Y_j}(X_j)+w_j\,\bar\phi^\circ_{Y_j}(X_j)\), with the limiting influences \(g_y\) and \(\bar\phi_y^\circ\) of Section~\ref{appendix:pairwise_kernel_mapping}, which have mean zero in each class. By Step~2 of that proof, and its counterpart for stratified \(K\)-fold, \(\|h_j-L_j\|_2\le\bar\varepsilon_n(\tilde v_j+w_j)\) with \(\bar\varepsilon_n\to0\), on an event of probability tending to one. The \(h_j-L_j\) are independent across \(j\), so with \(r_D\) the sum of the \(h_j-L_j\) and of the components on two or more observations,
	\[
	\mathbb E(r_D^2\mid I,Y)\le2\bar\varepsilon_n^2\sum_j(\tilde v_j+w_j)^2+2\sum_{k\ge2}\rho_n^{(k)}(I,Y),
	\]
	which is \(o(n^{-1})\) in probability by A4(i), since \(\sum_j(\tilde v_j+w_j)^2=O_P(n^{-1})\). Take \(a_j:=\max(\tilde v_j,w_j)\) and \(e_j:=|g_{Y_j}(X_j)|+|\bar\phi^\circ_{Y_j}(X_j)|\), which does not depend on \(n\).
	
	\noindent\emph{Step 11: The weights.}\quad
	Let \(s_n^2(I):=\Var(\sum_jL_j\mid I)\). In each case there is a constant \(V_D\) such that, in probability,
	\[
	\text{(D1)}\quad ns_n^2(I)\to V_D\ \text{ and }\ n\textstyle\sum_ja_j^2\ \text{is bounded},\qquad
	\text{(D2)}\quad n^{1/2}\max_ja_j\to0 .
	\]
	For a unit-level metric, \(s_n^2(I)=\sum_j(v_j,w_j)\,\Sigma_m\,(v_j,w_j)^\top\), with \(\Sigma_m\) the covariance matrix of \((\ell_j^\star-R_{p,n},\bar\phi(Z_j))\) at training size \(m\), and \(\Sigma_m\) converges, to \(\Sigma_\infty\) say, by A5.
	\begin{itemize}
		\item[(11a)] Balanced fixed-size designs. Here \(v_j=w_j=1/n\) for every \(I\), so \(ns_n^2(I)=\mathcal V_m\to\mathcal V_\infty\), \(n\sum_j(v_j^2+w_j^2)=2\), and \(n^{1/2}\max_ja_j=n^{-1/2}\), whatever the number of folds and repetitions.
		\item[(11b)] Monte Carlo cross-validation. With validation sets of size \(qn\), \(nv_j=N_j/(Sq)\) and \(nw_j=(S-N_j)/(Sp)\), where \(N_j\sim\mathrm{Binomial}(S,q)\) counts the splits that validate \(j\). So \(n\sum_jv_j^2\), \(n\sum_jw_j^2\), and \(n\sum_jv_jw_j\) are averages over \(j\) of bounded functions of \(N_j\). The vector \((N_1,\dots,N_n)\) is a sum over the \(S\) independent splits of the indicator vectors of uniformly random \(qn\)-subsets, so the joint law of \((N_j,N_{j'})\) differs from the product of its marginals by \(O(n^{-1})\) in total variation, and each average has variance \(O(n^{-1})\). The averages therefore converge in probability to their means, \(1+p/(qS)\), \(1+q/(pS)\), and \(1-1/S\) of \eqref{eq:role_moments}, and
		\[
		ns_n^2(I)\to\sigma_V^2\Big(1+\frac p{qS}\Big)+\sigma_T^2\Big(1+\frac q{pS}\Big)+2\gamma_0\Big(1-\frac1S\Big)=\mathcal V_\infty+\lim_nnQ_D,
		\]
		with the constants at their limits. Since \(v_j\le1/(nq)\) and \(w_j\le1/(np)\), (D1) and (D2) hold.
		\item[(11c)] Bootstrap cross-validation. Here \(v_j=N_j/M\), with \(N_j\) the number of resamples that exclude \(j\) and \(M=\sum_r|V_r|\), and \(w_j=\sum_rs_r\tilde I_{jr}/n\) with \(s_r=|V_r|/M\). By the law of large numbers over the multiplicities, \(M/n\to Se^{-1}\) and \(s_r\to1/S\) in probability. The argument of (11b), with \(\Cov(\tilde I_{jr},\tilde I_{j'r})=-1/n\) and all moments of \(\tilde I_{jr}\) bounded, gives the limits \(1+p/(qS)\), \(1+1/S\), and \(1-1/S\) of \eqref{eq:role_moments}, so \(ns_n^2(I)\to\mathcal V_\infty+\lim_nnQ_D\) with the bootstrap value of \(Q_D\). Also \(nv_j\le nS/M=O_P(1)\), and \(nw_j\le\max_rs_r\sum_r\tilde I_{jr}\), whose maximum over \(j\), a sum of \(S\) multiplicities each \(\mathrm{Binomial}(n,1/n)\), is \(O_P(\log n)\). Hence (D1) and (D2) hold.
		\item[(11d)] The within-fold AUC, given \((I,Y)\). Here \(ns_n^2(I,Y)=n\sum_j\{\tilde v_j^2\sigma_{V\mid Y_j}^2+w_j^2\sigma_{T\mid Y_j}^2+2\tilde v_jw_j\gamma_{0\mid Y_j}\}\). For stratified \(K\)-fold, Step~1 of its proof in Section~\ref{appendix:pairwise_kernel_mapping} gives \(\tilde v_j=1/n_{Y_j}+O(n^{-2})\) and \(w_j=1/n+O(n^{-2})\), so \(ns_n^2\to\sigma_A^2+\pi_1\sigma_{T\mid1}^2+\pi_0\sigma_{T\mid0}^2+2\gamma_A\). For a design independent of the labels, Step~3 of the proof of Corollary~\ref{prop:pairwise_reduction} gives \(\tilde v_j=v_j\pi_{Y_j}^{-1}\{1+O(n^{-1/3})\}\) and \(w_j=w_j^{\mathrm u}\{1+O(n^{-1/3})\}\) uniformly in \(j\), on an event of probability tending to one, with \(v_j\) and \(w_j^{\mathrm u}\) the unit-level weights. Given \(I\) the labels are i.i.d.\ and the unit-level weights do not depend on them, so by the law of large numbers over the labels and (11a)--(11b),
		\[
		ns_n^2(I,Y)\to\sigma_A^2\lim_nn\Gamma_D+(\pi_1\sigma_{T\mid1}^2+\pi_0\sigma_{T\mid0}^2)\lim_nn\Lambda_D+2\gamma_A\lim_nn\Omega_D=:V_D,
		\]
		which for stratified \(K\)-fold is the limit above, since there \(n\Gamma_D=n\Lambda_D=n\Omega_D=1\). In both cases \(\max_ja_j=O_P(n^{-1})\), so (D1) and (D2) hold.
	\end{itemize}
	
	\noindent\emph{Step 12: Truncation.}\quad
	By Steps 10 and 11 there are \(C\) and \(\eta_n\to0\) such that the event \(E_n\) on which
	\[
	|ns_n^2(I)-V_D|\le\eta_n,\qquad n\sum_ja_j^2\le C,\qquad n^{1/2}\max_ja_j\le\eta_n,\qquad n\,\mathbb E(r_D^2\mid I)\le\eta_n
	\]
	has probability tending to one. Let \(\varepsilon_n:=\eta_n^{1/2}\), and let \(\mathcal M(x)\) be the supremum over \(n\), and for the AUC over the two classes, of \(\mathbb E\{e_j^2\,\mathbb I(e_j>x)\}\). Uniform integrability gives \(\mathcal M(x)\to0\) as \(x\to\infty\). Since \(|L_j|\le a_je_j\), on \(E_n\)
	\begin{equation}\label{eq:F1_lindeberg}
		n\sum_j\mathbb E\big\{L_j^2\,\mathbb I(|L_j|>\varepsilon_nn^{-1/2})\,\big|\,I\big\}\le n\sum_ja_j^2\,\mathcal M\Big(\frac{\varepsilon_nn^{-1/2}}{\max_ja_j}\Big)\le C\,\mathcal M(\eta_n^{-1/2})\to0 .
	\end{equation}
	Let
	\[
	t_j:=L_j\,\mathbb I(|L_j|\le\varepsilon_nn^{-1/2})-\mathbb E\big\{L_j\,\mathbb I(|L_j|\le\varepsilon_nn^{-1/2})\,\big|\,I\big\},\qquad
	\widetilde R_D:=\mathbb E(\widehat R_D\mid I)+\sum_jt_j(U_j).
	\]
	Since \(L_j\) has mean zero, \(L_j-t_j\) is \(L_j\,\mathbb I(|L_j|>\varepsilon_nn^{-1/2})\) minus its mean. These are independent across \(j\), so by \eqref{eq:F1_rep} and \eqref{eq:F1_lindeberg}, on \(E_n\),
	\[
	n\Var(\widehat R_D-\widetilde R_D\mid I)\le2n\sum_j\mathbb E\big\{L_j^2\,\mathbb I(|L_j|>\varepsilon_nn^{-1/2})\,\big|\,I\big\}+2n\,\mathbb E(r_D^2\mid I)\to0 .
	\]
	\(\widetilde R_D\) is a sum of one-input terms, so \(\nabla_j\nabla_k\widetilde R_D=0\) for \(j\ne k\), and the third condition holds. At every point of evaluation \(|\nabla_j\widetilde R_D|=|t_j(x)-t_j(x')|\le2\varepsilon_nn^{-1/2}\), and \(\mathbb E\{(\nabla_j\widetilde R_D)^2\mid I\}\le2\Var(t_j\mid I)\le2a_j^2\sup\mathbb E(e_j^2)\). With \(n\sum_ja_j^2\le C\),
	\[
	\sum_j\mathbb E\big\{(\nabla_j\widetilde R_D)^4\,\big|\,I\big\}\le\frac{4\varepsilon_n^2}n\sum_j\mathbb E\big\{(\nabla_j\widetilde R_D)^2\,\big|\,I\big\}\le\frac{8C\varepsilon_n^2\sup\mathbb E(e_j^2)}{n^2}=o(n^{-2}),
	\]
	\[
	\sum_j\mathbb E\big\{|\nabla_j\widetilde R_D|^3\,\big|\,I\big\}\le\frac{2\varepsilon_n}{n^{1/2}}\sum_j\mathbb E\big\{(\nabla_j\widetilde R_D)^2\,\big|\,I\big\}\le\frac{4C\varepsilon_n\sup\mathbb E(e_j^2)}{n^{3/2}}=o(n^{-3/2}).
	\]
	So the conditions of the theorem hold. Finally, by \eqref{eq:F1_rep} and the Cauchy--Schwarz inequality, \(|n\Var(\widehat R_D\mid I)-ns_n^2(I)|\le\eta_n+2\{\eta_n\,ns_n^2(I)\}^{1/2}\) on \(E_n\), so \(n\Var(\widehat R_D\mid I)\to V_D\) and \(V=V_D\). For a unit-level metric, (11a)--(11c) give \(V=\mathcal V_\infty+\lim_nnQ_D\).
	
	\noindent\emph{Step 13: The label part of the AUC.}\quad
	Let \(W_2:=\sqrt n\{\mathbb E(\widehat R_D\mid I,Y)-\mu_m(\pi_1)\}\).
	\begin{itemize}
		\item[(13a)] Stratified \(K\)-fold. Step~3 of its proof in Section~\ref{appendix:pairwise_kernel_mapping} gives \(\mathbb E(\widehat R_D\mid I,Y)=\mu_{m_\star}(\tilde m_1/m_\star)+b_n\), with \(\sqrt n\,b_n\to0\), \(|m_\star-m|=O(1)\), and \(\tilde m_1/m_\star=\hat\pi_1+O(n^{-1})\), and the one-observation condition of Corollary~\ref{prop:pairwise_reduction} gives \(\mu_{m_\star}(\pi_1)-\mu_m(\pi_1)=o(n^{-1/2})\). By A4(iii) and the mean value theorem,
		\[
		W_2=\mu_\infty'(\pi_1)\sqrt n(\hat\pi_1-\pi_1)+o_P(1)\Rightarrow N\big(0,\ \pi_1\pi_0\,\mu_\infty'(\pi_1)^2\big),
		\]
		and here \(n\Lambda_D=1\).
		\item[(13b)] Designs independent of the labels. Step~4 of the proof of Corollary~\ref{prop:pairwise_reduction}, which uses A4(iii) alone, gives
		\[
		W_2=\mu_\infty'(\pi_1)\sqrt n\sum_lw_l^{\mathrm u}(Y_l-\pi_1)+o_P(1).
		\]
		Given \(I\) the labels are i.i.d., so the sum has independent bounded terms with mean zero, and \(n\sum_l(w_l^{\mathrm u})^2\to\lim_nn\Lambda_D\) in probability by (11a)--(11b). So
		\[
		W_2\Rightarrow N\big(0,\pi_1\pi_0\mu_\infty'(\pi_1)^2\lim_nn\Lambda_D\big).
		\]
	\end{itemize}
	In both cases Step~9 adds the two variances. By the display for \(\sigma_T^2\) in Section~\ref{subsec:pairwise_metrics} (Section~\ref{appendix:auc_label_part}), \(\pi_1\sigma_{T\mid1}^2+\pi_0\sigma_{T\mid0}^2+\pi_1\pi_0\mu_\infty'(\pi_1)^2=\sigma_T^2\). Under A4, Step~12 gives \(V=V_D\) of (11d), so the total is
	\[
	\sigma_A^2\lim_nn\Gamma_D+\sigma_T^2\lim_nn\Lambda_D+2\gamma_A\lim_nn\Omega_D=\sigma_A^2+\sigma_T^2+2\gamma_A+\lim_nnQ_D,
	\]
	with \(\sigma_A^2\) and \(\gamma_A\) in place of \(\sigma_V^2\) and \(\gamma_0\) in \(Q_D\).
\end{proof}

\subsection{The stability conditions of Austern and Zhou}
\label{appendix:proof_containment}

This proves the statement of Section~\ref{subsec:clt} on the conditions of \citet{austern2020}. Let \(D\) be \(K\)-fold cross-validation with \(K\) fixed and dividing \(n\), so that the design is fixed-size, assume A1, and let aggregation be pooled. Suppose that, uniformly over the training sets of the design,
\[
\|\nabla_j\ell_i^{(r)}\|_2 = o(n^{-1/2}),
\qquad
\|\nabla_j\nabla_{j'}\ell_i^{(r)}\|_2 = o(n^{-1}),
\qquad
\|\nabla_j\nabla_{j'}R_r\|_2 = o(n^{-3/2}),
\]
for \(i\in V_r\) and distinct \(j,j'\in T_r\), with \(R_r:=R(\hat f_{T_r})\). The stability conditions of \citet{austern2020} imply these, since they bound the same first and second differences of a held-out contribution, and of its conditional mean given the training set, which is the second difference of \(R_r\), in \(L_4\), and an \(L_2\) norm is at most the \(L_4\) norm. Then A2(i) holds, \(m\Var(R_r)=\sigma_T^2+o(1)\), \(\mathbb E\{\Var(\ell_i^{(r)}\mid\hat f_{T_r})\}=\sigma_V^2+o(1)\), and \(\gamma_0\) is the quantity \(\rho_{\mathrm{AZ}}\) of \citet{austern2020}.

\begin{proof}
	\noindent\emph{Step 1: The first condition is \(\beta_1'=o(n^{-1/2})\).}\quad
	By its definition in Section~\ref{appendix:stability_assumptions}, \(\beta_1'=\sup\|\nabla_j\ell_i^{(r)}\|_2\), which the first condition makes \(o(n^{-1/2})\).
	
	\noindent\emph{Step 2: The second and third conditions give \(\beta_2=o(n^{-3/2})\).}\quad
	Under pooled aggregation \(\bar\ell^{(r)}=|V_r|^{-1}\sum_{i'\in V_r}\ell_{i'}^{(r)}\), so
	\[
	\nabla_j\nabla_{j'}\bar\ell^{(r)}
	=
	\nabla_j\nabla_{j'}R_r+\frac{1}{|V_r|}\sum_{i'\in V_r}\mathsf u_{i'},
	\qquad
	\mathsf u_{i'}:=\nabla_j\nabla_{j'}\big(\ell_{i'}^{(r)}-R_r\big).
	\]
	Given the training set and its two replacement copies, the \(\mathsf u_{i'}\) for \(i'\in V_r\) are i.i.d.\ with mean zero, because each \(\ell_{i'}^{(r)}\) has conditional mean \(R_r\) under every one of the four training sets. Moreover \(\mathsf u_{i'}\) is \(\nabla_j\nabla_{j'}\ell_{i'}^{(r)}\) minus its conditional mean, so \(\|\mathsf u_{i'}\|_2\le\|\nabla_j\nabla_{j'}\ell_{i'}^{(r)}\|_2\). Hence
	\[
	\Big\|\frac{1}{|V_r|}\sum_{i'\in V_r}\mathsf u_{i'}\Big\|_2^2
	=
	\frac{\|\mathsf u_{i'}\|_2^2}{|V_r|}
	\le
	\frac{\|\nabla_j\nabla_{j'}\ell_{i'}^{(r)}\|_2^2}{|V_r|}.
	\]
	By Minkowski's inequality,
	\[
	\beta_2
	\le
	\|\nabla_j\nabla_{j'}R_r\|_2+|V_r|^{-1/2}\|\nabla_j\nabla_{j'}\ell_{i'}^{(r)}\|_2
	=
	o(n^{-3/2})+O(n^{-1/2})\,o(n^{-1})
	=
	o(n^{-3/2}).
	\]
	
	\noindent\emph{Step 3: A2(i).}\quad
	For \(K\)-fold, \(\max_iv_i=1/n\). The bound (a) of Section~\ref{appendix:stability_assumptions} at \(k=1\) with Steps 1 and 2 gives, for every realized \(I\),
	\[
	\rho_n^{(2+)}(I)
	\le
	\frac{3m}{n}\beta_1'^2+\frac{3m^2}{8}\beta_2^2
	=
	o(n^{-1})+O(n^2)\,o(n^{-3})
	=
	o(n^{-1}),
	\]
	which is A2(i).
	
	\noindent\emph{Step 4: The constants.}\quad
	(4a) By the definitions of \(\phi_i\) and \(\bar\phi\) in Section~\ref{subsec:assumptions},
	\[
	\mathbb E(R_r\mid Z_j=z)-R_{p,n}
	=
	\mathbb E_{Z_i}\big\{\mathbb E(\ell_i^{(r)}\mid Z_i,Z_j=z)-\mathbb E(\ell_i^{(r)}\mid Z_i)\big\}
	=
	\frac{\bar\phi(z)}{|T_r|},
	\]
	so
	\[
	|T_r|\,\Cov\big\{\ell_j^\star,\ \mathbb E(R_r\mid Z_j)\big\}
	=
	\Cov\{\ell_j^\star,\bar\phi(Z_j)\}
	=
	\gamma_0
	\]
	exactly. This is the quantity \(\rho_{\mathrm{AZ}}\) of \citet{austern2020}. Their \(\mathbb E(\beta_1^{l,n}\mid X_1)\) is \(\mathbb E(R_r\mid Z_1)-R_{p,n}\), and their \(\bar{\mathcal L}_{l,n}(X_1)\) is \(\ell_1^\star\).
	
	(4b) The single-observation part of \(R_r-R_{p,n}\) is \(|T_r|^{-1}\sum_{j\in T_r}\bar\phi(Z_j)\), with variance \(\sigma_T^2/|T_r|\). By Steps 1--2 of the proof of (a) in Section~\ref{appendix:stability_assumptions}, applied to \(R_r\) as a function of its training observations, the variance of the rest is at most
	\[
	\frac14\sum_{j<j'}\|\nabla_j\nabla_{j'}R_r\|_2^2
	=
	O(m^2)\,o(n^{-3})
	=
	o(n^{-1}).
	\]
	Hence \(|T_r|\Var(R_r)=\sigma_T^2+o(1)\), which is their \(\sigma_{2,\mathrm{AZ}}^2\).
	
	(4c) Since \(\mathbb E(\ell_i^{(r)}\mid Z_i)=\ell_i^\star\), the law of total variance and the Efron--Stein inequality in the \(m\) training observations at fixed \(Z_i\) give
	\[
	\Var(\ell_i^{(r)})-\sigma_V^2
	=
	\mathbb E\{\Var(\ell_i^{(r)}\mid Z_i)\}
	\le
	\frac12\sum_{j\in T_r}\mathbb E\{(\nabla_j\ell_i^{(r)})^2\}
	\le
	\frac12\,m\,\beta_1'^2
	=
	o(1).
	\]
	With (4b),
	\[
	\mathbb E\{\Var(\ell_i^{(r)}\mid\hat f_{T_r})\}
	=
	\Var(\ell_i^{(r)})-\Var(R_r)
	=
	\sigma_V^2+o(1),
	\]
	which is their \(\sigma_{1,\mathrm{AZ}}^2\).
	
\end{proof}

\subsection{An example where A2(i) fails: one-nearest-neighbour regression}
\label{appendix:clt_local}

This gives the example of Section~\ref{subsec:clt}. Let \(X_1,\dots,X_n\) be i.i.d.\ real-valued with a continuous distribution, and \(Y_i=\mu(X_i)+\epsilon_i\) with \(\mu\) bounded and \(\epsilon_i\) independent of \(X_i\), i.i.d., with mean zero, variance \(\sigma^2>0\), and finite moments of every order, as for Gaussian noise. Let \(D\) be \(K\)-fold cross-validation with \(K\ge2\) fixed and dividing \(n\), the model one-nearest-neighbour regression, which predicts at \(x\) the outcome of the training covariate nearest to \(x\), and the metric squared error. A1 holds: an observation is the nearest training neighbour of a given validated observation with probability at most \(2/m\) given its covariate, so \(|\bar\phi(z)|\le C(1+y^2)\), and the contributions have bounded moments.

The covariates are i.i.d.\ with a continuous distribution and independent of \(I\), so ties have probability zero, and the fold labels of the ordered covariates are a uniformly random arrangement of \(n/K\) copies of each label. With \(\mathrm{nn}(i)\) the training observation nearest to \(X_i\) in the split in which \(i\) is validated,
\[
\widehat R_D=\frac1n\sum_i\big(Y_i-Y_{\mathrm{nn}(i)}\big)^2 .
\]
We show that \(n\rho_n^{(2+)}(I)\ge4\sigma^4\) for every \(n\) and \(I\), so that A2(i) fails, and that the three conditions on differences of Theorem~\ref{thm:clt}(a) hold with \(\widetilde R_D=\widehat R_D\), so that \(\{\widehat R_D-R_{p,n}\}/\Var(\widehat R_D)^{1/2}\Rightarrow N(0,1)\).

\begin{proof}
	\noindent\emph{Step 1: A pair term orthogonal to every single observation.}\quad
	With \(b_i:=\mu(X_{\mathrm{nn}(i)})-\mu(X_i)\),
	\[
	\widehat R_D=\frac1n\sum_i\big(\epsilon_i-\epsilon_{\mathrm{nn}(i)}-b_i\big)^2 .
	\]
	Let
	\[
	P:=-\frac2n\sum_i\epsilon_i\epsilon_{\mathrm{nn}(i)}
	\]
	be its part that multiplies two different noises. Given the covariates, the noises are independent with mean zero, and \(\mathrm{nn}(i)\ne i\). For a square-integrable function \(u\) of one observation \(Z_j\), at least one of \(\epsilon_i\) and \(\epsilon_{\mathrm{nn}(i)}\) is independent of \(Z_j\) and of the other factors given the covariates, so
	\[
	\mathbb E\big\{\epsilon_i\epsilon_{\mathrm{nn}(i)}u(Z_j)\mid \mathbf X\big\}=0,
	\qquad \mathbf X:=(X_1,\dots,X_n).
	\]
	Hence \(P\) is orthogonal to the constants and to every function of a single observation. The projection of \(\widehat R_D\) onto the orthogonal complement of the subspace of such functions is the sum of its components of order two and higher, whose variance is \(\rho_n^{(2+)}(I)\). The same argument shows that \(P\) is uncorrelated with every \(\epsilon_i^2\), with every \(b_i\epsilon_k\), and with every function of \(\mathbf X\), so
	\[
	\Cov(\widehat R_D,P)=\Var(P).
	\]
	
	\noindent\emph{Step 2: A2(i) fails.}\quad
	By Step~1 and the Cauchy--Schwarz inequality applied to \(P\) and that projection,
	\[
	\rho_n^{(2+)}(I)\ge\frac{\Cov(\widehat R_D,P)^2}{\Var(P)}=\Var(P).
	\]
	Given the covariates, the products \(\epsilon_i\epsilon_k\) over distinct unordered pairs are uncorrelated with variance \(\sigma^4\). Each of the \(n\) arrows \(i\to\mathrm{nn}(i)\) contributes \(2/n\) to the coefficient of its pair, and a mutual pair receives two arrows, so with \(n_{\mathrm{mut}}\) mutual pairs
	\[
	\Var(P)=\mathbb E\{\Var(P\mid\mathbf X)\}=\frac{4\sigma^4}{n^2}\,\mathbb E(n+2n_{\mathrm{mut}})\ge\frac{4\sigma^4}{n}.
	\]
	Hence \(n\rho_n^{(2+)}(I)\ge4\sigma^4\) for every \(n\) and \(I\).
	
	\noindent\emph{Step 3: Runs and in-degrees.}\quad
	The nearest training neighbour of a position is the nearer of the nearest positions on either side that carry a different label, so \(j\) can be the nearest neighbour of \(i\) only if every position strictly between them carries the label of \(i\). Hence the in-degree \(d_j:=\#\{i:\mathrm{nn}(i)=j\}\) is at most the total length of the two maximal runs of equal labels adjacent to \(j\). The probability that \(L\) given consecutive positions carry one given label is
	\[
	\prod_{l<L}\frac{n/K-l}{n-l}\le K^{-L},
	\]
	so run lengths have geometric tails, and \(d_j\) has moments of every order, uniformly in \(n\) and \(j\). The same holds at every point at which Theorem~\ref{thm:clt}(a) evaluates differences, since replacing observations by independent copies leaves an i.i.d.\ sample independent of \(I\).
	
	\noindent\emph{Step 4: First differences.}\quad
	Replacing \(Z_j\) by an independent copy \(Z_j'\) changes only the term of \(j\), the terms of the observations whose nearest neighbour is \(j\) before the replacement, and those whose nearest neighbour is \(j\) after it, because removing or adding a candidate that is not an observation's nearest neighbour leaves that neighbour unchanged. With \(A_j\) this set of terms, \(|A_j|\le1+d_j+d_j'\), where \(d_j'\) is the in-degree of \(j\) after the replacement. Each term of \(A_j\) changes from \((Y_i-Y_a)^2\) to \((Y_i^\circ-Y_b)^2\), with \(a\) and \(b\) its nearest neighbours before and after and \(Y_i^\circ=Y_i\) unless \(i=j\), so
	\[
	|\nabla_j\widehat R_D|\le\frac{W_j}{n},
	\qquad
	W_j:=\sum_{i\in A_j}\big\{(Y_i-Y_a)^2+(Y_i^\circ-Y_b)^2\big\}.
	\]
	Given all covariates and \(I\), the set \(A_j\) is fixed and each outcome has bounded eighth moment, since \(\mu\) is bounded, so \(\mathbb E(W_j^4\mid\mathbf X,X_j',I)\le C|A_j|^4\). By Step~3,
	\[
	\mathbb E\{(\nabla_j\widehat R_D)^4\mid I\}\le\frac{C}{n^4}\,\mathbb E\{(1+d_j+d_j')^4\}=O(n^{-4}),
	\qquad
	\mathbb E\{|\nabla_j\widehat R_D|^3\mid I\}=O(n^{-3}),
	\]
	uniformly in \(j\) and over the evaluation points. Summing over \(j\), the first two conditions of Theorem~\ref{thm:clt}(a) hold, at \(O(n^{-3})\) and \(O(n^{-2})\).
	
	\noindent\emph{Step 5: Second differences.}\quad
	By Step~4, \(\nabla_j\nabla_k\widehat R_D=\nabla_j\widehat R_D(\mathcal D_n)-\nabla_j\widehat R_D(\mathcal D_n^{(k)})\) is zero unless replacing \(Z_k\) changes a term of \(A_j\) or which terms belong to it. That requires the original or replaced covariate of \(k\) to lie, in the ordering, within the runs adjacent to the original or replaced covariate of \(j\) or next to them. Call such \(j\) and \(k\) neighbours, and write \(N_k\) for the neighbours of \(k\). The rank of a copy among the sample is uniform and independent of the labels, so the number of copies that fall within \(w\) positions of a given point has mean at most \(2w+1\), and with the geometric run tails of Step~3, \(|N_k|\) has moments of every order, uniformly in \(n\) and \(k\). Since \(|\nabla_j\nabla_k\widehat R_D|\le n^{-1}(W_j+W_j^{(k)})\), with \(W_j^{(k)}\) the bound of Step~4 at \(\mathcal D_n^{(k)}\),
	\[
	\sum_{j\ne k}|\nabla_j\nabla_k\widehat R_D|\,|\nabla_j\widehat R_D|
	\le
	\frac1{n^2}\sum_{j\in N_k}\big(W_j+W_j^{(k)}\big)W_j .
	\]
	With \(F_j:=(W_j+W_j^{(k)})W_j\), the moment bounds of Step~4, which hold at every order because the noise has moments of every order, give \(\|\max_jF_j\|_4\le\{\sum_j\mathbb E(F_j^4)\}^{1/4}=O(n^{1/4})\). Hence
	\[
	\begin{split}
		\mathbb E\Big[\Big\{\sum_{j\ne k}|\nabla_j\nabla_k\widehat R_D|\,|\nabla_j\widehat R_D|\Big\}^2\,\Big|\,I\Big]
		&\le
		\frac1{n^4}\,\mathbb E\Big(|N_k|^2\max_jF_j^2\Big)\\
		&\le
		\frac1{n^4}\,\big\||N_k|\big\|_4^2\,\big\|\max_jF_j\big\|_4^2
		=O(n^{-7/2})
	\end{split}
	\]
	uniformly in \(k\). Summing over \(k\) gives \(O(n^{-5/2})\), the third condition.
	
	\noindent\emph{Step 6: Normality.}\quad
	The folds have equal size, so a permutation of the indices carries any partition to any other, and the observations are exchangeable. Hence \(\mathbb E(\widehat R_D\mid I)=R_{p,n}\) and \(\Var(\widehat R_D\mid I)=\Var(\widehat R_D)\) for every \(I\). By the Efron--Stein inequality and Step~4,
	\[
	n\Var(\widehat R_D)\le\frac n2\sum_j\mathbb E\{(\nabla_j\widehat R_D)^2\}=O(1),
	\]
	and by Step~2, \(n\Var(\widehat R_D)\ge n\rho_n^{(2+)}(I)\ge4\sigma^4\). Every subsequence has a further subsequence along which \(n\Var(\widehat R_D)\) converges to some \(V\ge4\sigma^4\), and along it Theorem~\ref{thm:clt}(a), with Steps 4 and 5, gives \(\sqrt n(\widehat R_D-R_{p,n})\Rightarrow N(0,V)\). Hence \(\{\widehat R_D-R_{p,n}\}/\Var(\widehat R_D)^{1/2}\Rightarrow N(0,1)\) along the whole sequence.
\end{proof}

\section{Further empirical results}
\label{appendix:validation_more}

\subsection{Constants at the training size}
\label{appendix:validation_finite_m}

The exact unit-level identity uses the constants at training size $m$. For the Nadaraya--Watson smoother the expansions below give
\begin{align*}
	\sigma_{V,m}^2&=\sigma_V^2+\frac{b_V}{m}+O(m^{-2}),&
	\sigma_{T,m}^2&=\sigma_T^2+\frac{b_C}{m}+O(m^{-2}),&
	\gamma_{0,m}&=\gamma_0+\frac{b_\gamma}{m}+O(m^{-2}).
\end{align*}
Then $b=b_V+b_C+2b_\gamma$, and the contribution missing from a limiting-constant prediction under a fixed-size design is
\begin{equation}
	\label{eq:general_finite_training_correction}
	\frac{b_V\Gamma_D+b_C\Lambda_D+2b_\gamma\Omega_D}{m}.
\end{equation}
It equals $b/(pn^2)$ for balanced designs. For Monte Carlo cross-validation, \eqref{eq:role_moments} gives instead
\begin{equation}
	\label{eq:mccv_finite_training_correction}
	\frac{1}{pn^2}\left[b+\frac1S\left(\frac pq b_V+\frac qp b_C-2b_\gamma\right)\right].
\end{equation}
Table~\ref{tab:second_constants} gives the limiting constants of the three metrics of Section~\ref{sec:empirical_validation} and the coefficient \(b\) for MSE and the Brier score.

\begin{table}[!htbp]
	\centering
	\caption{Limiting constants of the identity for the three metrics, computed from their definitions. For MSE and Brier score, $b$ is the coefficient of $1/m$ in $\mathcal V_m$. The AUC entries are the limits of the indicator smoothed to $\Phi(u/\varepsilon)$ as $\varepsilon\to0$, extrapolated from $\varepsilon\in\{0.01,0.02,0.04\}$ (Section~\ref{appendix:validation_finite_m}), with pair-level constants $\varsigma^2 = 0.0596$, $\omega_A = 0.0134$, $\omega_A' = 0.0133$, $\omega_A'' = -0.0224$, population AUC $0.8165$, and $\pi_1 = 0.4215$; $b_A$ is given in Section~\ref{appendix:validation_finite_m}.}
	\label{tab:second_constants}
	\footnotesize\setlength{\tabcolsep}{4pt}
	\begin{tabular}{lccccccccc}
		\hline
		& $\sigma_V^2$ & $\sigma_T^2$ & $\gamma_0$ & $b$ & $\mathbb{E}\{\phi_i(Z_j)^2\}$ & $\tau^2$ & $\kappa$ & $\sigma_\chi^2$ & $\xi$ \\
		\hline
		MSE   & 7.812   & 1.555   & $-2.127$ & 20.67  & 36.91  & 35.35  & 32.87  & 61.41  & $-41.69$ \\
		Brier & 0.01365 & 0.00253 & 0.00431  & 0.03849 & 0.2662 & 0.2637 & 0.2491 & 0.2762 & $-0.2434$ \\
		\hline
		& $\sigma_A^2$ & $\sigma_T^2$ & $\gamma_A$ & & & $\tau_A^2$ & $\kappa_A$ & $\sigma_{\chi,A}^2$ & $\xi_A$ \\
		AUC   & 0.1841 & 0.00185 & $-0.00574$ & & & 1.307 & 0.954 & 2.016 & $-1.323$ \\
		\hline
	\end{tabular}
\end{table}

\paragraph{Expected-performance functional.}
Let $\mathcal R_m(P)$ be expected squared-error performance when both the $m$ training observations and the independent validation observation have law $P$. Differentiating the product law in a mean-zero direction yields the exact influence function
\[
\operatorname{IF}_{\mathcal R_m}(z)
=\ell_m^\star(z)-\mathbb E(\ell_m^\star)+\bar\phi_m(z)=H_m(z).
\]
For the fixed-bandwidth Nadaraya--Watson fit with kernel $\mathcal K_h$ define
\[
\mathsf a_x(z)=\frac{\mathcal K_h(x-x_z)}{\mathbb E_P\{\mathcal K_h(x-X)\}},\qquad
f_P(x)=\frac{\mathbb E_P[\mathcal K_h(x-X)Y]}{\mathbb E_P\{\mathcal K_h(x-X)\}},\qquad
\psi_x(z)=\mathsf a_x(z)\{y_z-f_P(x)\},
\]
so that $f_P$ at the population $P$ is the population limit of the smoother of Section~\ref{sec:empirical_validation}.
The leading conditional bias and variance of the fitted prediction are $\beta_P(x)/m$ and $V_{1,P}(x)/m$, where
\[
\beta_P(x)=-\mathbb E_P[\mathsf a_x(Z)\psi_x(Z)],\qquad
V_{1,P}(x)=\mathbb E_P[\psi_x(Z)^2].
\]
With $r(z)=y-f_P(x)$, put
\[
\delta_P(z)=V_{1,P}(x)-2r(z)\beta_P(x),\qquad
\mathcal B(P)=\mathbb E_P[\delta_P(Z)].
\]
Expanding the squared error,
\[
\ell_m^\star(z)=r(z)^2+\frac{\delta_P(z)}m+O_{L_2}(m^{-2}),\qquad
\mathcal R_m(P)=\mathcal R_\infty(P)+\frac{\mathcal B(P)}m+O(m^{-2}).
\]
On the compact covariate domain the kernel denominator is bounded away from zero, so Taylor expansion of the ratio functional controls the remainder and its derivative in the population law. Differentiating the performance expansion yields $H_m=H_\infty+\operatorname{IF}_{\mathcal B}/m+O_{L_2}(m^{-2})$ and hence
\begin{equation}
	\label{eq:b_functional_derivative}
	b=2\mathbb E_P[H_\infty(Z)\operatorname{IF}_{\mathcal B}(Z)].
\end{equation}

\paragraph{Separate coefficients.}
Write $v_0(z)=r(z)^2-\mathcal R_\infty(P)$ and $t_0(z)=\bar\phi_\infty(z)$. For a centered perturbation direction $g$, define
\[
\dot{\mathcal B}[g]=\mathbb E_P[g(Z)\operatorname{IF}_{\mathcal B}(Z)].
\]
Expanding the three covariances gives
\begin{align*}
	b_V&=2\mathbb E_P[v_0\delta_P],\\
	b_C&=2\{\dot{\mathcal B}[t_0]-\mathbb E_P[t_0\delta_P]\},\\
	b_\gamma&=\dot{\mathcal B}[v_0]-\mathbb E_P[v_0\delta_P]+\mathbb E_P[t_0\delta_P].
\end{align*}
The combination $b_V+b_C+2b_\gamma$ is $2\dot{\mathcal B}[v_0+t_0]$, as required by \eqref{eq:b_functional_derivative}. Table~\ref{tab:finite_training_components} reports the coefficients. Tensor-product Gauss--Legendre quadrature of order $24$ per covariate dimension, with Gaussian outcome moments integrated by Gauss--Hermite quadrature and binary outcomes integrated exactly, gives $b=20.66960356$ for MSE and $b=0.03848590$ for Brier score. The order-$16$ and order-$24$ results differ by less than $4.5\times10^{-9}$ and $2.8\times10^{-8}$, respectively. Analytic directional derivatives agree with central finite differences of $\mathcal B(P)$ to numerical precision.

\begin{table}[!htbp]
	\centering
	\caption{Finite-training-size coefficients for the Nadaraya--Watson examples, from the expected-performance functional and its derivatives. The final column is $b_V+b_C+2b_\gamma$.}
	\label{tab:finite_training_components}
	\begin{tabular}{lrrrr}
		\hline
		Metric & $b_V$ & $b_C$ & $b_\gamma$ & $b$\\
		\hline
		MSE & 6.9075726 & -6.0375852 & 9.8998081 & 20.669604 \\
		Brier & -0.0060271482 & 0.012459268 & 0.016026891 & 0.038485902 \\
		\hline
	\end{tabular}
\end{table}

\paragraph{Order three.}
The same expansion of the smoother gives the kernels of order three and the $1/m$ terms of the pair constants. At the limit, $(\tau_{3}^2,\kappa_{3},\sigma_{\chi,3}^2,\xi_{3})=(99.75,\,68.97,\,148.1,\,-99.74)$ for MSE and $(0.6055,\,0.4132,\,0.6886,\,-0.5253)$ for the Brier score, with the signs of order two. The coefficients of $1/m$ in the pair constants are $27.7$, $35.5$, $106.3$, and $-49.6$ for MSE and $0.0020$, $0.0470$, $0.1908$, and $-0.0368$ for the Brier score. With the constants at the training size and the order-three term, at $n=80$ and $2\times10^5$ datasets the measured total variances of two-, four-, and eight-fold, leave-one-out, and Monte Carlo cross-validation at $p=0.5$ and $0.8$ are within one standard error of the prediction for both metrics ($0.997$--$1.003$), and the split noises within two ($0.994$--$1.016$), where the prediction through order two at the limiting constants is off by up to $9\%$ and $5\%$. At the sample sizes of Section~\ref{sec:empirical_validation} the order-three term changes the predicted totals by less than $0.1\%$ and the split noise of $K$-fold by at most $1\%$. With the constants at the training size and the order-three term, the $30$ unit-level split-noise ratios of Tables~\ref{tab:app_mse} and~\ref{tab:app_brier} lie in $0.986$--$1.010$ with standard errors of $0.35$--$0.9\%$, and the total-variance ratios in $0.986$--$1.004$, the deviations at each $n$ shared by the designs through the common datasets.

\paragraph{Finite training size for the AUC.}
The pair contribution is $\mathbb I(u_{ab}>D_{ab})$ with $u_{ab}=f_P(x_a)-f_P(x_b)$ and $D_{ab}=\Delta_T(x_b)-\Delta_T(x_a)$, where $\Delta_T=\widehat f_T-f_P$. Its expectation over the training set cannot be expanded in $\Delta_T$, because the indicator is not smooth, but it can be expanded in the law of $D_{ab}$: as a distribution in $u$,
\[
\mathbb P(D<u)-\mathbb I(u>0)=-\mathbb E(D)\,\delta(u)+\tfrac12\mathbb E(D^2)\,\delta'(u)-\tfrac16\mathbb E(D^3)\,\delta''(u)+O(m^{-2}),
\]
and the same holds for the conditional law given a training observation. The moments of $D_{ab}$ are moments of the sample means of $\psi_x$ and $\mathsf a_x-1$ at $x_a$ and $x_b$ over the training set, which are exact polynomials in $1/m$; under class-stratified folds the training set has fixed class counts, which changes these moments at order $1/m$ through within-class covariances, and we computed both laws. The validation influences $g_1$, $g_0$ and the training influence $\bar\phi_A$ at training size $m$ follow by integrating the expansion over the control, the case, or both, which turns the $\delta^{(k)}$ into derivatives of the score densities on the level sets $f_P(x_a)=f_P(x_b)$. We evaluate these level-set integrals by Gaussian smoothing of the hard indicator at $\varepsilon\in\{0.02,0.01,0.005\}$ and extrapolate $\varepsilon\to0$; the extrapolation error is $O(\varepsilon^2)$ and the three values agree with it. The limits agree with the AUC row of Table~\ref{tab:second_constants}, which extrapolates the quadrature of the smoothed indicator to $\varepsilon\to0$ from the same three widths, and at $\varepsilon=0.02$ both reproduce the earlier values $0.1827$, $0.00160$, $-0.00370$. The hard-indicator limits are
\[
\sigma_A^2=0.18407,\qquad \sigma_T^2=0.001855,\qquad \gamma_A=-0.005735,\qquad \sigma_A^2+\sigma_T^2+2\gamma_A=0.17445,
\]
with population AUC $0.81646$, and the coefficients of $1/m$ are
\[
b_{A,V}=-0.4220,\qquad b_{A,C}=-0.2136,\qquad b_{A,\gamma}=1.1849,\qquad b_A=b_{A,V}+b_{A,C}+2b_{A,\gamma}=1.7341,
\]
the same to four digits under fixed class counts and under a random training composition. The covariance term carries the correction: $\gamma_{A,m}=-0.0057+1.18/m$ changes sign near $m=200$, because the sampling noise of the fitted score smooths the indicator, which is also the direction in which the limits move when the smoothing width is raised. The expected estimate is $\mathbb E(\widehat R_D)=0.81646-0.817/m+O(m^{-2})$.

With $b_A/(pn^2)$ added to the first-order term, the total-variance ratios of the $K$-fold cells of Table~\ref{tab:app_auc} move from $1.038$--$1.060$, $1.025$--$1.036$, and $1.016$--$1.025$ to $0.992$--$0.997$, $0.999$--$1.004$, and $1.005$--$1.006$ at $n=250$, $500$, $1000$ (Table~\ref{tab:app_b}), and the split-noise ratios of Monte Carlo cross-validation at $p=0.5$ from $0.899$, $0.946$, $0.978$ to $1.001$, $1.001$, $1.007$. The pair constants of the AUC are at their limits, with no finite-training correction, and the split noise of $K$-fold, which is pure pair terms, differs from its prediction by up to $7\%$ at $n=250$ and $2\%$ at $n=1000$, decreasing with $n$; the totals at $p=0.9$, where the pair terms weigh most, remain $0.6$--$2.3\%$ high.

\FloatBarrier
\subsection{The numbers behind Figure~\ref{fig:identity}}
\label{appendix:validation_table}

Tables~\ref{tab:app_mse}--\ref{tab:app_auc} give the numbers behind Figure~\ref{fig:identity}. Table~\ref{tab:app_b} adds the design-weighted finite-training corrections of Section~\ref{appendix:validation_finite_m} to the first-order predictions. Total variances are over $10^5$ datasets and split noises over $6000$ datasets with $30$ redraws.

\begin{table}[!htbp]
	\centering
	\caption{Total-variance ratios with the design-weighted finite-training corrections to the three first-order constants ($b_V$, $b_C$, $b_\gamma$; for the AUC $b_{A,V}$, $b_{A,C}$, $b_{A,\gamma}$) added to the limiting prediction through order two. Numerators are the measurements of Tables~\ref{tab:app_mse}--\ref{tab:app_auc}.}
	\label{tab:app_b}
	\begin{tabular}{lccc ccc ccc}
		\hline
		&\multicolumn{3}{c}{MSE}&\multicolumn{3}{c}{Brier}&\multicolumn{3}{c}{AUC}\\
		\cline{2-4}\cline{5-7}\cline{8-10}
		$n$&250&500&1000&250&500&1000&250&500&1000\\
		\hline
		2-fold & 1.001 & 1.003 & 0.999 & 0.989 & 1.002 & 0.997 & 0.992 & 0.999 & 1.006 \\
		5-fold & 0.998 & 1.002 & 0.998 & 0.988 & 0.998 & 1.001 & 0.997 & 1.003 & 1.006 \\
		10-fold & 0.997 & 1.003 & 0.998 & 0.986 & 0.998 & 1.000 & 0.997 & 1.004 & 1.005 \\
		leave-one-out & 0.997 & 1.002 & 0.998 & 0.986 & 0.999 & 1.000 & --- & --- & --- \\
		MCCV p = 0.5 & 0.994 & 1.003 & 0.995 & 0.986 & 0.999 & 1.001 & 0.993 & 1.002 & 1.007 \\
		MCCV p = 0.9 & 0.997 & 1.004 & 0.999 & 0.991 & 0.999 & 0.999 & 1.023 & 1.009 & 1.006 \\
		\hline
	\end{tabular}
\end{table}

\begin{table}[!htbp]
	\centering
	\caption{MSE.\label{tab:app_mse} Total variance and split noise, measured (relative standard error in percent) against the limiting first-order prediction and limiting pair contribution. The ratios divide the measurement by the first-order prediction, by the sum of the two, and (at $m$) by the prediction through order three with the constants at the training size. $K$-fold designs have no first-order split noise.}
	\resizebox{\textwidth}{!}{%
		\begin{tabular}{ll rrrrrr rrrrrr}
			\hline
			& & \multicolumn{6}{c}{total variance} & \multicolumn{6}{c}{split noise} \\
			\cline{3-8}\cline{9-14}
			$n$ & design & measured (\%) & first & second & ratio & ratio (both) & at $m$ & measured (\%) & first & second & ratio & ratio (both) & at $m$ \\
			\hline
			250 & 2-fold & 2.429e-2 (0.47) & 2.045e-2 & 3.158e-3 & 1.188 & 1.029 & 0.998 & 2.957e-3 (0.79) & 0 & 2.905e-3 & --- & 1.018 & 1.003 \\
			250 & 5-fold & 2.142e-2 (0.45) & 2.045e-2 & 6.097e-4 & 1.047 & 1.017 & 0.997 & 3.611e-4 (0.49) & 0 & 3.581e-4 & --- & 1.008 & 0.994 \\
			250 & 10-fold & 2.114e-2 (0.45) & 2.045e-2 & 3.848e-4 & 1.034 & 1.015 & 0.997 & 1.351e-4 (0.43) & 0 & 1.334e-4 & --- & 1.013 & 0.998 \\
			250 & leave-one-out & 2.098e-2 (0.45) & 2.045e-2 & 2.512e-4 & 1.026 & 1.013 & 0.997 & 0 & 0 & 0 & --- & --- & --- \\
			250 & MCCV p = 0.5 & 2.699e-2 (0.45) & 2.590e-2 & 6.486e-4 & 1.042 & 1.017 & 0.993 & 5.745e-3 (0.40) & 5.448e-3 & 3.959e-4 & 1.055 & 0.983 & 0.992 \\
			250 & MCCV p = 0.9 & 5.153e-2 (0.46) & 5.035e-2 & 9.151e-4 & 1.023 & 1.005 & 0.996 & 3.054e-2 (0.42) & 2.990e-2 & 6.638e-4 & 1.021 & 0.999 & 0.997 \\
			500 & 2-fold & 1.122e-2 (0.46) & 1.023e-2 & 7.904e-4 & 1.097 & 1.018 & 1.003 & 7.326e-4 (0.77) & 0 & 7.278e-4 & --- & 1.007 & 0.999 \\
			500 & 5-fold & 1.051e-2 (0.45) & 1.023e-2 & 1.528e-4 & 1.027 & 1.012 & 1.002 & 9.097e-5 (0.49) & 0 & 9.026e-5 & --- & 1.008 & 1.001 \\
			500 & 10-fold & 1.045e-2 (0.45) & 1.023e-2 & 9.651e-5 & 1.022 & 1.012 & 1.003 & 3.390e-5 (0.41) & 0 & 3.398e-5 & --- & 0.998 & 0.991 \\
			500 & leave-one-out & 1.040e-2 (0.45) & 1.023e-2 & 6.250e-5 & 1.017 & 1.010 & 1.002 & 0 & 0 & 0 & --- & --- & --- \\
			500 & MCCV p = 0.5 & 1.330e-2 (0.45) & 1.295e-2 & 1.619e-4 & 1.027 & 1.014 & 1.003 & 2.787e-3 (0.37) & 2.724e-3 & 9.923e-5 & 1.023 & 0.987 & 0.992 \\
			500 & MCCV p = 0.9 & 2.560e-2 (0.45) & 2.517e-2 & 2.285e-4 & 1.017 & 1.008 & 1.004 & 1.516e-2 (0.39) & 1.495e-2 & 1.660e-4 & 1.014 & 1.003 & 1.002 \\
			1000 & 2-fold & 5.349e-3 (0.45) & 5.113e-3 & 1.977e-4 & 1.046 & 1.007 & 0.999 & 1.846e-4 (0.76) & 0 & 1.821e-4 & --- & 1.014 & 1.010 \\
			1000 & 5-fold & 5.166e-3 (0.45) & 5.113e-3 & 3.825e-5 & 1.010 & 1.003 & 0.998 & 2.269e-5 (0.47) & 0 & 2.266e-5 & --- & 1.001 & 0.998 \\
			1000 & 10-fold & 5.152e-3 (0.45) & 5.113e-3 & 2.416e-5 & 1.008 & 1.003 & 0.998 & 8.622e-6 (0.40) & 0 & 8.574e-6 & --- & 1.006 & 1.002 \\
			1000 & leave-one-out & 5.140e-3 (0.45) & 5.113e-3 & 1.559e-5 & 1.005 & 1.002 & 0.998 & 0 & 0 & 0 & --- & --- & --- \\
			1000 & MCCV p = 0.5 & 6.523e-3 (0.45) & 6.475e-3 & 4.045e-5 & 1.007 & 1.001 & 0.995 & 1.392e-3 (0.36) & 1.362e-3 & 2.484e-5 & 1.022 & 1.004 & 1.006 \\
			1000 & MCCV p = 0.9 & 1.266e-2 (0.45) & 1.259e-2 & 5.710e-5 & 1.006 & 1.001 & 0.999 & 7.596e-3 (0.36) & 7.474e-3 & 4.151e-5 & 1.016 & 1.011 & 1.010 \\
			\hline
	\end{tabular}}
\end{table}

\begin{table}[!htbp]
	\centering
	\caption{Brier score.\label{tab:app_brier} Total variance and split noise, measured (relative standard error in percent) against the limiting first-order prediction and limiting pair contribution. The ratios divide the measurement by the first-order prediction, by the sum of the two, and (at $m$) by the prediction through order three with the constants at the training size. $K$-fold designs have no first-order split noise.}
	\resizebox{\textwidth}{!}{%
		\begin{tabular}{ll rrrrrr rrrrrr}
			\hline
			& & \multicolumn{6}{c}{total variance} & \multicolumn{6}{c}{split noise} \\
			\cline{3-8}\cline{9-14}
			$n$ & design & measured (\%) & first & second & ratio & ratio (both) & at $m$ & measured (\%) & first & second & ratio & ratio (both) & at $m$ \\
			\hline
			250 & 2-fold & 1.199e-4 (0.47) & 9.918e-5 & 2.079e-5 & 1.209 & 0.999 & 0.988 & 1.799e-5 (0.73) & 0 & 1.813e-5 & --- & 0.992 & 0.986 \\
			250 & 5-fold & 1.039e-4 (0.45) & 9.918e-5 & 5.184e-6 & 1.048 & 0.996 & 0.988 & 2.556e-6 (0.47) & 0 & 2.531e-6 & --- & 1.010 & 1.002 \\
			250 & 10-fold & 1.020e-4 (0.45) & 9.918e-5 & 3.625e-6 & 1.028 & 0.992 & 0.986 & 9.876e-7 (0.40) & 0 & 9.743e-7 & --- & 1.014 & 1.005 \\
			250 & leave-one-out & 1.010e-4 (0.45) & 9.918e-5 & 2.649e-6 & 1.018 & 0.992 & 0.986 & 0 & 0 & 0 & --- & --- & --- \\
			250 & MCCV p = 0.5 & 1.068e-4 (0.45) & 1.022e-4 & 4.961e-6 & 1.045 & 0.997 & 0.986 & 5.216e-6 (0.56) & 3.024e-6 & 2.298e-6 & 1.725 & 0.980 & 0.991 \\
			250 & MCCV p = 0.9 & 1.515e-4 (0.45) & 1.450e-4 & 7.346e-6 & 1.045 & 0.994 & 0.991 & 5.037e-5 (0.45) & 4.579e-5 & 4.695e-6 & 1.100 & 0.998 & 1.000 \\
			500 & 2-fold & 5.521e-5 (0.46) & 4.959e-5 & 5.203e-6 & 1.113 & 1.008 & 1.002 & 4.547e-6 (0.74) & 0 & 4.542e-6 & --- & 1.001 & 0.998 \\
			500 & 5-fold & 5.096e-5 (0.45) & 4.959e-5 & 1.298e-6 & 1.028 & 1.001 & 0.998 & 6.410e-7 (0.46) & 0 & 6.380e-7 & --- & 1.005 & 1.001 \\
			500 & 10-fold & 5.058e-5 (0.45) & 4.959e-5 & 9.077e-7 & 1.020 & 1.002 & 0.998 & 2.511e-7 (0.40) & 0 & 2.482e-7 & --- & 1.012 & 1.008 \\
			500 & leave-one-out & 5.033e-5 (0.45) & 4.959e-5 & 6.593e-7 & 1.015 & 1.002 & 0.998 & 0 & 0 & 0 & --- & --- & --- \\
			500 & MCCV p = 0.5 & 5.255e-5 (0.45) & 5.110e-5 & 1.237e-6 & 1.028 & 1.004 & 0.998 & 2.073e-6 (0.53) & 1.512e-6 & 5.759e-7 & 1.371 & 0.993 & 1.001 \\
			500 & MCCV p = 0.9 & 7.437e-5 (0.45) & 7.248e-5 & 1.834e-6 & 1.026 & 1.001 & 0.999 & 2.393e-5 (0.40) & 2.289e-5 & 1.174e-6 & 1.045 & 0.994 & 0.996 \\
			1000 & 2-fold & 2.610e-5 (0.45) & 2.479e-5 & 1.301e-6 & 1.053 & 1.000 & 0.997 & 1.146e-6 (0.75) & 0 & 1.137e-6 & --- & 1.008 & 1.007 \\
			1000 & 5-fold & 2.519e-5 (0.45) & 2.479e-5 & 3.247e-7 & 1.016 & 1.003 & 1.001 & 1.604e-7 (0.46) & 0 & 1.601e-7 & --- & 1.002 & 1.000 \\
			1000 & 10-fold & 2.507e-5 (0.45) & 2.479e-5 & 2.271e-7 & 1.011 & 1.002 & 1.000 & 6.245e-8 (0.40) & 0 & 6.261e-8 & --- & 0.997 & 0.995 \\
			1000 & leave-one-out & 2.500e-5 (0.45) & 2.479e-5 & 1.645e-7 & 1.008 & 1.002 & 1.000 & 0 & 0 & 0 & --- & --- & --- \\
			1000 & MCCV p = 0.5 & 2.596e-5 (0.45) & 2.555e-5 & 3.088e-7 & 1.016 & 1.004 & 1.001 & 8.928e-7 (0.46) & 7.561e-7 & 1.441e-7 & 1.181 & 0.992 & 0.997 \\
			1000 & MCCV p = 0.9 & 3.671e-5 (0.45) & 3.624e-5 & 4.581e-7 & 1.013 & 1.000 & 0.999 & 1.184e-5 (0.37) & 1.145e-5 & 2.936e-7 & 1.034 & 1.008 & 1.009 \\
			\hline
	\end{tabular}}
\end{table}

\begin{table}[!htbp]
	\centering
	\caption{AUC.\label{tab:app_auc} Total variance and split noise, measured (relative standard error in percent) against the limiting first-order prediction and limiting pair contribution. The ratios divide the measurement by the first-order prediction, by the sum of the two, and (at $m$) by the sum with the first-order constants at the training size, which adds $b_A$. $K$-fold designs have no first-order split noise.}
	\resizebox{\textwidth}{!}{%
		\begin{tabular}{ll rrrrrr rrrrrr}
			\hline
			& & \multicolumn{6}{c}{total variance} & \multicolumn{6}{c}{split noise} \\
			\cline{3-8}\cline{9-14}
			$n$ & design & measured (\%) & first & second & ratio & ratio (both) & at $m$ & measured (\%) & first & second & ratio & ratio (both) & at $m$ \\
			\hline
			250 & 2-fold & 8.567e-4 (0.49) & 6.977e-4 & 1.107e-4 & 1.228 & 1.060 & 0.992 & 9.022e-5 (0.90) & 0 & 9.659e-5 & --- & 0.934 & 0.934 \\
			250 & 5-fold & 7.703e-4 (0.45) & 6.977e-4 & 4.049e-5 & 1.104 & 1.044 & 0.997 & 2.772e-5 (0.49) & 0 & 2.637e-5 & --- & 1.051 & 1.051 \\
			250 & 10-fold & 7.788e-4 (0.45) & 6.977e-4 & 5.280e-5 & 1.116 & 1.038 & 0.997 & 4.041e-5 (0.39) & 0 & 3.869e-5 & --- & 1.044 & 1.044 \\
			250 & MCCV p = 0.5 & 8.455e-4 (0.45) & 7.766e-4 & 2.907e-5 & 1.089 & 1.049 & 0.993 & 8.434e-5 (0.40) & 7.894e-5 & 1.492e-5 & 1.068 & 0.899 & 1.001 \\
			250 & MCCV p = 0.9 & 1.496e-3 (0.45) & 1.365e-3 & 7.706e-5 & 1.096 & 1.037 & 1.023 & 7.556e-4 (0.39) & 6.672e-4 & 6.295e-5 & 1.132 & 1.035 & 1.051 \\
			500 & 2-fold & 3.901e-4 (0.46) & 3.488e-4 & 2.772e-5 & 1.118 & 1.036 & 0.999 & 2.354e-5 (0.85) & 0 & 2.420e-5 & --- & 0.973 & 0.973 \\
			500 & 5-fold & 3.689e-4 (0.45) & 3.488e-4 & 1.014e-5 & 1.058 & 1.028 & 1.003 & 6.789e-6 (0.45) & 0 & 6.617e-6 & --- & 1.026 & 1.026 \\
			500 & 10-fold & 3.711e-4 (0.45) & 3.488e-4 & 1.321e-5 & 1.064 & 1.025 & 1.004 & 9.969e-6 (0.37) & 0 & 9.694e-6 & --- & 1.028 & 1.028 \\
			500 & MCCV p = 0.5 & 4.078e-4 (0.45) & 3.883e-4 & 7.259e-6 & 1.050 & 1.031 & 1.002 & 4.086e-5 (0.37) & 3.947e-5 & 3.737e-6 & 1.035 & 0.946 & 1.001 \\
			500 & MCCV p = 0.9 & 7.131e-4 (0.45) & 6.824e-4 & 1.926e-5 & 1.045 & 1.016 & 1.009 & 3.566e-4 (0.37) & 3.336e-4 & 1.574e-5 & 1.069 & 1.021 & 1.029 \\
			1000 & 2-fold & 1.859e-4 (0.45) & 1.744e-4 & 6.934e-6 & 1.066 & 1.025 & 1.006 & 6.042e-6 (0.82) & 0 & 6.055e-6 & --- & 0.998 & 0.998 \\
			1000 & 5-fold & 1.802e-4 (0.45) & 1.744e-4 & 2.535e-6 & 1.033 & 1.018 & 1.006 & 1.691e-6 (0.42) & 0 & 1.657e-6 & --- & 1.020 & 1.020 \\
			1000 & 10-fold & 1.806e-4 (0.45) & 1.744e-4 & 3.304e-6 & 1.035 & 1.016 & 1.005 & 2.457e-6 (0.36) & 0 & 2.426e-6 & --- & 1.013 & 1.013 \\
			1000 & MCCV p = 0.5 & 2.003e-4 (0.45) & 1.942e-4 & 1.814e-6 & 1.032 & 1.022 & 1.007 & 2.021e-5 (0.36) & 1.974e-5 & 9.353e-7 & 1.024 & 0.978 & 1.007 \\
			1000 & MCCV p = 0.9 & 3.493e-4 (0.45) & 3.412e-4 & 4.813e-6 & 1.024 & 1.009 & 1.006 & 1.710e-4 (0.35) & 1.668e-4 & 3.935e-6 & 1.025 & 1.002 & 1.006 \\
			\hline
	\end{tabular}}
\end{table}

\FloatBarrier
\subsection{Exactness on the sample mean}
\label{appendix:validation_mean}

For the sample mean under squared error every interaction is between two observations, so the identity stops at order two, is exact, and can be checked against a closed form.
The population is $\mathrm{Gamma}(2)$, so that $\mu_4 \ne 3\sigma^4$ and $\sigma_V^2 = \mu_4 - \sigma^4$ is tested as well. All seven constants are in closed form. With $\epsilon:=Y-\mathbb E(Y)$, the pair kernels are $\phi^{(2)}(Z_i;Z_j)=-2\epsilon_i\epsilon_j$ and $\chi(Z_j,Z_k)=2\epsilon_j\epsilon_k$, so $\tau_n^2=\kappa_n=\sigma_{\chi,n}^2=-\xi_n=4\sigma^4$, the values of Section~\ref{subsec:variance_decomposition}. At $n = 24$, for one realized fold assignment under each of $2$-, $4$-, $8$-fold, leave-one-out, $3\times4$-fold, and Monte Carlo cross-validation at $(p, S) = (0.5, 5)$ and $(0.75, 10)$, the variance of $\widehat R_D$ over $2\times10^5$ datasets is compared with the identity at the realized weights. The ratios of measured to predicted variance lie in $[0.998, 1.013]$, each within $1.6$ standard errors of one, and the first-order part alone is low by $12\%$ at $2$-fold, $4\%$ at $4$-fold and at Monte Carlo $p = 0.5$, and $2\%$ at leave-one-out. For leave-one-out the variance has the closed form $\frac{n}{(n-1)^2}\{\mu_4 - \frac{n-3}{n-1}\sigma^4\}$, which the identity reproduces to four digits.

\FloatBarrier
\subsection{Simulation check of balance at order two}
\label{appendix:validation_pair_balanced}

The check compares two designs of five-fold cross-validation repeated six times, on \(n=25\) observations: the pair-balanced design given by the six parallel classes of the affine plane of order five, with the observations in random order, and six independent random partitions. Data are \(Y=\sin(3X)+\epsilon\) with \(X\) uniform on \([0,1]\) and \(\epsilon\sim N(0,0.5^2)\). Two models are evaluated under squared error: the training-set mean, whose components of order three and higher are zero, so that the pair-balanced design is predicted to have no split noise at all, and one-nearest-neighbour regression, whose pair block is of first order and which has interactions of every order through its covariates. For each of 1500 datasets the estimate is computed under 60 inclusion structures of each design, giving \(\Var_I(\widehat R_D\mid\mathcal D_n)\) at that dataset, and the split noise is the average over datasets. The sampling variance is the variance across datasets of the average over inclusion structures.

\begin{table}[!htbp]
	\centering
	\caption{Split noise and sampling variance of pair-balanced and independently repeated five-fold cross-validation, \(n=25\), six repetitions, 1500 datasets. For the sample mean the identity's exact finite-\(n\) value of the independent design's split noise is \(6.25\times10^{-5}\), from Step~3 of Section~\ref{appendix:proof_order_balance} with \(\tau^2=\kappa=\sigma_\chi^2=-\xi=4\Var(Y)^2\).}
	\label{tab:pair_balanced}
	\begin{tabular}{lccc}
		\hline
		& split noise, pair-balanced & split noise, independent & sampling variance \\
		\hline
		sample mean & \(3.5\times10^{-33}\) & \(6.07\times10^{-5}\) & \(9.92\times10^{-3}\) \\
		1-NN regression & \(1.39\times10^{-4}\) & \(9.78\times10^{-4}\) & \(3.54\times10^{-2}\) \\
		\hline
	\end{tabular}
\end{table}

For the sample mean the pair-balanced design's split noise is zero to machine precision, and the independent design's agrees with the identity to three percent. For one-nearest-neighbour the pair-balanced design removes six sevenths of the split noise, from 2.8 percent of the sampling variance to 0.4 percent, the remainder being its split noise of order three and higher, which pair balance leaves in place. The sampling variances of the two designs agree to 0.1 percent for both models, as they must at a common training size.

\FloatBarrier
\subsection{Simulation check of normality where A2(i) fails}
\label{appendix:validation_soclt}

This checks the conclusion of Theorem~\ref{thm:clt}, that the estimate is normal at the identity's full variance although the components of order two and higher carry variance of order \(1/n\), for the example of Section~\ref{appendix:clt_local}. The check uses squared error and five-fold cross-validation with one repetition, and each replicate draws a new dataset and a new partition, so the reported variances are unconditional. It uses $X \sim \mathrm{Uniform}(0,1)$ and $Y = \sin(3X) + \epsilon$ with $\epsilon \sim N(0, 0.5^2)$, at $n \in \{250, 1000, 4000\}$ with $20{,}000$, $10{,}000$, and $4{,}000$ datasets. One-nearest-neighbour regression is a linear smoother, whose prediction is a linear combination of the training outcomes with smoother weights that depend only on the covariates and the partition. Under squared error the estimate given these is then a quadratic form $\epsilon^\top\mathbf Q\epsilon+\mathbf L^\top\epsilon+c$ in the noises $\epsilon_k=Y_k-\mathbb E(Y_k\mid X_k)$, with $\mathbf Q$, $\mathbf L$, and $c$ computed from the realized smoother weights, and with Gaussian noise of variance $\sigma^2$
\[
\Var(\widehat R_D\mid\mathbf X,I)=2\sigma^4\sum_jQ_{jj}^2+\sigma^2\sum_jL_j^2+4\sigma^4\sum_{j<k}Q_{jk}^2
\]
exactly, the last sum being the \emph{pair block}.

The unconditional variance $V := \mathbb{E}\{\Var(\widehat R_D \mid \mathbf X, I)\} + \Var\{\mathbb{E}(\widehat R_D \mid \mathbf X, I)\}$ is computed from the exact conditional moments, by averaging the conditional variances and taking the sample variance of the conditional means over datasets, and its pair part is the pair block, averaged in the same way. Table~\ref{tab:soclt} reports the ratio of the sample variance of $\widehat R_D$ across datasets to $V$, the pair share $s$, which is the expected pair block as a fraction of $V$, the skewness and excess kurtosis of $\widehat R_D$, and two coverages. The first is the fraction of datasets with $|\widehat R_D - \bar R| \le 1.96\sqrt V$, with $\bar R$ the mean over datasets. It uses the exact variance and so tests only the shape of the law. The second puts $(1 - s)V$, the variance with the pair block omitted, in place of $V$, and under a normal law it is $2\Phi(1.96\sqrt{1 - s}) - 1$. The pair block is the variance of the noise pair term given the covariates, not the pair component of A2, and in the full observations it spreads over orders two and higher. The argument of Steps 1 and 2 of Section~\ref{appendix:clt_local} shows that the components of order two and higher carry at least the expected pair block, so the single-observation variance of Theorem~\ref{thm:variance_identity} is at most $(1 - s)V$, and an interval at that variance covers at most what the $(1 - s)V$ column reports.

\begin{table}[!htbp]
	\centering
	\caption{Theorem~\ref{thm:clt} in simulation: five-fold cross-validation, squared error, $Y = \sin(3X) + N(0, 0.5^2)$. $V$ is the identity's unconditional variance from the exact conditional moments and $s$ the share of $V$ in the pair block. Coverage is the fraction of datasets within $1.96$ standard deviations of the mean, at the full variance $V$ and at $(1 - s)V$, the variance with the pair block omitted, with the normal-law prediction $2\Phi(1.96\sqrt{1 - s}) - 1$ in parentheses.}
	\label{tab:soclt}
	\resizebox{\textwidth}{!}{%
		\begin{tabular}{llrcccccc}
			\hline
			Model & $n$ & datasets & $\widehat{\Var}/V$ & $s$ & skew & ex.\ kurt. & cov.\ $V$ & cov.\ $(1-s)V$ \\
			\hline
			1-NN & 250 & 20{,}000 & 1.00 & 0.405 & 0.31 & 0.11 & 0.951 & 0.871 (0.869) \\
			1-NN & 1000 & 10{,}000 & 1.01 & 0.405 & 0.13 & 0.06 & 0.949 & 0.870 (0.869) \\
			1-NN & 4000 & 4{,}000 & 0.99 & 0.405 & 0.07 & $-0.05$ & 0.952 & 0.871 (0.869) \\
			\hline
	\end{tabular}}
\end{table}

\begin{figure}[!htbp]
	\centering
	\includegraphics[width=0.55\textwidth,trim=0 0 238bp 24bp,clip]{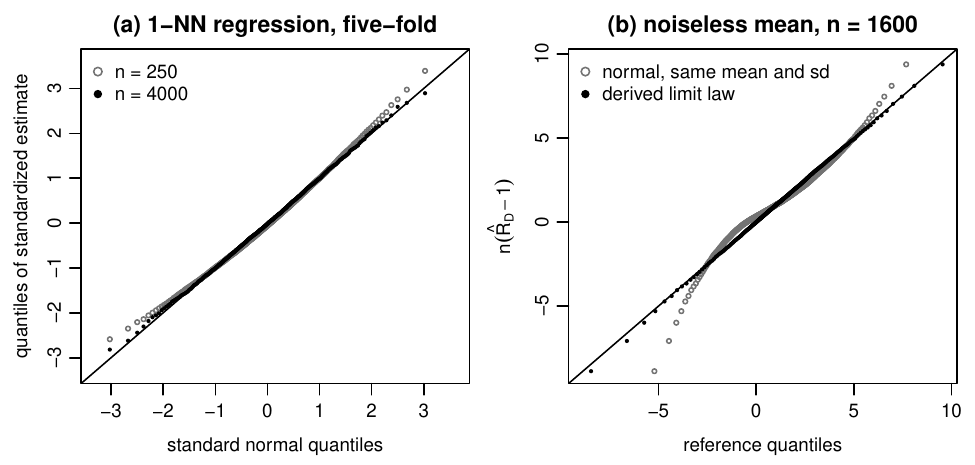}
	\caption{Theorem~\ref{thm:clt} in simulation. Quantiles of the standardized five-fold estimate for one-nearest-neighbour regression, $(\widehat R_D - \bar R)/\sqrt V$ with $V$ the identity's variance, against standard normal quantiles at $n = 250$ (open) and $n = 4000$ (filled). The pair block is $40\%$ of $V$ at both sample sizes.}
	\label{fig:soclt}
\end{figure}

For one-nearest-neighbour regression the measured variance is the identity's value to within one percent at every $n$ (the standard error of the ratio is $0.01$--$0.02$), and the pair share is $0.405$ at $n = 250$, $1000$, and $4000$. The interaction component is of first order and does not vanish as $n$ grows, as Section~\ref{appendix:clt_local} shows. The standardized law is nevertheless consistent with the normal approximation at the full variance, with skewness falling from $0.31$ to $0.07$ and coverage $0.949$--$0.952$ at every $n$ (Figure~\ref{fig:soclt}). An interval that omits the pair block covers $0.870$--$0.871$ at every $n$, which is $2\Phi(1.96\sqrt{0.595}) - 1 = 0.869$, and $0.794$--$0.799$ at the $90\%$ level against $0.795$. A first-order interval can do no better, and the shortfall does not decrease with $n$.

\FloatBarrier
\section{Notation}
\label{appendix:notation}

{\footnotesize
	\begin{longtable}{@{}p{0.27\textwidth}p{0.55\textwidth}p{0.12\textwidth}@{}}
		\hline
		Symbol & Meaning & Section\\
		\hline
		\endhead
		\multicolumn{3}{@{}l}{\emph{Data, design, estimate}}\\
		$\mathcal D_n$, $Z_i=(X_i,Y_i)$, $P$ & sample, observation, population & \ref{subsec:data_algorithm_metric}\\
		$\hat f_T$; $\ell_i(f)$; $\mathcal B$ & model fit to training set $T$; contribution of unit $i$; set of basic units & \ref{subsec:data_algorithm_metric}\\
		$D$; $p$, $q=1-p$; $S$; $K$, $R$ & design; training and validation fractions; number of splits; folds, repetitions & \ref{subsec:cv_design}\\
		$T_r$, $V_r$; $I=(I_{ir})$; $\tilde I_{ir}$ & training and validation sets of split $r$; inclusion structure; bootstrap multiplicity & \ref{subsec:cv_design}\\
		$m$; $M$; $N_i$ & common training size; number of evaluations; times unit $i$ is validated & \ref{subsec:cv_design}, \ref{subsec:weighted_representation}\\
		$\widehat R_D$; $R_{p,n}$; $\widehat R_{K,R}$ & estimate; its target; average of $R$ repeated $K$-fold estimates & \ref{subsec:cv_design}, \ref{subsec:split_noise}\\
		\multicolumn{3}{@{}l}{\emph{Weights}}\\
		$v_i$, $w_i$ & validation and training (role) weights & \ref{subsec:weighted_representation}\\
		$c^{(k)}_{i;\mathcal J}$, $d^{(k)}_{\mathcal S}$; $c_{ij}$, $d_{ij}$ & weights of order $k$; pair weights & \ref{subsec:representation}\\
		$\Gamma_D$, $\Lambda_D$, $\Omega_D$ & first-order role moments, $\mathbb E_I$ of $\sum v_i^2$, $\sum w_i^2$, $\sum v_iw_i$ & \ref{subsec:variance_decomposition}\\
		$\Theta_D^{(k)}$, $\Theta_D'^{(k)}$, $\Psi_D^{(k)}$, $\Psi_D'^{(k)}$ & role moments of order $k$ & \ref{subsec:variance_decomposition}\\
		\multicolumn{3}{@{}l}{\emph{Influences and constants}}\\
		$\ell_i^\star$; $\phi_i$; $\bar\phi$ & validation influence; training influence on unit $i$; on expected performance & \ref{subsec:assumptions}\\
		$g_{\mathcal S}$; $h_{\mathcal S}$, $h_i$; $\varrho_D$ & Hoeffding component; components of $\widehat R_D$ given $I$; their part on two or more observations & \ref{subsec:assumptions}, \ref{subsec:representation}\\
		$\rho_n^{(k)}(I)$; $\rho_n^{(k+)}(I)$ & variance of the components of order $k$; of orders $k$ and above & \ref{subsec:assumptions}\\
		$\nabla_j$, $\nabla_j\nabla_k$ & change when $Z_j$, or $Z_j$ and $Z_k$, are replaced by independent copies & \ref{subsec:assumptions}\\
		$R(f)$ & expected performance of $f$ & \ref{subsec:representation}\\
		$\phi^{(k)}$, $\chi^{(k)}$; $\chi$ & interactions of order $k$; training--training interaction at order two & \ref{subsec:representation}\\
		$\sigma_V^2$, $\sigma_T^2$, $\gamma_0$ & variances of the two influences and their covariance & \ref{subsec:representation}\\
		$\tau_{k,n}^2$, $\kappa_{k,n}$, $\sigma_{\chi,k,n}^2$, $\xi_{k,n}$ & constants of order $k$; pair constants $\tau_n^2$, $\kappa_n$, $\sigma_{\chi,n}^2$, $\xi_n$ at $k=2$, limits $\tau^2$, $\kappa$, $\sigma_\chi^2$, $\xi$ & \ref{subsec:variance_decomposition}, \ref{subsec:loo}\\
		$f_p^\star$; $\psi$ & limit of the fitted model; influence function of the fit & \ref{appendix:stability_assumptions}\\
		$\sigma_\ell^2$; $C_\ell$; $\omega$, $\gamma^{\mathrm{BG}}$ & variance of a held-out contribution; covariance at independent training sets; within- and across-fold covariances & \ref{subsec:refitting}\\
		$\Sigma$ & covariance matrix of $(\ell_i^\star,\bar\phi(Z_i))$ & \ref{subsec:design_comparison}\\
		\multicolumn{3}{@{}l}{\emph{Design comparison}}\\
		$H$, $H_m$; $\mathcal V$, $\mathcal V_m$, $\mathcal V_\infty$ & combined influence; its variance, at training size $m$, and the limit & \ref{subsec:design_comparison}, \ref{subsec:endpoint_variance_optimality}\\
		$Q_D$ & imbalance penalty & \ref{subsec:design_comparison}\\
		$\Var_{\mathrm{samp}}(m)$; $\Pi_n^{(k)}$; $\mathcal N_D^{(k)}$, $\mathcal N_{K,1}$ & sampling variance at training size $m$; its term of order $k$; split noise of order $k$, of one partition & \ref{subsec:split_noise}\\
		$\lambda_{ij}$, $\bar\lambda$ & partitions in which $i$ and $j$ share a fold, and their average & \ref{subsec:design_comparisons_fixed_p}\\
		$G(K)$; $b$, $\gamma$ & split-noise coefficient; coefficient and rate in $\mathcal V_m-\mathcal V_\infty$ & \ref{subsec:loo}\\
		\multicolumn{3}{@{}l}{\emph{Pairwise metrics}}\\
		$\mathcal C(s,t)$; $\bar\phi_Y$ & pair contribution; training influence given the labels & \ref{subsec:data_algorithm_metric}, \ref{subsec:assumptions}\\
		$n_y$, $\pi_y$; $m_1$, $m_0$ & class count and population proportion; training cases and controls & \ref{subsec:cv_design}, \ref{subsec:weighted_representation}, \ref{subsec:pairwise_metrics}\\
		$\tilde v_i$ & validation weight through the pairs & \ref{subsec:weighted_representation}\\
		$\sigma_{V\mid y}^2$, $\sigma_{T\mid y}^2$, $\gamma_{0\mid y}$; $\sigma_A^2$, $\gamma_A$ & constants within class $y$; pairwise counterparts of $\sigma_V^2$ and $\gamma_0$ & \ref{subsec:pairwise_metrics}\\
		$\mu_m(m_1/m)$, $\mu_\infty'$; $R_{p,n}(Y)$ & expected value of the metric for $m_1$ cases in a training set of $m$, limiting derivative of $\mu_m$ in the case fraction; target given the labels & \ref{subsec:cv_design}, \ref{subsec:pairwise_metrics}\\
		$\mathcal W_{1,n}$; $\tau_A^2$, $\kappa_A$, $\sigma_{\chi,A}^2$, $\xi_A$ & first-order sampling variance of a pairwise metric; pair-level constants & \ref{subsec:folds_auc}\\
		$\varsigma^2$, $\omega_A$, $\omega_A'$, $\omega_A''$; $\lambda_1,\dots,\lambda_4$, $G_A$, $G_s$; $b_A$, $\ell_A$ & pair-specific constants; fold-count coefficients, and $G$ at same-class pair constants; finite-training and class-composition corrections & \ref{subsec:folds_auc}, \ref{app:endpoint_auc}\\
		\multicolumn{3}{@{}l}{\emph{Normality}}\\
		$\widetilde R_D$; $V$ & approximation to the estimate; limiting variance & \ref{sec:normality}\\
		\hline
	\end{longtable}
}

\end{document}